\documentclass[11pt]{article}

\usepackage[final]{acl}

\usepackage{times}
\usepackage{latexsym}
\usepackage{booktabs}

\usepackage[T1]{fontenc}

\usepackage[utf8]{inputenc}

\usepackage{microtype}

\usepackage{inconsolata}
\usepackage{array}

\usepackage{subcaption}
\usepackage{graphicx}
\usepackage{adjustbox}
\usepackage{float}
\usepackage{amsmath}
\usepackage{algorithm}
\usepackage{algorithmic}
\usepackage{amsthm}
\usepackage{amsmath}
\usepackage[table]{xcolor}
\usepackage{enumitem}
\usepackage{array}
\usepackage{makecell}
\usepackage{multirow}
\usepackage{amsthm}
\usepackage{pifont}
\usepackage{longtable}
\newtheorem{problem}{Problem}
\newtheorem{definition}{Definition}

\newcommand{\blfootnote}[1]{%
  \begingroup
  \renewcommand\thefootnote{}\footnote{#1}%
  \addtocounter{footnote}{-1}%
  \endgroup
}
\title{Train Smarter, Not Harder: Switching Signal-Guided 
Training in Active Learning}

\definecolor{porpol}{RGB}{120, 0, 163}

\newcommand{\fone}{F1}

\author{
  Nagham Omar\textsuperscript{*} \quad
  Maya Rozenshtein\textsuperscript{*} \quad
  Evgeny Mishlyakov\textsuperscript{*} \quad
  Avigdor Gal \\
  Faculty of Data and Decision Sciences, Technion \\
  {\small\texttt{\{nagham.omar, rmaya, ym\}@campus.technion.ac.il}}, 
  {\small\texttt{avigal@technion.ac.il}}
}

\begin{document}

\maketitle

\blfootnote{\textsuperscript{*}Equal contribution.}
\blfootnote{Accepted to EMNLP 2026 Main Conference.}
\begin{abstract}
Training strategy, namely whether to retrain from scratch or
fine-tune from the previous checkpoint, is an overlooked decision
variable in active learning. We show that this choice has exploitable
structure: retraining is most useful in early rounds, when each batch
can substantially reshape the labeled distribution, while fine-tuning
becomes safer once the model trajectory stabilizes. We propose
\textsc{\textbf{HybridAL}}, an adaptive training schedule that monitors
an online stabilization signal and switches from retraining to
fine-tuning after sustained stabilization. Two complementary signals, spectral exponent change
$\Delta\alpha$ (weight-based) and accuracy change $\Delta$Acc
(validation-based), span different points on the
time--calibration trade-off. Across
three encoder backbones and six text-classification tasks
(five seeds each), \textsc{\textbf{HybridAL}} keeps endpoint
macro-\fone{} non-inferior to retraining and fine-tuning at a
$0.010$ margin, saves up to $49\%$ of retraining time, and recovers a
substantial fraction of retraining's calibration advantage as measured
by negative log-likelihood (NLL). Compared with schedules that switch
at a pre-committed round, \textsc{\textbf{HybridAL}} obtains lower
NLL at moderate additional cost, showing that trajectory-dependent
switching provides a stronger time--calibration trade-off than fixed
early switching\footnote{The implementation is available at:
\url{https://github.com/naghamo/hybridAL}}.
\end{abstract}
\section{Introduction}\label{sec:introduction}
\begin{figure}[h]
  \centering
  \includegraphics[width=0.5\textwidth]{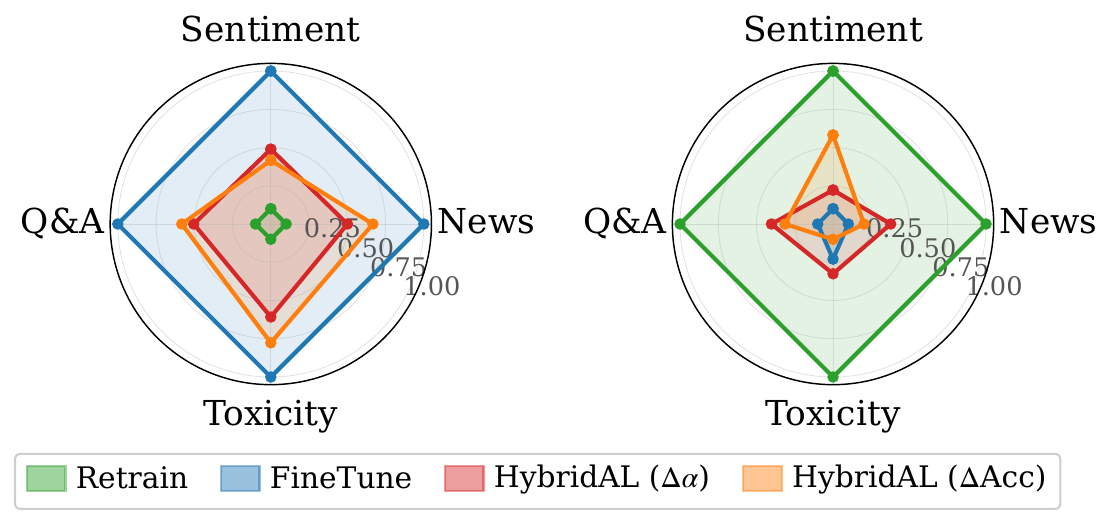}
  \caption{\textbf{No single training strategy dominates.}
  \textbf{Left:} speed (bigger = faster).
  \textbf{Right:} calibration (bigger = lower negative
  log-likelihood (NLL), i.e.\ better-calibrated predictions).
  Axes are min--max normalized per task-family dataset.
  \textsc{\textbf{HybridAL}} variants use $\Delta\alpha$ (spectral exponent
  change) and $\Delta$Acc (accuracy change) as switching
  signals.}
  \label{fig:radar}
\end{figure}


Active learning (AL) reduces annotation costs by iteratively
selecting informative unlabeled examples for labeling, rather than
annotating a large dataset in a single pass~\citep{settles2009active}.
Large language models (LLMs) have been proposed as scalable
annotators and judges across NLP and beyond~\citep{tan2024large, li2025generation}, and surpass crowd workers on some
annotation tasks~\citep{gilardi2023chatgpt}. However, their
annotations remain task-dependent and can exhibit systematic
biases~\citep{chen2024humans, ashktorab2025evalassist, calderon2025alternative, szymanski2025limitations}. These limitations mean that AL remains necessary for deciding
which examples to annotate under a limited
budget~\citep{ren2021survey}. At the same time, as LLM-based annotation reduces per-sample
cost, updating the model each round becomes the dominant
bottleneck in the AL loop, a cost that grows with each
acquisition step~\citep{scala2025efficient}; as practitioners
can afford more rounds, training-time savings become
increasingly valuable.

While much AL research optimizes which samples to acquire
~\citep{settles2009active, ash2019deep}, the choice of how to update
the model after each round remains underexplored~\citep{munagala2022clactive}. Two strategies
dominate practice: \textsc{Retrain}, which reinitializes from the original
pre-trained weights (or from random initialization when no
pre-trained backbone is used) and trains on all accumulated
labeled data, and
\textsc{FineTune}, which continues from the previous checkpoint.
Neither is uniformly preferable. \textsc{Retrain} is robust but
computationally redundant as the model matures, whereas
\textsc{FineTune} is efficient but can suffer from warm-starting
degradation in early, high-variance acquisition
rounds~\citep{ash2020warm}. This suggests a natural asymmetry:
early rounds benefit from retraining, while later rounds can often
be handled by fine-tuning once the model trajectory stabilizes.

We address this gap with \textsc{\textbf{HybridAL}}, an adaptive
training schedule for pool-based AL that switches from full
retraining to incremental fine-tuning after sustained stabilization.
The motivation is illustrated in Figure~\ref{fig:radar}: \textsc{Retrain} produces well-calibrated predictions, as reflected
by low test negative log-likelihood (NLL), but is slow; \textsc{FineTune} is efficient but incurs a calibration
penalty. This penalty matters because uncertainty-based
acquisition functions, the most widely used family in
AL~\citep{settles2009active, ren2021survey}, rank candidates by predicted
probabilities, so probability quality during training may affect
which examples are queried. We treat this as motivation for
retraining early, not as an established effect. No single strategy
dominates both speed and calibration. \textsc{\textbf{HybridAL}}
therefore monitors a switching signal after each round and switches
when the model trajectory enters a low-change regime.

A central question is which signal best detects stabilization. We
evaluate eight candidates spanning performance-based metrics
(e.g., accuracy
change) and model-based metrics
(e.g., spectral exponent change~\citep{martin2021implicit}
and representation similarity~\citep{kornblith2019similarity}).
Two signals emerge as complementary operating points: the
spectral exponent change ($\Delta\alpha$), a weight-based
signal that favors time savings, and the validation accuracy
change ($\Delta$Acc), which favors calibration. Both
maintain comparable final \fone{}.

\textbf{Our contributions} are threefold:
\begin{enumerate}[leftmargin=*,noitemsep,topsep=0pt]
    \item We identify training strategy as an overlooked decision
    variable in AL and propose
    \textsc{\textbf{HybridAL}} (Algorithm~\ref{alg:hybridal}), an
    adaptive schedule that switches from \textsc{Retrain} to
    \textsc{FineTune} once stabilization is detected.

    \item We introduce stabilization detection
    (Definition~\ref{def:stabilization}), a general online criterion
    over model-trajectory signals, and identify two complementary
    signals: $\Delta\alpha$ (weight-based, fastest, no additional validation pass) and $\Delta$Acc (validation-based, best calibration).

    \item We evaluate \textsc{\textbf{HybridAL}} across three
    encoder backbones and six text-classification tasks
    (5 seeds each) and show that endpoint \fone{} is
    non-inferior to both single-strategy baselines at a $0.010$ margin, roughly three quarters of the seed-to-seed standard
deviation (two one-sided tests, TOST), that it saves up to $49\%$ of
    retraining time, and that it achieves a stronger
    time--calibration trade-off than schedules that switch at a
    pre-committed round, establishing that adaptive timing, not
    switching itself, drives the calibration gain.
\end{enumerate}

\section{Related Work}

\medskip \noindent\textbf{Existing training regimes.} 
Retraining a model from scratch each round is often recommended for robust generalization, though it remains computationally expensive~\citep{beck2021effective}.
Conversely, fine-tuning is computationally efficient but frequently degrades generalization due to warm-start bias~\citep{ash2020warm}. While some configurations attempt to train exclusively on newly acquired data, this strategy risks catastrophic forgetting~\citep{munagala2022clactive, das2023accelerating} unless mitigated by replay-based methods that interleave a small buffer of previously labeled examples during updates~\citep{rolnick2019experience}. 

Despite these trade-offs, existing active learning pipelines apply a single training strategy uniformly across all selection rounds, ignoring a fundamental asymmetry: early rounds operate in a high-information, high-variance regime where each batch drastically reshapes the data distribution. Fine-tuning prematurely in this phase induces a severe loss of plasticity, permanently degrading the network's capacity to absorb new concepts~\citep{dohare2024loss}. Conversely, later rounds provide only marginal refinements to an already-stable model. Once a model's internal representations geometrically mature and stabilize, phenomena observable via spectral self-regularization~\citep{martin2021implicit} and neural collapse~\citep{papyan2020prevalence}, fine-tuning becomes safer and more efficient.

\medskip
\noindent\textbf{Adaptive methods and motivation for performance-based signals.}
Prior AL efficiency literature focuses mainly on other aspects of the pipeline. Recent advancements have introduced adaptive frameworks that dynamically switch between acquisition strategies mid-process, using multi-armed bandits \citep{zhang2023algorithm}, deep imitation learning \citep{liu2018learning}, or budget-aware heuristics that transition from typicality to uncertainty sampling as the labeled pool grows \citep{hacohen2022active, hacohen2023select}. Other works utilize dynamic performance signals to alter the AL pipeline mid-stream. For instance, performance plateaus and confidence metrics are frequently used as stopping criteria to terminate the AL loop~\citep{vlachos2008stopping, zhu2008multi}. Similarly, some work has used performance deltas as reward signals for reinforcement learning-based acquisition~\citep{fang2017learning}, and in stream-based AL, concept drift has been used to trigger model ensemble updates~\citep{han2024adaptive}. Current Green AI frameworks borrow AL-inspired iterative sampling and utilize adaptive performance signals, such as tracking loss stagnation to dynamically trigger shifts in the training regimen, to reduce computational costs on already fully labeled datasets \citep{scala2024play, scala2025efficient}. While the latter methods alter training to facilitate data pruning when all
labels are available, they do not inherently operate in an environment where labels are acquired iteratively. 

To our knowledge, prior pool-based AL work has not treated the
choice between \textsc{Retrain} and \textsc{FineTune} as an
online decision variable. \textsc{\textbf{HybridAL}} targets
this gap by adapting the training strategy while keeping the
acquisition protocol fixed.

\section{\textsc{\textbf{HybridAL}}: Adaptive Training Strategy Switching}
\label{sec:method}
We present \textsc{\textbf{HybridAL}}, an adaptive training
method for pool-based AL that switches from full
retraining to incremental fine-tuning by detecting when the
model has stabilized. We define the switching problem
(\S\ref{sec:problem}), introduce a stabilization detection
mechanism (\S\ref{subsec:stab}), and present the complete
algorithm (\S\ref{subsec:algorithm}).

\subsection{Model \& Problem Definition}
\label{sec:problem}

Let $\mathcal{D} = \{(x_i, y_i)\}_{i=1}^{N}$ denote a dataset over input space $\mathcal{X}$ and label space $\mathcal{Y} = \{1, \dots, C\}$ for $C$-class classification. We consider a standard pool-based AL setting~\citep{settles2009active} with initial labeled and unlabeled pools $\mathcal{L}_0$ and $\mathcal{U}_0$. AL proceeds for $T$ rounds under a fixed labeling budget $|\mathcal{L}_0| + nT$, where $n$ is the acquisition batch size. 
At each round $t$, a model $f_{\theta_t}: \mathcal{X} \to \mathcal{Y}$ with parameters $\theta_t$ is trained on $\mathcal{L}_{t-1}$ according to a strategy $s_t$. The model then selects a batch $\mathcal{Q}_t \subset \mathcal{U}_{t-1}$ of size $n$ via an acquisition function.
The pools are then updated as 
$\mathcal{L}_t = \mathcal{L}_{t-1} \cup \mathcal{Q}_t$ and 
$\mathcal{U}_t = \mathcal{U}_{t-1} \setminus \mathcal{Q}_t$.

In this work, we address the challenge of choosing $s_t$ as a
function of history up to round $t$. Standard paradigms
typically restrict $s_t$ to a constant strategy across all
rounds. \textsc{Retrain} reinitializes from pre-trained weights
and trains on all of $\mathcal{L}_{t-1}$, producing robust
generalization~\citep{ash2020warm, beck2021effective} at
growing cumulative cost. \textsc{FineTune} continues from
$\theta_{t-1}$, reducing per-round cost through warm-starting,
but inheriting biases from previous checkpoints that can
degrade generalization~\citep{ash2020warm}, particularly in
early rounds~\citep{beck2021effective}. A third approach (which
we do not consider in our solution but include as a baseline)
is \textsc{NewOnly}, which trains only on the newly acquired
batch, discarding historical data and risking catastrophic
forgetting~\citep{munagala2022clactive, das2023accelerating}.

To combine the early-stage robustness of \textsc{Retrain} with
the late-stage efficiency of
\textsc{FineTune}, we
consider schedules that switch once from \textsc{Retrain} to
\textsc{FineTune}. We first state the switching objective as
an offline problem, then explain why it must be approximated
online.

\begin{problem}[Training Strategy Switching]\label{prob:offline Switch}
Find a switching point $t^* \in \{1,\dots,T+1\}$ that defines
\begin{equation}
s_\tau(t^*) =
\begin{cases}
\textsc{Retrain} & \text{if } \tau < t^*,\\
\textsc{FineTune} & \text{if } \tau \geq t^*.
\end{cases}
\end{equation}
Here $t^*=T+1$ recovers pure \textsc{Retrain}, and $t^*=1$
recovers pure \textsc{FineTune}. The ideal switch improves the
time--calibration trade-off while preserving endpoint
classification performance:
\begin{equation}
\min_{t \in \{1,\dots,T+1\}}
\ \mathrm{Time}(t) + \lambda\,
\mathrm{Calib}(f_{\theta_T}^{(t)})
\end{equation}
subject to
\begin{equation}
\mathrm{Perf}(f_{\theta_T}^{(t)})
\geq
\max\!\left(
\mathrm{Perf}_{\textsc{Retrain}},
\mathrm{Perf}_{\textsc{FineTune}}
\right)-\delta,
\end{equation}
where $f_{\theta_T}^{(t)}$ is the final model obtained by
switching at round $t$, $\mathrm{Time}(t)$ is the cumulative training time across all
$T$ rounds under switch point $t$, $\mathrm{Calib}(\cdot)$ is a calibration error
measure, $\mathrm{Perf}(\cdot)$ is a task performance metric
(e.g., macro-\fone{}), $\lambda \geq 0$ controls the
time--calibration trade-off, and $\delta \geq 0$ is an allowed
performance tolerance.
\end{problem}

Problem~\ref{prob:offline Switch} depends on endpoint quantities
that are known only in post-hoc analysis. Evaluating
$\mathrm{Perf}(f_{\theta_T}^{(t)})$,
$\mathrm{Calib}(f_{\theta_T}^{(t)})$, or even
$\mathrm{Time}(t)$ for a candidate switch point $t$ would
require running the full $T$-round AL loop under that choice.
\textsc{\textbf{HybridAL}} therefore approximates this
objective online using the stabilization criterion in
Definition~\ref{def:stabilization} as a tractable proxy.

\subsection{Stabilization Detection as a Switching Signal}
\label{subsec:stab}

Early rounds operate with small pools where
each batch constitutes a $n / |\mathcal{L}_t|$ distributional
shift; in this regime, warm-starting degrades
generalization~\citep{ash2020warm}, and the penalty compounds
across rounds. As the pool grows,
the per-round shift shrinks, checkpoint quality improves, and
the gap between strategies
vanishes. This asymmetry motivates
switching from \textsc{Retrain} to \textsc{FineTune}, rather
than the reverse. The remaining question is \emph{when}.
\paragraph{The stabilization hypothesis.}
AL exhibits a regime transition: learning dynamics shift from
rapid exploration (high information gain, large distributional
shifts, unstable representations) to gradual refinement
(diminishing returns, converged representations). The
transition point varies by task and dataset complexity, so a
fixed switching round cannot suit all settings. We formalize
the detection of this transition as follows.

\begin{definition}[Stabilization Point]
\label{def:stabilization}
Let $S(f_{\theta_t})$ denote a switching signal evaluated after 
round $t$. The signal change is 
$\Delta S_t = |S(f_{\theta_t}) - S(f_{\theta_{t-1}})|$ for 
$t \geq 1$. The stabilization point is the earliest round $t^*$ 
at which $\Delta S_t$ remains below threshold $\varepsilon$ for 
$k$ consecutive rounds:
\begin{equation}
t^* = \min \left\{ t \geq k \;\middle|\; 
  \max_{i \in [t-k+1,\, t]} \Delta S_i < \varepsilon \right\}
\end{equation}
where $\varepsilon > 0$ is the sensitivity threshold and 
$k \geq 1$ the patience parameter.
\end{definition}

The threshold $\varepsilon$ controls how much round-to-round
change is tolerated before declaring stabilization: smaller
values require the signal to flatten more before switching.

When $\Delta S_t$ is large, the model trajectory is still changing
substantially and retraining remains safer. When $\Delta S_t$ stays
below $\varepsilon$, the trajectory has entered a low-change regime:
the warm-starting penalty is less likely to dominate, and fine-tuning
becomes a more efficient update. Crucially, $\Delta S_t$ is a relative
measure of change, not an absolute performance level, making it less sensitive to task difficulty.

\paragraph{The role of patience.}
A single low $\Delta S_t$ may result from noise: an uninformative 
batch, a temporary plateau, or class sampling 
imbalance~\citep{ren2021survey}. The patience parameter $k$ requires $k$ 
consecutive sub-threshold rounds before switching, filtering 
transient fluctuations. This corresponds exactly to the 
$\max_{i \in [t-k+1,\, t]} \Delta S_i < \varepsilon$ condition: 
a single excursion above $\varepsilon$ within the window resets 
the counter. \textsc{\textbf{HybridAL}} thus monitors the signal online and switches 
only when stabilization is confirmed, adapting to each task's 
trajectory.

\subsection{The \textsc{\textbf{HybridAL}} Algorithm}
\label{subsec:algorithm}

Algorithm~\ref{alg:hybridal} presents the complete procedure with 
three state variables: current strategy $s$, stabilization counter 
$\text{stable\_count}$, and previous signal value $S_{\text{prev}}$.

\begin{algorithm}[t]
\caption{\textsc{\textbf{HybridAL}}: Adaptive Training Strategy Switching}
\label{alg:hybridal}
\begin{algorithmic}[1]
\REQUIRE Unlabeled pool $\mathcal{U}_0$, labeled pool $\mathcal{L}_0$, validation set $\mathcal{V}$, rounds $T$, batch size $n$, acquisition function $A$, threshold $\varepsilon$, patience $k$
\ENSURE Final model $f_{\theta_T}$
\STATE $s \gets \textsc{Retrain}$;\; $\text{stable\_count} \gets 0$;\; $S_{\text{prev}} \gets 0$
\FOR{$t = 1$ to $T$}
    \IF{$s = \textsc{Retrain}$}
        \STATE $f_{\theta_t} \gets \text{TrainFromScratch}(\mathcal{L}_{t-1})$
    \ELSE
        \STATE $f_{\theta_t} \gets \text{FineTune}(f_{\theta_{t-1}}, \mathcal{L}_{t-1})$
    \ENDIF
    \STATE $\mathcal{Q}_t \gets A(f_{\theta_t}, \mathcal{U}_{t-1}, n)$
    \STATE $\mathcal{L}_t \gets \mathcal{L}_{t-1} \cup \mathcal{Q}_t$;\; $\mathcal{U}_t \gets \mathcal{U}_{t-1} \setminus \mathcal{Q}_t$
    \STATE $S_{\text{curr}} \gets S(f_{\theta_t})$
    \STATE $\Delta S_t \gets |S_{\text{curr}} - S_{\text{prev}}|$;\; $S_{\text{prev}} \gets S_{\text{curr}}$
    \IF{$\Delta S_t < \varepsilon$}
        \STATE $\text{stable\_count} \gets \text{stable\_count} + 1$
        \IF{$\text{stable\_count} \geq k$ \AND $s = \textsc{Retrain}$}
            \STATE $s \gets \textsc{FineTune}$ \COMMENT{Permanent switch}
        \ENDIF
    \ELSE
        \STATE $\text{stable\_count} \gets 0$
    \ENDIF
\ENDFOR
\RETURN $f_{\theta_T}$
\end{algorithmic}
\end{algorithm}

At each round, \textsc{\textbf{HybridAL}} trains $f_{\theta_t}$ according to the 
current strategy (Lines 3--6), selects and labels a batch 
(Lines 7--8), and computes the switching signal and its change 
(Lines 9--10). Performance-based signals ({\em e.g.}, $\Delta$Acc) require a forward pass on $\mathcal{V}$. Model-based 
signals ({\em e.g.}, $\Delta\alpha$) are computed directly from the 
weights. The algorithm permanently switches to \textsc{FineTune} after $k$ consecutive rounds with $\Delta S_t < \varepsilon$
(Line 14).

\paragraph{Key properties.}
The switch is \emph{irreversible}: once $s_t = \textsc{FineTune}$,
the algorithm never reverts, ensuring monotonically decreasing
per-round cost. This design is deliberate: after switching,
the model is updated via warm-starting rather than
re-initialization, which changes the optimization dynamics.
Under these new dynamics, signal values can fluctuate even
when the model remains performant. A reversible variant would
misinterpret such fluctuations as instability and repeatedly
revert to \textsc{Retrain}, losing the cost guarantee without
improving endpoint performance
(Appendix~\ref{app:post_switch}).

Stabilization of classification decisions does not guarantee
stabilization of the full probability distribution;
\textsc{\textbf{HybridAL}} mitigates calibration drift by
retraining during the early rounds, when probability estimates
are most sensitive to the training data composition. This
motivates the design, since uncertainty-based acquisition ranks
candidates by predicted probabilities and post-hoc
recalibration~\citep{guo2017calibration} applies only to the
final model, leaving acquisition decisions already made during
training unchanged (Appendix~\ref{app:temp_scaling}). Empirically,
acquisition quality is maintained after the switch
(Appendix~\ref{app:batch_composition}); whether improved calibration
yields better acquisition utility remains open.

The time savings arise because \textsc{FineTune} converges in
fewer epochs than \textsc{Retrain} under early stopping. The
validation set $\mathcal{V}$ is held fixed, shared by all
strategies, and does not consume labeling budget. \textsc{\textbf{HybridAL}} introduces two
hyperparameters: $\varepsilon$ and $k$, whose selection is
described in \S\ref{sec:experiments}.
\section{Experiments}\label{sec:experiments}\label{sec:results}

We evaluate \textsc{\textbf{HybridAL}} against single-strategy
baselines and non-adaptive schedules that switch at a fixed
pre-committed round, across three encoder backbones and six
text-classification datasets spanning binary and multi-class
regimes, with five seeds per cell. After describing the protocol
(\S\ref{sec:setup}), the remainder of this section establishes
three claims:

\begin{itemize}
\item \textsc{\textbf{HybridAL}} is non-inferior to both
single-strategy baselines across backbones and task
difficulties (\S\ref{sec:f1_parity}).
\item \textsc{\textbf{HybridAL}} retains most of
\textsc{Retrain}'s calibration while capturing the bulk of
\textsc{FineTune}'s training-time savings
(\S\ref{sec:time_nll}).
\item \textsc{\textbf{HybridAL}} achieves a stronger
time--calibration trade-off than fixed early-switch schedules
by adapting the switch point to the model trajectory
(\S\ref{sec:adaptivity}).
\end{itemize}

\subsection{Experimental Setup}\label{sec:setup}

\paragraph{Datasets.}
We use six English text-classification benchmarks
(Table~\ref{tab:datasets}): three binary (IMDb, Jigsaw, SST-2)
and three multi-class (TweetEval, AG News, Yahoo Answers).
Yahoo Answers is stratified-downsampled to $60$k (uniform
across classes). Per-split sizes are in
Appendix~\ref{app:datasets}.

\begin{table}[H]
\centering
\resizebox{\columnwidth}{!}{%
\begin{tabular}{l|p{3cm}|c|c}
\toprule
\textbf{Dataset} & \textbf{Domain} & \textbf{C} & \textbf{Size} \\
\midrule
\multicolumn{4}{l}{\textit{Binary}} \\
IMDb \citep{maas2011learning} & sentiment analysis & 2 & 50,000 \\
SST-2 \citep{socher2013recursive} & sentiment analysis & 2 & 68,221 \\
Jigsaw \citep{wulczyn2017ex} & toxicity detection & 2 & 159,571 \\
\midrule
\multicolumn{4}{l}{\textit{Multi-class}} \\
TweetEval \citep{barbieri2020tweeteval} & sentiment analysis & 3 & 59,899 \\
AG News \citep{zhang2015character} & topic classification & 4 & 127{,}600 \\
Yahoo \citep{zhang2015character} & topic (Q\&A) & 10 & 60{,}000  \\
\bottomrule
\end{tabular}
}
\caption{Text classification datasets, showing total size
and number of classes (C).}
\label{tab:datasets}
\end{table}

\paragraph{Models.}
We evaluate DistilBERT~\citep{sanh2019distilbert} ($\sim$66M),
BERT-base~\citep{devlin2019bert} ($\sim$110M), and
RoBERTa-base~\citep{liu2019roberta} ($\sim$125M). DistilBERT is
the default for ablations; all three appear in the main results.

\paragraph{Active learning protocol.}
Each run starts from a class-stratified pool of
$|\mathcal{L}_0|{=}200$ and proceeds for $T{=}25$ rounds,
acquiring $n{=}32$ examples per round (final budget
$1{,}000$). A stratified per-dataset validation set $\mathcal{V}$
($477$ to $1{,}596$ labels) is held fixed across rounds, separate
from the labeling budget and AL pools, and is shared by all methods
for early stopping and per-round evaluation. Validation-label
assumptions and a size-sensitivity study are in
Appendix~\ref{app:val_size}. Each configuration is repeated over 5 seeds
($42$--$46$). Sensitivity to $|\mathcal{L}_0|$, $n$, and
the entropy pre-filter size $N{=}1{,}000$ is in
Appendix~\ref{app:robustness}.

\paragraph{Acquisition functions.}
\textsc{Entropy}~\citep{settles2009active} is the default;
ablations with \textsc{Random} and
\textsc{BADGE}~\citep{ash2019deep} are in
Appendix~\ref{app:sampler_ablation}.

\paragraph{Switching signals.}
We evaluate eight signals capturing round-to-round model change
(Table~\ref{tab:signal_summary}): four performance-based,
measured on $\mathcal{V}$: macro-F1 change $\Delta$F1,
accuracy change $\Delta$Acc, cross-entropy change $\Delta$Loss,
and gradient $\ell_2$ norm; and four model-based: spectral
exponent change
$\Delta\alpha$~\citep{martin2021implicit}, the mean
power-law tail exponent of each layer's eigenvalue spectrum
differenced between rounds; $\ell_2$ weight distance
$\|\theta_t{-}\theta_{t-1}\|_2$; representational similarity
$1{-}\mathrm{CKA}$~\citep{kornblith2019similarity}; and
within-class feature concentration change
$\Delta$NC~\citep{papyan2020prevalence}. Of the model-based
signals, $\Delta\alpha$ and weight distance operate on weight
matrices alone; CKA and $\Delta$NC require a forward pass on
$\mathcal{V}$ to extract representations. All strategies use
$\mathcal{V}$ for early stopping and per-round evaluation
regardless of signal choice; the distinction is whether the
signal adds an extra pass each round. Based on a preliminary
signal comparison on DistilBERT
(Table~\ref{tab:signal_summary}), the main results use
$\Delta\alpha$ and $\Delta$Acc: they have the two highest
fire rates ($97\%$ and $93\%$), each is Pareto-optimal on
(F1, time) within its family (model-based and
performance-based, respectively), and the two are mutually
uncorrelated ($\rho \approx 0$), capturing complementary
information. Signal normalization and full per-signal results
are in Appendix~\ref{app:signal_ablation}.
\begin{table}[H]
  \centering
  \small
  \setlength{\tabcolsep}{2pt}
  \begin{tabular}{llccc}
  \toprule
  Signal & Type & Extra val.\ pass? & Fire rate & Mean $t^{\star}$ \\
  \midrule
  $\Delta$Acc$^{\star}$  & performance & yes &  93\% &  9.1 \\
  $\Delta$F1             & performance & yes &  90\% & 10.0 \\
  $\Delta$Loss           & performance & yes &  43\% & 13.4 \\
  Grad.\ norm            & performance & yes &  40\% &  8.5 \\
  \midrule
  $\Delta\alpha^{\star}$ & model-based & no  &  97\% &  7.9 \\
   $\ell_2$ distance            & model-based & no  &   0\% &  \ding{55}\\
  $\Delta$NC             & model-based & yes &  67\% & 12.4 \\
  CKA                    & model-based & yes &  33\% &  5.4 \\
  \bottomrule
  \end{tabular}
  \caption{Signal ablation (DistilBERT, 6 datasets $\times$
  5 seeds, normalised $\varepsilon{=}0.5$, $k{=}3$). Fire
  rate: fraction that switched; mean $t^{\star}$: switch
  round among firing cells. $^{\star}$Selected for main
  results.}
  \label{tab:signal_summary}
\end{table}

\paragraph{Methods.}
Three single-strategy baselines: \textsc{Retrain},
\textsc{FineTune}, and \textsc{NewOnly} (trains only on the new
batch). Two \textsc{\textbf{HybridAL}} variants: $\Delta\alpha$
and $\Delta$Acc (Algorithm~\ref{alg:hybridal}). Four
non-adaptive ablations: \textsc{FixedSwitch@$k$} for
$k\in\{3,5,7,10\}$, spanning the range around the mean switch
rounds of the two selected signals
(Table~\ref{tab:signal_summary}), which switch unconditionally
at a pre-committed round, isolating whether the gain comes
from switching itself or from adaptive timing.

\paragraph{Hyperparameters.}
We tune $(\varepsilon, k)$ on IMDb and AG News (one binary,
one multi-class; 3 seeds, 15 rounds) by selecting the cell
within $0.5\%$ of top validation \fone{} that minimises a
normalised time--NLL score
(Appendix~\ref{app:hyperparam_tuning}):
$(\varepsilon^*, k^*) = (10^{-4}, 3)$ for $\Delta\alpha$ and
$(5{\times}10^{-3}, 2)$ for $\Delta$Acc, applied without
retuning. The two-order-of-magnitude $\varepsilon$ gap
reflects different signal units, not sensitivity (\fone{}
varies ${\leq}1.1$~percentage points (pp) across the grid).

\paragraph{Training and evaluation.}
All strategies use AdamW~\citep{loshchilov2017decoupled}
(lr$=2{\times}10^{-5}$, weight decay$=10^{-3}$), batch size
$16$, up to $10$ epochs with early stopping (patience $2$).
\textsc{FineTune} converges in $3.4$ epochs on average vs.\
$5.5$ for \textsc{Retrain}
(Appendix~\ref{app:early_stopping}). We report macro-F1, test NLL, and wall-clock time; for
\textsc{\textbf{HybridAL}} variants we additionally report
the mean switch round $t^{\star}$ and the switch rate.
Significance is assessed via paired two-sided $t$-test at
$\alpha = 0.05$; non-inferiority is assessed by TOST. Experiments ran on two NVIDIA RTX 2080 Ti
GPUs with PyTorch~2.6~\citep{paszke2019pytorch} and
HuggingFace Transformers~\citep{wolf2020transformers}. Full
pairwise results are in Appendix~\ref{app:main}.

\begin{figure*}[t]
  \centering
  \begin{subfigure}[b]{0.48\textwidth}
    \includegraphics[width=\linewidth]{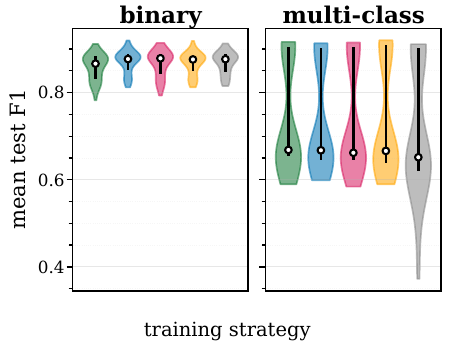}
    \caption{Endpoint test \fone{} by task family.}
    \label{fig:f1_violins}
  \end{subfigure}\hfill
  \begin{subfigure}[b]{0.48\textwidth}
    \includegraphics[width=\linewidth]{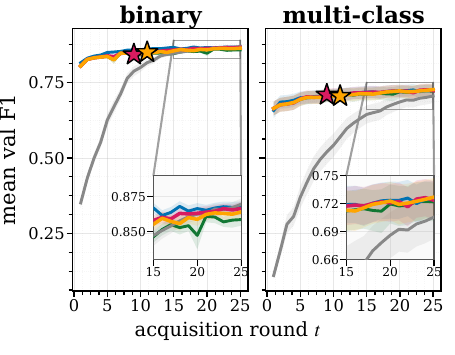}
    \caption{Per-round validation \fone{} by task family.}
    \label{fig:f1_trajectory}
  \end{subfigure}

  \vspace{4pt}
  \centerline{\includegraphics[width=0.70\textwidth]{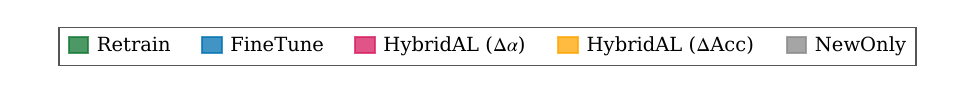}}

\caption{Test \fone{} preservation.
\textbf{(a)} Endpoint test \fone{} distributions per task family
(45 cells: 3 backbones $\times$ 3 datasets $\times$ 5 seeds);
median (white dot) and IQR (black bar) overlaid.
\textbf{(b)} Per-round validation \fone{}; $\pm 1$~SEM bands;
stars mark each \textsc{\textbf{HybridAL}} variant's mean switch round
$t^{\star}$; insets zoom the convergence region
(rounds~15-25). Method colors follow the shared legend below.}
\label{fig:f1}
\end{figure*}

\subsection{Preserving F1}
\label{sec:f1_parity}

Figure~\ref{fig:f1_violins} shows endpoint test \fone{}
distributions per task family. Both \textsc{\textbf{HybridAL}}
variants remain close to \textsc{Retrain} and \textsc{FineTune} on
every backbone: pooled across the six datasets, their means lie within
$0.5$--$0.9$ pp of one another. We test this formally with TOST on the paired differences over all
$90$ (backbone, dataset, seed) cells, anchoring the margin to
\textsc{Retrain}'s mean seed-to-seed \fone{} standard deviation
of $0.0131$. At $\delta{=}0.010$ both variants are non-inferior
to \textsc{Retrain} and \textsc{FineTune} individually and to the
per-cell better of the two; at $\delta{=}0.005$ three of the four
hybrid--baseline pairs pass, the exception being $\Delta$Acc vs.\
\textsc{FineTune} ($p{=}0.051$). Per-cell tests and full bounds
are in Appendix~\ref{app:tost}. Thus, the main
effect of the training schedule is not endpoint \fone{}, but the
time--calibration trade-off analyzed below. Pooled across all six datasets, \textsc{NewOnly}'s mean
trails by $\approx 1.7$~pp on DistilBERT and $2.7$~pp on
BERT relative to the strongest non-\textsc{NewOnly} method,
with the deficit statistically significant on Yahoo Answers
across all backbones (paired $t$-test, $p<0.05$;
Appendix~\ref{app:f1}). The gap concentrates on the hardest
multi-class tasks relative to \textsc{Retrain}: TweetEval
($-8.2$~pp DistilBERT, $-5.8$~pp BERT) and Yahoo Answers
($-2.9$~pp DistilBERT, $-5.9$~pp BERT); on RoBERTa it
vanishes (${\leq}0.2$~pp pooled).

Figure~\ref{fig:f1_trajectory} confirms these conclusions hold
throughout training. Round-by-round, the mean validation \fone{}
of \textsc{Retrain}, \textsc{FineTune}, \textsc{\textbf{HybridAL}}($\Delta\alpha$), and
\textsc{\textbf{HybridAL}}($\Delta$Acc) is nearly identical from the first
acquisition round onward; the insets (rounds~15--25) show the
four curves stay within $\approx 1$~pp of each other at
convergence. The two \textsc{\textbf{HybridAL}} variants therefore track the
single-strategy baselines at every round, and \fone{} shows no
inflection at \textsc{\textbf{HybridAL}}'s mean switch round $t^{\star}$ (stars),
so the \textsc{Retrain}$\to$\textsc{FineTune} handoff does not disrupt learning.
\textsc{NewOnly}, by contrast, converges visibly slower on both
task families and especially on multi-class, and even at round
$T$ remains below the other four methods.

\begin{figure*}[t]
  \centering
  \begin{subfigure}[b]{0.48\textwidth}
    \includegraphics[width=\linewidth]{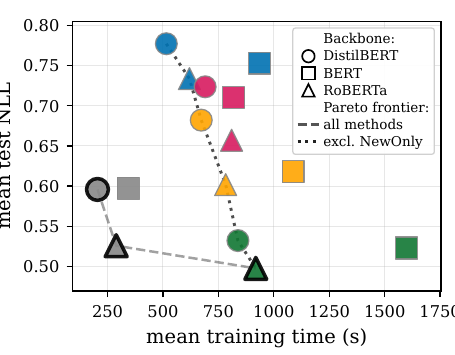}
    \caption{Time--NLL Pareto plane.}
    \label{fig:pareto_nll_time}
  \end{subfigure}\hfill
  \begin{subfigure}[b]{0.48\textwidth}
    \includegraphics[width=\linewidth]{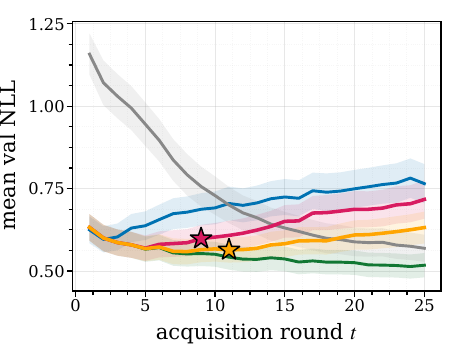}
    \caption{Per-round validation NLL.}
    \label{fig:nll_trajectory}
  \end{subfigure}

  \vspace{4pt}
  \centerline{\includegraphics[width=0.70\textwidth]{figures/main_results_f1_legend.pdf}}

 \caption{Time--calibration trade-off.
\textbf{(a)} Mean training time vs.\ mean test NLL per
(method, backbone) (color = method, shape = backbone); dashed
line is the all-method Pareto frontier, dotted excludes
\textsc{NewOnly}.
\textbf{(b)} Per-round mean validation NLL pooled across
6 datasets $\times$ 3 backbones $\times$ 5 seeds; $\pm 1$~SEM
bands; stars mark each HybridAL variant's mean switch round
$t^{\star}$. Method colors follow the shared legend below.}

\end{figure*}
\subsection{\textsc{\textbf{HybridAL}} Trades Time for Calibration}
\label{sec:time_nll}

Preserving \fone{} alone is insufficient: the entropy sampler
ranks candidates by predicted probability, so calibration during
the loop is what those decisions rest on. We therefore
turn to the time--calibration trade-off, measured through NLL,
which has a clear structure
(Figure~\ref{fig:pareto_nll_time}). \textsc{Retrain} achieves the lowest test NLL on every
backbone ($0.498$ RoBERTa to $0.532$ DistilBERT), but is the
slowest method ($838$--$1{,}599$~s). \textsc{FineTune} is
$33$--$41$\% faster, but incurs $44$--$47$\% higher NLL. On
DistilBERT and RoBERTa, both \textsc{\textbf{HybridAL}}
variants lie between these extremes. On BERT,
\textsc{\textbf{HybridAL}}($\Delta\alpha$) Pareto-dominates
\textsc{FineTune}: it is faster ($819$~s vs.\ $936$~s) and
better calibrated (NLL $0.710$ vs.\ $0.753$). Across all
backbones, \textsc{\textbf{HybridAL}}($\Delta$Acc) saves
$15$--$32$\% of \textsc{Retrain}'s time at only
$18$--$28$\% higher NLL, reclaiming $39$--$59$\% of
\textsc{FineTune}'s raw NLL gap;
\textsc{\textbf{HybridAL}}($\Delta\alpha$) saves
$12$--$49$\% of \textsc{Retrain}'s time at $32$--$36$\%
higher NLL. Thus, $\Delta$Acc is the safer default for calibration,
while $\Delta\alpha$ favors speed. A temperature scaling
analysis confirms that this ordering reflects training-time overconfidence inherited from warm-starting
(Appendix~\ref{app:temp_scaling}). Both variants lie on the
substantive Pareto frontier, excluding \textsc{NewOnly}, a
region not reached by any single-strategy pool-trained method.

Figure~\ref{fig:nll_trajectory} shows that this ordering is
stable throughout training: from approximately round $5$ onward,
validation NLL follows \textsc{Retrain} $<$
\textsc{\textbf{HybridAL}}($\Delta$Acc) $<$
\textsc{\textbf{HybridAL}}($\Delta\alpha$) $<$
\textsc{FineTune}, with \textsc{FineTune} climbing late.
\textsc{NewOnly} is the cheapest method and appears on the
all-methods Pareto frontier, but it is not directly comparable
to the pool-trained strategies: it discards the accumulated
labeled set and trains only on the newest batch. This makes its
endpoint NLL misleading for AL, since it begins as the
worst-NLL method and only catches up late, so early acquisition
decisions are made from poorly calibrated predictions. Together
with its \fone{} deficit on hard multi-class tasks with smaller
backbones (\S\ref{sec:f1_parity}), this is why we exclude
\textsc{NewOnly} from the substantive frontier.

\begin{figure}[t]
  \centering
  \includegraphics[width=\columnwidth]{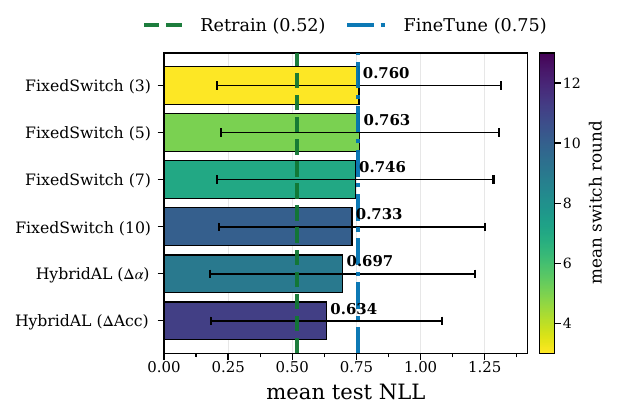}
  \caption{Mean test NLL. Bar color encodes mean switch round.
\textsc{Retrain} (dashed) and \textsc{FineTune} (dash-dot)
mark the calibration ceiling and floor.}
  \label{fig:switch_tradeoff}
\end{figure}

\subsection{Adaptive Switching}
\label{sec:adaptivity}

A natural question is whether \textsc{\textbf{HybridAL}}'s
gains require adaptive timing or simply result from switching
at any round. We compare against \textsc{FixedSwitch}
schedules that switch unconditionally at round
$k\in\{3,5,7,10\}$. Pooled mean \fone{} agrees within $0.010$ on
every backbone across all methods (Appendix~\ref{app:f1}); the
difference lies in calibration. \textsc{FixedSwitch} commits
to an early switch and is faster, but
Figure~\ref{fig:switch_tradeoff} shows its NLL clusters near
\textsc{FineTune}'s ($\approx 0.75$), while both
\textsc{\textbf{HybridAL}} variants move closer to
\textsc{Retrain}'s ($\approx 0.52$), with
\textsc{\textbf{HybridAL}}($\Delta$Acc) the lowest
non-\textsc{Retrain} method.
\textsc{\textbf{HybridAL}}'s empirical switch lands later
(mean $t^{\star}\!\approx\!9$--$12$; range $3$--$25$ across
cells), using the extra retraining rounds to improve
calibration. The switch also varies by dataset (e.g.,
$t^{\star}\!\approx\!6$ on TweetEval vs.\ $11$ on Yahoo
Answers for $\Delta\alpha$), confirming that no fixed schedule reproduces this
adaptation. The stabilization criterion
(Definition~\ref{def:stabilization}) thus adapts to each
task without per-dataset tuning, achieving a stronger
time--calibration trade-off than any pre-committed schedule.
Full per-dataset results are in Appendix~\ref{app:main}.

\section{Conclusion}
\label{sec:conclusion}

Training strategy is a decision variable that AL
pipelines often leave fixed. We introduced
\textsc{\textbf{HybridAL}}, an adaptive schedule that uses
retraining while the model trajectory is changing rapidly and
switches to fine-tuning after sustained stabilization. Across 9
methods, 3 encoder backbones, 6 text-classification benchmarks,
and 5 seeds, \textsc{\textbf{HybridAL}} keeps endpoint test
\fone{} non-inferior to the pool-trained baselines at a $0.010$
margin, substantially reduces wall-clock training time relative to full retraining, and recovers much of \textsc{Retrain}'s NLL-based calibration
advantage. Compared with fixed early-switch schedules,
\textsc{\textbf{HybridAL}} obtains a stronger time--calibration
trade-off by adapting the switch point to the model trajectory,
consistent with prior evidence that warm-starting costs are
concentrated in early rounds~\citep{ash2020warm, beck2021effective}.
The two \textsc{\textbf{HybridAL}} variants offer complementary
operating points: \textsc{\textbf{HybridAL}}($\Delta$Acc) favors
lower NLL when a validation set is available, while
\textsc{\textbf{HybridAL}}($\Delta\alpha$) favors speed and uses
weight statistics without an additional validation forward pass. Both share the
same switching mechanism, so practitioners can choose the signal
according to the desired time--calibration trade-off.

Future work could switch to a
damped per-round controller that selects the training strategy
at each acquisition step while avoiding the oscillation of
na\"ive reversibility
(Appendix~\ref{app:post_switch}). Replay-based
extensions~\citep{rolnick2019experience, das2023accelerating}
could also combine naturally with
\textsc{\textbf{HybridAL}}'s switching logic, especially for
stronger backbones where \textsc{NewOnly}'s \fone{} deficit
already narrows.

\section*{Limitations}
\label{sec:limitations}

\paragraph{Irreversible switching.}
\textsc{\textbf{HybridAL}}'s switch is irreversible: once it moves to \textsc{FineTune}, it cannot revert to \textsc{Retrain} even if the model later exhibits large
round-over-round drift (e.g., from a batch introducing a
previously rare class). The irreversibility guarantees
monotonically decreasing per-round cost but bounds the calibration
we can recover. Empirically, post-switch signal values frequently
re-cross $\varepsilon$ on every dataset
(Appendix~\ref{app:post_switch}), so a reversible variant
would oscillate between strategies rather than settle. The
irreversible design prevents this instability, but it means
\textsc{\textbf{HybridAL}} cannot recover if a genuinely novel regime emerges late
in training. Designing a damped or hysteresis-based reversal
mechanism that avoids oscillation remains future work.

\paragraph{Residual calibration drift.}
\textsc{\textbf{HybridAL}} leaves residual calibration drift relative to \textsc{Retrain}.
Even \textsc{\textbf{HybridAL}}($\Delta$Acc) remains $18$--$28\%$ above \textsc{Retrain}'s
NLL on every backbone (\S\ref{sec:time_nll}); applications that
demand strict endpoint probability calibration (e.g., selective
prediction with hard thresholds) may still require full retraining
each round, potentially combined with post-hoc
recalibration~\citep{guo2017calibration}.
However, for the common AL setting where calibration quality
during acquisition matters most, \textsc{\textbf{HybridAL}} retains
more of \textsc{Retrain}'s calibration through the early rounds
(\S\ref{sec:time_nll}).

\paragraph{Acquisition-utility evidence is indirect.}
We do not isolate the effect of calibration quality on acquisition
utility. After the switch, the examples \textsc{\textbf{HybridAL}}
acquires overlap little with those the \textsc{Retrain} run
selects, yet class balance and endpoint \fone{} are maintained
(Appendix~\ref{app:batch_composition}); this shows acquisition
quality is not degraded, not that better calibration improves it.
Isolating the effect would require refitting a temperature
$\tau_t$ before acquisition at every round and comparing against
native probabilities, which we leave to future work
(Appendix~\ref{app:temp_scaling}).

\paragraph{Structural lower bound of pool-based AL.}
The same lower bound surfaces in \textsc{NewOnly}, which trains
only on the newly acquired batch and is the cheapest method we
test. \textsc{NewOnly} is well-calibrated at convergence but
begins AL as the worst-calibrated method
(\S\ref{sec:time_nll}). This is a structural property of
pool-based AL: cumulative-data training is what supplies
early-round calibration, and no schedule that omits it can
match \textsc{Retrain} in the first several rounds. \textsc{\textbf{HybridAL}} stays in
\textsc{Retrain} until stabilization, but it cannot remove the
underlying constraint.

\paragraph{No formal stabilization guarantee.}
The stabilization detection in
Definition~\ref{def:stabilization} is empirical: we have no
formal guarantee on when the switch fires for an unseen
dataset, nor a closed-form bound on the calibration loss it
incurs. The hyperparameters $(\varepsilon, k)$ were tuned on a
two-dataset subset (\S\ref{sec:setup}) and applied without
retuning to every main-results experiment; a task whose
stabilization profile differs substantially from our six
benchmarks may need fresh tuning.

\paragraph{Generalization beyond encoders.}
Our results cover encoder-based text classification with three
backbones under $150$M parameters. Two aspects of
\textsc{\textbf{HybridAL}} are tied to that setting. First,
$\Delta\alpha$ is a spectral statistic of the weight matrices, so
its scale depends on architecture and depth; the tuned
$(\varepsilon, k)$ would not carry over to models with different
spectra, and \S\ref{sec:setup}'s thresholds would need refitting.
Second, the time savings come from \textsc{FineTune} converging in
fewer epochs under early stopping, which assumes full-parameter
updates; under parameter-efficient tuning the per-round cost gap
that \textsc{\textbf{HybridAL}} exploits is much smaller. Whether the stabilization transition itself appears in
decoder-based models, substantially larger backbones, or tasks
beyond classification remains for future work.

\section*{Ethical Considerations}

\paragraph{Data.} All six datasets are public research benchmarks
used for their original classification tasks; we redistribute
neither data nor annotations. The Jigsaw corpus
\citep{wulczyn2017ex} contains toxic user comments, and automated
toxicity classifiers are known to exhibit demographic biases
\citep{sap2019risk}; \textsc{\textbf{HybridAL}} is a training schedule and does not
mitigate them.

\paragraph{Models.} We build on BERT, DistilBERT, and RoBERTa,
which inherit biases from web-scale pretraining
\citep{bender2021dangers}. Our results characterize average
behavior across seeds and datasets, not worst-case behavior on
specific subpopulations.

\paragraph{Compute.} The full set of experiments (1{,}626 runs
across main results, ablations, and sensitivity studies) took
approximately $280$ GPU-hours on two RTX 2080 Ti cards. \textsc{\textbf{HybridAL}}
itself reduces per-cycle training time by $12$--$49\%$ relative
to full retraining (\S\ref{sec:time_nll}), partially offsetting
the cost of pool-based AL.

\paragraph{Scope.} No human subjects were involved. \textsc{\textbf{HybridAL}} is
intended for research use; we do not recommend it for
safety-critical applications where miscalibrated probabilities
carry direct welfare consequences.

\section*{Acknowledgments}
 This work is partially supported by the Lando Kravetz Fund, Technion's Grant 2073351.

\bibliography{references}

@article{schuirmann1987comparison,
  title={A comparison of the two one-sided tests procedure and the power approach for assessing the equivalence of average bioavailability},
  author={Schuirmann, Donald J},
  journal={Journal of pharmacokinetics and biopharmaceutics},
  volume={15},
  number={6},
  pages={657--680},
  year={1987},
  publisher={Springer}
}

@article{zhang2015character,
  title={Character-level convolutional networks for text classification},
  author={Zhang, Xiang and Zhao, Junbo and LeCun, Yann},
  journal={Advances in neural information processing systems},
  volume={28},
  year={2015}
}

@inproceedings{devlin2019bert,
  title={Bert: Pre-training of deep bidirectional transformers for language understanding},
  author={Devlin, Jacob and Chang, Ming-Wei and Lee, Kenton and Toutanova, Kristina},
  booktitle={Proceedings of the 2019 conference of the North American chapter of the association for computational linguistics: human language technologies, volume 1 (long and short papers)},
  pages={4171--4186},
  year={2019}
}

@inproceedings{barbieri2020tweeteval,
  title={TweetEval: Unified benchmark and comparative evaluation for tweet classification},
  author={Barbieri, Francesco and Camacho-Collados, Jose and Anke, Luis Espinosa and Neves, Leonardo},
  booktitle={Findings of the association for computational linguistics: EMNLP 2020},
  pages={1644--1650},
  year={2020}
}

@inproceedings{maas2011learning,
  title={Learning word vectors for sentiment analysis},
  author={Maas, Andrew and Daly, Raymond E and Pham, Peter T and Huang, Dan and Ng, Andrew Y and Potts, Christopher},
  booktitle={Proceedings of the 49th annual meeting of the association for computational linguistics: Human language technologies},
  pages={142--150},
  year={2011}
}

@inproceedings{socher2013recursive,
  title={Recursive deep models for semantic compositionality over a sentiment treebank},
  author={Socher, Richard and Perelygin, Alex and Wu, Jean and Chuang, Jason and Manning, Christopher D and Ng, Andrew Y and Potts, Christopher},
  booktitle={Proceedings of the 2013 conference on empirical methods in natural language processing},
  pages={1631--1642},
  year={2013}
}

@article{liu2019roberta,
  title={Roberta: A robustly optimized bert pretraining approach},
  author={Liu, Yinhan and Ott, Myle and Goyal, Naman and Du, Jingfei and Joshi, Mandar and Chen, Danqi and Levy, Omer and Lewis, Mike and Zettlemoyer, Luke and Stoyanov, Veselin},
  journal={arXiv preprint arXiv:1907.11692},
  year={2019}
}

@article{sanh2019distilbert,
  title={DistilBERT, a distilled version of BERT: smaller, faster, cheaper and lighter},
  author={Sanh, Victor and Debut, Lysandre and Chaumond, Julien and Wolf, Thomas},
  journal={arXiv preprint arXiv:1910.01108},
  year={2019}
}

@article{ash2019deep,
  title={Deep batch active learning by diverse, uncertain gradient lower bounds},
  author={Ash, Jordan T and Zhang, Chicheng and Krishnamurthy, Akshay and Langford, John and Agarwal, Alekh},
  journal={arXiv preprint arXiv:1906.03671},
  year={2019}
}

@article{beck2021effective,
  title={Effective evaluation of deep active learning on image classification tasks},
  author={Beck, Nathan and Sivasubramanian, Durga and Dani, Apurva and Ramakrishnan, Ganesh and Iyer, Rishabh},
  journal={arXiv preprint arXiv:2106.15324},
  year={2021}
}

@article{ren2021survey,
  title={A survey of deep active learning},
  author={Ren, Pengzhen and Xiao, Yun and Chang, Xiaojun and Huang, Po-Yao and Li, Zhihui and Gupta, Brij B and Chen, Xiaojiang and Wang, Xin},
  journal={ACM computing surveys (CSUR)},
  volume={54},
  number={9},
  pages={1--40},
  year={2021},
  publisher={ACM New York, NY}
}

@article{ash2020warm,
  title={On warm-starting neural network training},
  author={Ash, Jordan and Adams, Ryan P},
  journal={Advances in neural information processing systems},
  volume={33},
  pages={3884--3894},
  year={2020}
}

@inproceedings{tan2024large,
  title={Large language models for data annotation and synthesis: A survey},
  author={Tan, Zhen and Li, Dawei and Wang, Song and Beigi, Alimohammad and Jiang, Bohan and Bhattacharjee, Amrita and Karami, Mansooreh and Li, Jundong and Cheng, Lu and Liu, Huan},
  booktitle={Proceedings of the 2024 Conference on Empirical Methods in Natural Language Processing},
  pages={930--957},
  year={2024}
}

@inproceedings{li2025generation,
  title={From generation to judgment: Opportunities and challenges of llm-as-a-judge},
  author={Li, Dawei and Jiang, Bohan and Huang, Liangjie and Beigi, Alimohammad and Zhao, Chengshuai and Tan, Zhen and Bhattacharjee, Amrita and Jiang, Yuxuan and Chen, Canyu and Wu, Tianhao and others},
  booktitle={Proceedings of the 2025 Conference on Empirical Methods in Natural Language Processing},
  pages={2757--2791},
  year={2025}
}

@article{gilardi2023chatgpt,
  title={ChatGPT outperforms crowd workers for text-annotation tasks},
  author={Gilardi, Fabrizio and Alizadeh, Meysam and Kubli, Ma{\"e}l},
  journal={Proceedings of the National Academy of Sciences},
  volume={120},
  number={30},
  pages={e2305016120},
  year={2023},
  publisher={National Academy of Sciences}
}

@inproceedings{ashktorab2025evalassist,
  title={EvalAssist: Insights on Task-Specific Evaluations and AI-Assisted Judgment Strategy Preferences},
  author={Ashktorab, Zahra and Desmond, Michael and Pan, Qian and Johnson, James M and Santill{\'a}n Cooper, Martin and Daly, Elizabeth M and Nair, Rahul and Pedapati, Tejaswini and Do, Hyo Jin and Geyer, Werner},
  booktitle={Proceedings of the 38th Annual ACM Symposium on User Interface Software and Technology},
  pages={1--23},
  year={2025}
}

@inproceedings{chen2024humans,
  title={Humans or LLMs as the judge? a study on judgement bias},
  author={Chen, Guiming Hardy and Chen, Shunian and Liu, Ziche and Jiang, Feng and Wang, Benyou},
  booktitle={Proceedings of the 2024 Conference on Empirical Methods in Natural Language Processing},
  pages={8301--8327},
  year={2024}
}

@inproceedings{szymanski2025limitations,
  title={Limitations of the llm-as-a-judge approach for evaluating llm outputs in expert knowledge tasks},
  author={Szymanski, Annalisa and Ziems, Noah and Eicher-Miller, Heather A and Li, Toby Jia-Jun and Jiang, Meng and Metoyer, Ronald A},
  booktitle={Proceedings of the 30th international conference on intelligent user interfaces},
  pages={952--966},
  year={2025}
}

@inproceedings{calderon2025alternative,
  title={The alternative annotator test for llm-as-a-judge: How to statistically justify replacing human annotators with llms},
  author={Calderon, Nitay and Reichart, Roi and Dror, Rotem},
  booktitle={Proceedings of the 63rd Annual Meeting of the Association for Computational Linguistics (Volume 1: Long Papers)},
  pages={16051--16081},
  year={2025}
}

@techreport{settles2009active,
  title={Active learning literature survey},
  author={Settles, Burr},
  year={2009},
  institution={University of Wisconsin-Madison Department of Computer Sciences},
  number={1648}
}

@inproceedings{kornblith2019similarity,
  title={Similarity of neural network representations revisited},
  author={Kornblith, Simon and Norouzi, Mohammad and Lee, Honglak and Hinton, Geoffrey},
  booktitle={International conference on machine learning},
  pages={3519--3529},
  year={2019},
  organization={PMlR}
}

@inproceedings{wolf2020transformers,
  title={Transformers: State-of-the-art natural language processing},
  author={Wolf, Thomas and Debut, Lysandre and Sanh, Victor and Chaumond, Julien and Delangue, Clement and Moi, Anthony and Cistac, Pierric and Rault, Tim and Louf, R{\'e}mi and Funtowicz, Morgan and others},
  booktitle={Proceedings of the 2020 conference on empirical methods in natural language processing: system demonstrations},
  pages={38--45},
  year={2020}
}

@article{paszke2019pytorch,
  title={Pytorch: An imperative style, high-performance deep learning library},
  author={Paszke, Adam and Gross, Sam and Massa, Francisco and Lerer, Adam and Bradbury, James and Chanan, Gregory and Killeen, Trevor and Lin, Zeming and Gimelshein, Natalia and Antiga, Luca and others},
  journal={Advances in neural information processing systems},
  volume={32},
  year={2019}
}

@article{loshchilov2017decoupled,
  title={Decoupled weight decay regularization},
  author={Loshchilov, Ilya and Hutter, Frank},
  journal={arXiv preprint arXiv:1711.05101},
  year={2017}
}

@inproceedings{wulczyn2017ex,
  title={Ex machina: Personal attacks seen at scale},
  author={Wulczyn, Ellery and Thain, Nithum and Dixon, Lucas},
  booktitle={Proceedings of the 26th international conference on world wide web},
  pages={1391--1399},
  year={2017}
}

@article{das2023accelerating,
  title={Accelerating batch active learning using continual learning techniques},
  author={Das, Arnav and Bhatt, Gantavya and Bhalerao, Megh and Gao, Vianne and Yang, Rui and Bilmes, Jeff},
  journal={arXiv preprint arXiv:2305.06408},
  year={2023}
}

@inproceedings{sap2019risk,
  title={The risk of racial bias in hate speech detection},
  author={Sap, Maarten and Card, Dallas and Gabriel, Saadia and Choi, Yejin and Smith, Noah A},
  booktitle={Proceedings of the 57th annual meeting of the association for computational linguistics},
  pages={1668--1678},
  year={2019}
}

@inproceedings{bender2021dangers,
  title={On the dangers of stochastic parrots: Can language models be too big?},
  author={Bender, Emily M and Gebru, Timnit and McMillan-Major, Angelina and Shmitchell, Shmargaret},
  booktitle={Proceedings of the 2021 ACM conference on fairness, accountability, and transparency},
  pages={610--623},
  year={2021}
}

@inproceedings{munagala2022clactive,
  title={Clactive: Episodic memories for rapid active learning},
  author={Munagala, Sri Aurobindo and Subramanian, Sidhant and Karthik, Shyamgopal and Prabhu, Ameya and Namboodiri, Anoop},
  booktitle={Conference on Lifelong Learning Agents},
  pages={430--440},
  year={2022},
  organization={PMLR}
}

@inproceedings{guo2017calibration,
  title={On calibration of modern neural networks},
  author={Guo, Chuan and Pleiss, Geoff and Sun, Yu and Weinberger, Kilian Q},
  booktitle={International conference on machine learning},
  pages={1321--1330},
  year={2017},
  organization={PMLR}
}

@article{rolnick2019experience,
  title={Experience replay for continual learning},
  author={Rolnick, David and Ahuja, Arun and Schwarz, Jonathan and Lillicrap, Timothy and Wayne, Gregory},
  journal={Advances in neural information processing systems},
  volume={32},
  year={2019}
}

@article{dohare2024loss,
  title={Loss of plasticity in deep continual learning},
  author={Dohare, Shibhansh and Hernandez-Garcia, J Fernando and Lan, Qingfeng and Rahman, Parash and Mahmood, A Rupam and Sutton, Richard S},
  journal={Nature},
  volume={632},
  number={8026},
  pages={768--774},
  year={2024},
  publisher={Nature Publishing Group UK London}
}

@article{martin2021implicit,
  title={Implicit self-regularization in deep neural networks: Evidence from random matrix theory and implications for learning},
  author={Martin, Charles H and Mahoney, Michael W},
  journal={Journal of Machine Learning Research},
  volume={22},
  number={165},
  pages={1--73},
  year={2021}
}

@article{papyan2020prevalence,
  title={Prevalence of neural collapse during the terminal phase of deep learning training},
  author={Papyan, Vardan and Han, XY and Donoho, David L},
  journal={Proceedings of the National Academy of Sciences},
  volume={117},
  number={40},
  pages={24652--24663},
  year={2020},
  publisher={National Academy of Sciences}
}

@inproceedings{scala2024play,
  title={Play it straight: An intelligent data pruning technique for green-ai},
  author={Scala, Francesco and Flesca, Sergio and Pontieri, Luigi},
  booktitle={International conference on discovery science},
  pages={69--85},
  year={2024},
  organization={Springer}
}

@article{scala2025efficient,
  title={An efficient model training framework for green AI},
  author={Scala, Francesco and Flesca, Sergio and Pontieri, Luigi},
  journal={Machine Learning},
  volume={114},
  number={12},
  pages={275},
  year={2025},
  publisher={Springer}
}

@inproceedings{liu2018learning,
  title={Learning how to actively learn: A deep imitation learning approach},
  author={Liu, Ming and Buntine, Wray and Haffari, Gholamreza},
  booktitle={Proceedings of the 56th Annual Meeting of the Association for Computational Linguistics (Volume 1: Long Papers)},
  pages={1874--1883},
  year={2018}
}

@inproceedings{hacohen2022active,
  title={Active Learning on a Budget: Opposite Strategies Suit High and Low Budgets},
  author={Hacohen, Guy and Dekel, Avihu and Weinshall, Daphna},
  booktitle={International Conference on Machine Learning},
  pages={8175--8195},
  year={2022},
  organization={PMLR}
}

@article{hacohen2023select,
  title={How to select which active learning strategy is best suited for your specific problem and budget},
  author={Hacohen, Guy and Weinshall, Daphna},
  journal={Advances in Neural Information Processing Systems},
  volume={36},
  pages={13395--13407},
  year={2023}
}

@inproceedings{fang2017learning,
  title={Learning how to active learn: A deep reinforcement learning approach},
  author={Fang, Meng and Li, Yuan and Cohn, Trevor},
  booktitle={Proceedings of the 2017 conference on empirical methods in natural language processing},
  pages={595--605},
  year={2017}
}

@article{zhang2023algorithm,
  title={Algorithm selection for deep active learning with imbalanced datasets},
  author={Zhang, Jifan and Shao, Shuai and Verma, Saurabh and Nowak, Robert},
  journal={Advances in Neural Information Processing Systems},
  volume={36},
  pages={9614--9647},
  year={2023}
}

@article{han2024adaptive,
  title={An adaptive active learning method for multiclass imbalanced data streams with concept drift},
  author={Han, Meng and Li, Chunpeng and Meng, Fanxing and He, Feifei and Zhang, Ruihua},
  journal={Applied Sciences},
  volume={14},
  number={16},
  pages={7176},
  year={2024},
  publisher={MDPI}
}

@inproceedings{zhu2008multi,
  title={Multi-criteria-based strategy to stop active learning for data annotation},
  author={Zhu, Jingbo and Wang, Huizhen and Hovy, Eduard},
  booktitle={Proceedings of the 22nd International Conference on Computational Linguistics (Coling 2008)},
  pages={1129--1136},
  year={2008}
}

@article{vlachos2008stopping,
  title={A stopping criterion for active learning},
  author={Vlachos, Andreas},
  journal={Computer Speech \& Language},
  volume={22},
  number={3},
  pages={295--312},
  year={2008},
  publisher={Elsevier}
}

\appendix

\section{Dataset Details}\label{app:datasets}

We briefly describe each dataset below; per-class distributions for
the splits used in our experiments are in
Table~\ref{tab:datasets-class-distribution}.

\paragraph{IMDb~\citep{maas2011learning}.}
50{,}000 movie reviews from the Internet Movie Database, labelled
positive (review score $\geq 7$) or negative ($\leq 4$). Train and test splits are class-balanced 50/50; we use an
80/20 train/test partition with a $1\%$ validation holdout
from train.

\paragraph{Jigsaw~\citep{wulczyn2017ex}.}
159{,}571 comments from English Wikipedia talk pages, originally
annotated for six types of toxicity (toxic, severely toxic, obscene,
threatening, insulting, and identity-hate). We binarize using the
primary toxicity label, yielding a heavily imbalanced binary task
($\approx 9.6\%$ positive class).

\paragraph{SST-2~\citep{socher2013recursive}.}
Movie-review snippets from the Stanford Sentiment Treebank with
binary positive/negative sentiment labels. We use the GLUE
formulation, whose held-out validation split (used here as the
test set) is class-balanced at ${\approx}49/51\%$.

\paragraph{TweetEval~\citep{barbieri2020tweeteval}.}
The sentiment subtask of TweetEval. Tweets are labelled negative,
neutral, or positive. The official train/test class distributions
differ substantially (Table~\ref{tab:datasets-class-distribution}),
making this dataset a useful stress test under label-distribution
shift.

\paragraph{AG News~\citep{zhang2015character}.}
A balanced four-class news-topic classification corpus drawn from
the AG news collection. Classes are \emph{World}, \emph{Sports},
\emph{Business}, and \emph{Sci/Tech}, with $30{,}000$
training examples per class in the original release
($29{,}700$ in our AL pool after the $1\%$ validation
holdout).

\paragraph{Yahoo Answers~\citep{zhang2015character}.}
A 10-class question-topic classification corpus derived from the
Yahoo!\ Answers comprehensive Q\&A dataset. The original release
contains 1.4M training and 60k test examples; for computational
tractability we stratified-downsample to $6{,}000$ documents per
class ($4{,}950$ train, $50$ validation held out from train,
$1{,}000$ test), preserving the uniform class distribution.

\begin{table*}[t]
\centering
\small
\setlength{\tabcolsep}{4pt}
\begin{tabular}{llrrrrrr}
\toprule
Dataset & Class & Train ($n$) & Train (\%) & Val.\ ($n$) & Val.\ (\%) & Test ($n$) & Test (\%) \\
\midrule
\multirow{3}{*}{IMDb}
  & negative              & 19{,}750  & 50.00 &   250 & 50.00 &  5{,}000 & 50.00 \\
  & positive              & 19{,}750  & 50.00 &   250 & 50.00 &  5{,}000 & 50.00 \\
  & \textit{Total}        & \textit{39{,}500}  & --- & \textit{500} & --- & \textit{10{,}000} & --- \\
\midrule
\multirow{3}{*}{Jigsaw}
  & non-toxic             & 113{,}978 & 90.42 & 1{,}443 & 90.41 & 28{,}856 & 90.42 \\
  & toxic                 & 12{,}082  &  9.58 &   153 &  9.59 &  3{,}059 &  9.58 \\
  & \textit{Total}        & \textit{126{,}060} & --- & \textit{1{,}596} & --- & \textit{31{,}915} & --- \\
\midrule
\multirow{3}{*}{SST-2}
  & negative              & 29{,}482  & 44.22 &   298 & 44.21 &     428 & 49.08 \\
  & positive              & 37{,}193  & 55.78 &   376 & 55.79 &     444 & 50.92 \\
  & \textit{Total}        & \textit{66{,}675}  & --- & \textit{674} & --- & \textit{872} & --- \\
\midrule
\multirow{4}{*}{TweetEval}
  & negative              &  7{,}331  & 15.55 &    74 & 15.51 &  3{,}972 & 32.33 \\
  & neutral               & 21{,}326  & 45.24 &   216 & 45.28 &  5{,}937 & 48.33 \\
  & positive              & 18{,}481  & 39.21 &   187 & 39.20 &  2{,}375 & 19.33 \\
  & \textit{Total}        & \textit{47{,}138}  & --- & \textit{477} & --- & \textit{12{,}284} & --- \\
\midrule
\multirow{5}{*}{AG News}
  & World                 & 29{,}700  & 25.00 &   300 & 25.00 &  1{,}900 & 25.00 \\
  & Sports                & 29{,}700  & 25.00 &   300 & 25.00 &  1{,}900 & 25.00 \\
  & Business              & 29{,}700  & 25.00 &   300 & 25.00 &  1{,}900 & 25.00 \\
  & Sci/Tech              & 29{,}700  & 25.00 &   300 & 25.00 &  1{,}900 & 25.00 \\
  & \textit{Total}        & \textit{118{,}800} & --- & \textit{1{,}200} & --- & \textit{7{,}600} & --- \\
\midrule
\multirow{11}{*}{Yahoo Answers}
  & Society \& Culture       & 4{,}950 & 10.00 & 50 & 10.00 & 1{,}000 & 10.00 \\
  & Science \& Mathematics   & 4{,}950 & 10.00 & 50 & 10.00 & 1{,}000 & 10.00 \\
  & Health                   & 4{,}950 & 10.00 & 50 & 10.00 & 1{,}000 & 10.00 \\
  & Education \& Reference   & 4{,}950 & 10.00 & 50 & 10.00 & 1{,}000 & 10.00 \\
  & Computers \& Internet    & 4{,}950 & 10.00 & 50 & 10.00 & 1{,}000 & 10.00 \\
  & Sports                   & 4{,}950 & 10.00 & 50 & 10.00 & 1{,}000 & 10.00 \\
  & Business \& Finance      & 4{,}950 & 10.00 & 50 & 10.00 & 1{,}000 & 10.00 \\
  & Entertainment \& Music   & 4{,}950 & 10.00 & 50 & 10.00 & 1{,}000 & 10.00 \\
  & Family \& Relationships  & 4{,}950 & 10.00 & 50 & 10.00 & 1{,}000 & 10.00 \\
  & Politics \& Government   & 4{,}950 & 10.00 & 50 & 10.00 & 1{,}000 & 10.00 \\
  & \textit{Total}           & \textit{49{,}500}  & --- & \textit{500} & --- & \textit{10{,}000} & --- \\
\bottomrule
\end{tabular}
\caption{Per-class distribution of each dataset across train, validation
and test splits (seed=42 splits as used in our experiments). Train counts
reflect the AL pool after validation holdout; \textit{Total} rows give
the size of each split.}
\label{tab:datasets-class-distribution}
\end{table*}

\section{Main Results: Full Breakdown}
\label{app:main}

This appendix provides the complete per-cell breakdown of the
main results: test \fone{} (\S\ref{app:f1}), non-inferiority
tests (\S\ref{app:tost}), training time and test NLL
(\S\ref{app:time_nll}), post-switch signal stability
(\S\ref{app:post_switch}), and the temperature analysis behind
the calibration differences (\S\ref{app:temp_scaling}).

\subsection{Per-dataset Test \fone{}}
\label{app:f1}

Table~\ref{tab:appendix_f1_full} reports test \fone{} for every
(method $\times$ backbone $\times$ dataset) cell across 5
seeds. The four \textsc{FixedSwitch} variants ($k\in\{3,5,7,10\}$) and
the two \textsc{\textbf{HybridAL}} variants land within the same envelope as
\textsc{Retrain} and \textsc{FineTune}: no cell shows a gap larger than $0.035$
\fone{} across these eight methods (mean spread $0.015$),
confirming that switching the training strategy at any round
leaves test \fone{} within the same envelope; formal
non-inferiority tests are in \S\ref{app:tost}. \textsc{\textbf{HybridAL}}($\Delta\alpha$) is
statistically indistinguishable from both \textsc{Retrain} and \textsc{FineTune}
on $16/18$ (backbone, dataset) cells (paired $t$-test,
$p>0.05$); \textsc{\textbf{HybridAL}}($\Delta$Acc) is indistinguishable from
\textsc{Retrain} on $17/18$ cells and from \textsc{FineTune} on $15/18$ cells.
\textsc{NewOnly}'s deficit on DistilBERT and BERT is localized
to the hard multi-class datasets (TweetEval, Yahoo Answers)
and disappears on the easier binary tasks (IMDb, Jigsaw,
SST-2); the gap closes on RoBERTa (\S\ref{sec:f1_parity}).

\begin{table*}[t]
  \centering
  \scriptsize
  \setlength{\tabcolsep}{3pt}
  \resizebox{\textwidth}{!}{%
  \begin{tabular}{lcccccc}
  \toprule
  \multicolumn{7}{l}{\textbf{DistilBERT}} \\
  Method & IMDb & AGNews & Jigsaw & SST-2 & TwtEv & Yahoo \\
  \midrule
  \textsc{Retrain} & \cellcolor{green!15} 0.819\,$\pm$\,0.021 & \cellcolor{green!10} 0.900\,$\pm$\,0.004 & \cellcolor{green!36} 0.881\,$\pm$\,0.003 & \cellcolor{green!0} 0.835\,$\pm$\,0.021 & \cellcolor{green!55} 0.634\,$\pm$\,0.023 & \cellcolor{green!55} 0.667\,$\pm$\,0.004$^{\ddagger\S\P\star\bullet\circ}$ \\
  \textsc{FineTune} & \cellcolor{green!27} 0.823\,$\pm$\,0.008$^{\P}$ & \cellcolor{green!37} 0.904\,$\pm$\,0.005 & \cellcolor{green!53} 0.883\,$\pm$\,0.004 & \cellcolor{green!51} 0.858\,$\pm$\,0.018 & \cellcolor{green!49} 0.626\,$\pm$\,0.021 & \cellcolor{green!32} 0.655\,$\pm$\,0.006$^{\S}$ \\
  \textsc{NewOnly} & \cellcolor{green!27} 0.823\,$\pm$\,0.006 & \cellcolor{green!0} 0.899\,$\pm$\,0.001 & \cellcolor{green!41} 0.882\,$\pm$\,0.007 & \cellcolor{green!41} 0.854\,$\pm$\,0.009 & \cellcolor{green!0} 0.552\,$\pm$\,0.117 & \cellcolor{green!0} 0.637\,$\pm$\,0.009 \\
  \textsc{\textbf{HybridAL}} ($\Delta\alpha$) & \cellcolor{green!0} 0.814\,$\pm$\,0.014 & \cellcolor{green!26} 0.902\,$\pm$\,0.007 & \cellcolor{green!33} 0.881\,$\pm$\,0.004$^{\circ}$ & \cellcolor{green!18} 0.844\,$\pm$\,0.022 & \cellcolor{green!50} 0.627\,$\pm$\,0.025 & \cellcolor{green!36} 0.656\,$\pm$\,0.004$^{\S}$ \\
  \textsc{\textbf{HybridAL}} ($\Delta$Acc) & \cellcolor{green!38} 0.827\,$\pm$\,0.006 & \cellcolor{green!46} 0.905\,$\pm$\,0.005 & \cellcolor{green!24} 0.880\,$\pm$\,0.005 & \cellcolor{green!49} 0.857\,$\pm$\,0.012 & \cellcolor{green!52} 0.629\,$\pm$\,0.018 & \cellcolor{green!31} 0.654\,$\pm$\,0.005$^{\S}$ \\
  \textsc{FixedSwitch} (3) & \cellcolor{green!30} 0.824\,$\pm$\,0.006 & \cellcolor{green!55} 0.906\,$\pm$\,0.004$^{\S}$ & \cellcolor{green!23} 0.880\,$\pm$\,0.004 & \cellcolor{green!38} 0.852\,$\pm$\,0.023 & \cellcolor{green!46} 0.621\,$\pm$\,0.022 & \cellcolor{green!33} 0.655\,$\pm$\,0.005$^{\S}$ \\
  \textsc{FixedSwitch} (5) & \cellcolor{green!18} 0.820\,$\pm$\,0.014 & \cellcolor{green!19} 0.902\,$\pm$\,0.006 & \cellcolor{green!5} 0.878\,$\pm$\,0.006 & \cellcolor{green!55} 0.860\,$\pm$\,0.015 & \cellcolor{green!47} 0.622\,$\pm$\,0.008 & \cellcolor{green!36} 0.657\,$\pm$\,0.006$^{\S\star}$ \\
  \textsc{FixedSwitch} (7) & \cellcolor{green!55} 0.833\,$\pm$\,0.006$^{\bullet}$ & \cellcolor{green!36} 0.904\,$\pm$\,0.003$^{\S}$ & \cellcolor{green!0} 0.877\,$\pm$\,0.005 & \cellcolor{green!33} 0.850\,$\pm$\,0.020 & \cellcolor{green!46} 0.621\,$\pm$\,0.023 & \cellcolor{green!33} 0.655\,$\pm$\,0.007$^{\S}$ \\
  \textsc{FixedSwitch} (10) & \cellcolor{green!41} 0.828\,$\pm$\,0.005 & \cellcolor{green!38} 0.904\,$\pm$\,0.007 & \cellcolor{green!55} 0.883\,$\pm$\,0.005 & \cellcolor{green!50} 0.857\,$\pm$\,0.016 & \cellcolor{green!51} 0.628\,$\pm$\,0.019 & \cellcolor{green!39} 0.658\,$\pm$\,0.008$^{\S}$ \\
  \midrule
  \addlinespace[2pt]
  \multicolumn{7}{l}{\textbf{BERT}} \\
  Method & IMDb & AGNews & Jigsaw & SST-2 & TwtEv & Yahoo \\
  \midrule
  \textsc{Retrain} & \cellcolor{green!1} 0.826\,$\pm$\,0.013 & \cellcolor{green!41} 0.906\,$\pm$\,0.004$^{\S}$ & \cellcolor{green!0} 0.870\,$\pm$\,0.014 & \cellcolor{green!15} 0.875\,$\pm$\,0.021 & \cellcolor{green!36} 0.608\,$\pm$\,0.013 & \cellcolor{green!55} 0.667\,$\pm$\,0.002$^{\S\P\diamond}$ \\
  \textsc{FineTune} & \cellcolor{green!42} 0.840\,$\pm$\,0.010$^{\dagger\star}$ & \cellcolor{green!35} 0.905\,$\pm$\,0.004 & \cellcolor{green!40} 0.880\,$\pm$\,0.007 & \cellcolor{green!35} 0.885\,$\pm$\,0.007$^{\star}$ & \cellcolor{green!44} 0.622\,$\pm$\,0.014 & \cellcolor{green!55} 0.666\,$\pm$\,0.005$^{\P\star\diamond}$ \\
  \textsc{NewOnly} & \cellcolor{green!31} 0.836\,$\pm$\,0.012 & \cellcolor{green!0} 0.899\,$\pm$\,0.004 & \cellcolor{green!25} 0.876\,$\pm$\,0.008 & \cellcolor{green!17} 0.876\,$\pm$\,0.009 & \cellcolor{green!0} 0.550\,$\pm$\,0.090 & \cellcolor{green!0} 0.608\,$\pm$\,0.045 \\
  \textsc{\textbf{HybridAL}} ($\Delta\alpha$) & \cellcolor{green!25} 0.834\,$\pm$\,0.012 & \cellcolor{green!27} 0.903\,$\pm$\,0.006$^{\S}$ & \cellcolor{green!55} 0.884\,$\pm$\,0.006$^{\S}$ & \cellcolor{green!55} 0.895\,$\pm$\,0.011$^{\S\star}$ & \cellcolor{green!51} 0.633\,$\pm$\,0.019 & \cellcolor{green!47} 0.658\,$\pm$\,0.006 \\
  \textsc{\textbf{HybridAL}} ($\Delta$Acc) & \cellcolor{green!0} 0.826\,$\pm$\,0.013 & \cellcolor{green!55} 0.908\,$\pm$\,0.002$^{\S}$ & \cellcolor{green!49} 0.883\,$\pm$\,0.009 & \cellcolor{green!0} 0.868\,$\pm$\,0.013 & \cellcolor{green!36} 0.608\,$\pm$\,0.013 & \cellcolor{green!51} 0.662\,$\pm$\,0.007 \\
  \textsc{FixedSwitch} (3) & \cellcolor{green!55} 0.845\,$\pm$\,0.007$^{\dagger\star}$ & \cellcolor{green!42} 0.906\,$\pm$\,0.005 & \cellcolor{green!50} 0.883\,$\pm$\,0.006$^{\S}$ & \cellcolor{green!27} 0.881\,$\pm$\,0.017 & \cellcolor{green!46} 0.624\,$\pm$\,0.021 & \cellcolor{green!51} 0.662\,$\pm$\,0.013 \\
  \textsc{FixedSwitch} (5) & \cellcolor{green!36} 0.838\,$\pm$\,0.010 & \cellcolor{green!25} 0.903\,$\pm$\,0.005 & \cellcolor{green!32} 0.878\,$\pm$\,0.016 & \cellcolor{green!40} 0.887\,$\pm$\,0.006 & \cellcolor{green!44} 0.621\,$\pm$\,0.010 & \cellcolor{green!53} 0.665\,$\pm$\,0.007 \\
  \textsc{FixedSwitch} (7) & \cellcolor{green!53} 0.844\,$\pm$\,0.006$^{\dagger\star}$ & \cellcolor{green!51} 0.908\,$\pm$\,0.005$^{\S}$ & \cellcolor{green!49} 0.883\,$\pm$\,0.005$^{\S}$ & \cellcolor{green!25} 0.880\,$\pm$\,0.017 & \cellcolor{green!45} 0.622\,$\pm$\,0.024 & \cellcolor{green!43} 0.654\,$\pm$\,0.008 \\
  \textsc{FixedSwitch} (10) & \cellcolor{green!40} 0.839\,$\pm$\,0.011 & \cellcolor{green!40} 0.906\,$\pm$\,0.003$^{\S}$ & \cellcolor{green!40} 0.880\,$\pm$\,0.011 & \cellcolor{green!44} 0.889\,$\pm$\,0.010 & \cellcolor{green!55} 0.639\,$\pm$\,0.009$^{\dagger\ddagger\star}$ & \cellcolor{green!45} 0.656\,$\pm$\,0.012 \\
  \midrule
  \addlinespace[2pt]
  \multicolumn{7}{l}{\textbf{RoBERTa}} \\
  Method & IMDb & AGNews & Jigsaw & SST-2 & TwtEv & Yahoo \\
  \midrule
  \textsc{Retrain} & \cellcolor{green!0} 0.862\,$\pm$\,0.033 & \cellcolor{green!36} 0.911\,$\pm$\,0.005 & \cellcolor{green!0} 0.877\,$\pm$\,0.013 & \cellcolor{green!0} 0.889\,$\pm$\,0.015 & \cellcolor{green!9} 0.647\,$\pm$\,0.022 & \cellcolor{green!44} 0.672\,$\pm$\,0.006 \\
  \textsc{FineTune} & \cellcolor{green!36} 0.878\,$\pm$\,0.005 & \cellcolor{green!0} 0.904\,$\pm$\,0.008 & \cellcolor{green!33} 0.882\,$\pm$\,0.007 & \cellcolor{green!45} 0.906\,$\pm$\,0.009 & \cellcolor{green!44} 0.659\,$\pm$\,0.015 & \cellcolor{green!26} 0.668\,$\pm$\,0.009 \\
  \textsc{NewOnly} & \cellcolor{green!55} 0.887\,$\pm$\,0.003$^{\ddagger\P\star\circ}$ & \cellcolor{green!15} 0.907\,$\pm$\,0.003 & \cellcolor{green!19} 0.880\,$\pm$\,0.008 & \cellcolor{green!39} 0.904\,$\pm$\,0.008 & \cellcolor{green!9} 0.647\,$\pm$\,0.026 & \cellcolor{green!0} 0.663\,$\pm$\,0.014 \\
  \textsc{\textbf{HybridAL}} ($\Delta\alpha$) & \cellcolor{green!25} 0.873\,$\pm$\,0.006 & \cellcolor{green!41} 0.912\,$\pm$\,0.005 & \cellcolor{green!20} 0.880\,$\pm$\,0.012 & \cellcolor{green!32} 0.901\,$\pm$\,0.013 & \cellcolor{green!23} 0.652\,$\pm$\,0.020 & \cellcolor{green!24} 0.668\,$\pm$\,0.007 \\
  \textsc{\textbf{HybridAL}} ($\Delta$Acc) & \cellcolor{green!45} 0.882\,$\pm$\,0.004$^{\P}$ & \cellcolor{green!55} 0.914\,$\pm$\,0.007 & \cellcolor{green!38} 0.883\,$\pm$\,0.004 & \cellcolor{green!17} 0.895\,$\pm$\,0.019 & \cellcolor{green!5} 0.646\,$\pm$\,0.021 & \cellcolor{green!41} 0.672\,$\pm$\,0.007$^{\S}$ \\
  \textsc{FixedSwitch} (3) & \cellcolor{green!21} 0.872\,$\pm$\,0.015 & \cellcolor{green!40} 0.911\,$\pm$\,0.006 & \cellcolor{green!31} 0.882\,$\pm$\,0.005 & \cellcolor{green!41} 0.905\,$\pm$\,0.010 & \cellcolor{green!0} 0.644\,$\pm$\,0.016 & \cellcolor{green!40} 0.672\,$\pm$\,0.005 \\
  \textsc{FixedSwitch} (5) & \cellcolor{green!33} 0.877\,$\pm$\,0.005 & \cellcolor{green!30} 0.910\,$\pm$\,0.004 & \cellcolor{green!53} 0.885\,$\pm$\,0.004 & \cellcolor{green!55} 0.910\,$\pm$\,0.010 & \cellcolor{green!24} 0.652\,$\pm$\,0.012 & \cellcolor{green!40} 0.672\,$\pm$\,0.005 \\
  \textsc{FixedSwitch} (7) & \cellcolor{green!44} 0.882\,$\pm$\,0.007$^{\ddagger}$ & \cellcolor{green!41} 0.912\,$\pm$\,0.006 & \cellcolor{green!30} 0.882\,$\pm$\,0.004 & \cellcolor{green!32} 0.901\,$\pm$\,0.012 & \cellcolor{green!55} 0.662\,$\pm$\,0.010 & \cellcolor{green!55} 0.675\,$\pm$\,0.007$^{\P}$ \\
  \textsc{FixedSwitch} (10) & \cellcolor{green!44} 0.882\,$\pm$\,0.008 & \cellcolor{green!41} 0.912\,$\pm$\,0.004 & \cellcolor{green!55} 0.886\,$\pm$\,0.004 & \cellcolor{green!46} 0.907\,$\pm$\,0.010$^{\dagger}$ & \cellcolor{green!37} 0.657\,$\pm$\,0.015$^{\bullet}$ & \cellcolor{green!39} 0.671\,$\pm$\,0.003 \\
  \bottomrule
  \end{tabular}%
  }
  \caption{Full per-(method $\times$ backbone $\times$ dataset) test \fone{} (mean $\pm$ std across 5 seeds). Cell shading is per column within each backbone (greener~$=$~higher~$=$~better). Superscripts indicate that the cell is significantly \emph{higher} than the reference method (paired $t$-test, $p<0.05$, two-sided): $\dagger$ \textsc{Retrain}, $\ddagger$ \textsc{FineTune}, $\S$ \textsc{NewOnly}, $\P$ \textsc{\textbf{HybridAL}} ($\Delta\alpha$), $\star$ \textsc{\textbf{HybridAL}} ($\Delta$Acc), $\bullet$ \textsc{FixedSwitch} (3), $\circ$ \textsc{FixedSwitch} (5), $\diamond$ \textsc{FixedSwitch} (7), $\triangle$ \textsc{FixedSwitch} (10).}
  \label{tab:appendix_f1_full}
\end{table*}

\subsection{Non-Inferiority of Endpoint \fone{}}
\label{app:tost}

Table~\ref{tab:tost_pooled} reports one-sided $95\%$ lower
confidence bounds on the mean paired \fone{} difference between
each \textsc{\textbf{HybridAL}} variant and each reference method.
A non-significant paired $t$-test does not establish equivalence,
so we assess non-inferiority directly with the two one-sided tests
procedure \citep{schuirmann1987comparison}.
For each comparison we form the paired difference
$d = \fone{}(\textsc{\textbf{HybridAL}}) - \fone{}(\text{reference})$
over all $90$ (backbone, dataset, seed) cells and test
$H_0\!: \mathrm{E}[d] \leq -\delta$ against
$H_1\!: \mathrm{E}[d] > -\delta$ at $\alpha{=}0.05$, equivalently
a paired $t$-test with the null shifted by the margin $\delta$.
We anchor $\delta$ to the noise floor of the experiment:
\textsc{Retrain}'s seed-to-seed \fone{} standard deviation
averages $0.0131$ across the $18$ (backbone, dataset) cells
(median $0.013$, max $0.033$), so $\delta{=}0.005$ is about one
third of that floor and $\delta{=}0.010$ about three quarters.
\begin{table}[t]
  \centering
  \small
  \setlength{\tabcolsep}{4pt}
  \resizebox{\columnwidth}{!}{%
  \begin{tabular}{ll|c|cc}
  \toprule
  Variant & Reference & Lower bound & $\delta{=}0.005$ & $\delta{=}0.010$ \\
  \midrule
  $\Delta\alpha$  & \textsc{Retrain}  & $+0.0005$ & \ding{51} & \ding{51} \\
  $\Delta\alpha$  & \textsc{FineTune} & $-0.0045$ & \ding{51} & \ding{51} \\
  $\Delta$Acc     & \textsc{Retrain}  & $-0.0001$ & \ding{51} & \ding{51} \\
  $\Delta$Acc     & \textsc{FineTune} & $-0.0050$ & \ding{55} & \ding{51} \\
  \midrule
  $\Delta\alpha$  & per-cell best     & $-0.0086$ & \ding{55} & \ding{51} \\
  $\Delta$Acc     & per-cell best     & $-0.0091$ & \ding{55} & \ding{51} \\
  \bottomrule
  \end{tabular}}
  \caption{TOST non-inferiority of endpoint \fone{}: one-sided
  $95\%$ lower confidence bounds on the mean paired difference,
  $n{=}90$ (backbone, dataset, seed) cells. \ding{51}:
  non-inferior at that margin.}
  \label{tab:tost_pooled}
\end{table}

Both variants are non-inferior to \textsc{Retrain} and
\textsc{FineTune} individually at $\delta{=}0.010$, and to the
per-cell better of the two baselines (whichever is higher in each
individual cell) at the same margin. At the tighter
$\delta{=}0.005$, three of the four hybrid--baseline pairs pass;
\textsc{\textbf{HybridAL}}($\Delta$Acc) vs.\ \textsc{FineTune}
misses with $p{=}0.051$ and passes at $\delta{=}0.010$.
\textsc{\textbf{HybridAL}}($\Delta\alpha$) vs.\ \textsc{Retrain}
has a positive lower bound and is therefore non-inferior at any
$\delta \geq 0$. Running TOST separately within each cell would
use $n{=}5$ seeds and is underpowered by construction, so the
pooled test is the appropriate instrument for a macro-level claim
about endpoint \fone{}; all per-cell mean differences lie below
$0.025$ \fone{}, so no single cell contributes a heterogeneous
effect large enough to change the conclusion.

\subsection{Per-dataset Training Time and Test NLL}
\label{app:time_nll}

Table~\ref{tab:appendix_nll_full} reports mean test NLL and
Table~\ref{tab:appendix_time_full} mean training time in seconds,
for every (method $\times$ backbone $\times$ dataset) cell across
5 seeds. Both metrics are lower-is-better; per-column heatmap
shading within each backbone (greener~$=$~lower); superscripts
mark paired-$t$-test significance ($p<0.05$, two-sided) per
the caption legend.

The breakdown confirms the per-backbone trade-off discussed in
\S\ref{sec:time_nll}: \textsc{Retrain} reaches the lowest NLL on essentially every
cell but is the slowest; \textsc{FineTune} carries the highest
NLL among pool-trained methods. The four \textsc{FixedSwitch}
variants are consistently faster than
\textsc{\textbf{HybridAL}} on every backbone. On BERT,
\textsc{\textbf{HybridAL}}($\Delta\alpha$) is also faster
than \textsc{FineTune} (mean $819$~s vs.\ $936$~s), driven
by early switching on several datasets.
\textsc{FixedSwitch} NLL clusters near
\textsc{FineTune}'s, while both
\textsc{\textbf{HybridAL}} variants pull substantially
closer to \textsc{Retrain} (\S\ref{sec:adaptivity}).
\textsc{NewOnly} is the cheapest method overall and converges
to NLL between \textsc{Retrain}'s and \textsc{FineTune}'s,
but its \fone{} deficit on hard multi-class tasks
(\S\ref{sec:f1_parity}) keeps it off the joint Pareto
frontier.

\begin{table*}[t]
  \centering
  \setlength{\tabcolsep}{3pt}
  \resizebox{\textwidth}{!}{%
  \begin{tabular}{l|ccccccc}
  \toprule
  Method & IMDb & AGNews & Jigsaw & SST-2 & TwtEv & Yahoo & \textbf{Mean} \\
  \midrule
  \multicolumn{8}{l}{\textbf{DistilBERT}} \\
  \midrule
  \textsc{Retrain} & \cellcolor{green!55} 0.426\,$\pm$\,0.039$^{\ddagger\P\bullet\circ\diamond\triangle}$ & \cellcolor{green!55} 0.320\,$\pm$\,0.008$^{\ddagger\S\P\star\bullet\circ\diamond\triangle}$ & \cellcolor{green!55} 0.133\,$\pm$\,0.008$^{\ddagger\star\bullet\diamond}$ & \cellcolor{green!55} 0.391\,$\pm$\,0.029 & \cellcolor{green!55} 0.814\,$\pm$\,0.066$^{\ddagger\P\star\bullet\circ\diamond\triangle}$ & \cellcolor{green!55} 1.108\,$\pm$\,0.018$^{\ddagger\S\P\star\bullet\circ\diamond\triangle}$ & \cellcolor{green!55} \textbf{0.532}\,$\pm$\,0.336$^{\ddagger\S\P\star\bullet\circ\diamond\triangle}$ \\
  \textsc{FineTune} & \cellcolor{green!12} 0.614\,$\pm$\,0.075 & \cellcolor{green!0} 0.448\,$\pm$\,0.042 & \cellcolor{green!28} 0.153\,$\pm$\,0.016 & \cellcolor{green!26} 0.460\,$\pm$\,0.053 & \cellcolor{green!4} 1.379\,$\pm$\,0.075 & \cellcolor{green!0} 1.609\,$\pm$\,0.030 & \cellcolor{green!5} \textbf{0.777}\,$\pm$\,0.540 \\
  \textsc{NewOnly} & \cellcolor{green!38} 0.501\,$\pm$\,0.036$^{\ddagger\bullet\circ\diamond\triangle}$ & \cellcolor{green!37} 0.362\,$\pm$\,0.013$^{\ddagger\star\bullet\circ\diamond\triangle}$ & \cellcolor{green!55} 0.133\,$\pm$\,0.007$^{\ddagger\star\bullet\diamond}$ & \cellcolor{green!36} 0.435\,$\pm$\,0.041 & \cellcolor{green!49} 0.880\,$\pm$\,0.105$^{\ddagger\bullet\circ\diamond\triangle}$ & \cellcolor{green!38} 1.265\,$\pm$\,0.046$^{\ddagger\star\bullet\circ\diamond\triangle}$ & \cellcolor{green!42} \textbf{0.596}\,$\pm$\,0.382$^{\ddagger\P\star\bullet\circ\diamond\triangle}$ \\
  \textsc{\textbf{HybridAL}} ($\Delta\alpha$) & \cellcolor{green!6} 0.641\,$\pm$\,0.140 & \cellcolor{green!17} 0.408\,$\pm$\,0.059 & \cellcolor{green!32} 0.150\,$\pm$\,0.023$^{\circ}$ & \cellcolor{green!46} 0.412\,$\pm$\,0.038$^{\triangle}$ & \cellcolor{green!12} 1.288\,$\pm$\,0.321 & \cellcolor{green!19} 1.441\,$\pm$\,0.192 & \cellcolor{green!16} \textbf{0.724}\,$\pm$\,0.508$^{\circ}$ \\
  \textsc{\textbf{HybridAL}} ($\Delta$Acc) & \cellcolor{green!48} 0.457\,$\pm$\,0.079$^{\ddagger\bullet\circ\diamond\triangle}$ & \cellcolor{green!6} 0.433\,$\pm$\,0.023 & \cellcolor{green!24} 0.156\,$\pm$\,0.008 & \cellcolor{green!54} 0.393\,$\pm$\,0.032$^{\triangle}$ & \cellcolor{green!24} 1.163\,$\pm$\,0.246 & \cellcolor{green!13} 1.490\,$\pm$\,0.175 & \cellcolor{green!24} \textbf{0.682}\,$\pm$\,0.498$^{\ddagger\bullet\circ\diamond\triangle}$ \\
  \textsc{FixedSwitch} (3) & \cellcolor{green!9} 0.628\,$\pm$\,0.051 & \cellcolor{green!15} 0.414\,$\pm$\,0.027 & \cellcolor{green!14} 0.164\,$\pm$\,0.020 & \cellcolor{green!17} 0.481\,$\pm$\,0.102 & \cellcolor{green!3} 1.398\,$\pm$\,0.071 & \cellcolor{green!4} 1.579\,$\pm$\,0.033 & \cellcolor{green!5} \textbf{0.777}\,$\pm$\,0.535 \\
  \textsc{FixedSwitch} (5) & \cellcolor{green!1} 0.665\,$\pm$\,0.088 & \cellcolor{green!2} 0.444\,$\pm$\,0.017 & \cellcolor{green!10} 0.167\,$\pm$\,0.030 & \cellcolor{green!12} 0.493\,$\pm$\,0.157 & \cellcolor{green!0} 1.427\,$\pm$\,0.113 & \cellcolor{green!0} 1.612\,$\pm$\,0.078 & \cellcolor{green!0} \textbf{0.801}\,$\pm$\,0.547 \\
  \textsc{FixedSwitch} (7) & \cellcolor{green!13} 0.610\,$\pm$\,0.021$^{\triangle}$ & \cellcolor{green!8} 0.429\,$\pm$\,0.028 & \cellcolor{green!0} 0.174\,$\pm$\,0.027 & \cellcolor{green!10} 0.496\,$\pm$\,0.098 & \cellcolor{green!5} 1.369\,$\pm$\,0.093 & \cellcolor{green!7} 1.551\,$\pm$\,0.065 & \cellcolor{green!6} \textbf{0.772}\,$\pm$\,0.519 \\
  \textsc{FixedSwitch} (10) & \cellcolor{green!0} 0.668\,$\pm$\,0.029 & \cellcolor{green!10} 0.426\,$\pm$\,0.030 & \cellcolor{green!24} 0.156\,$\pm$\,0.020 & \cellcolor{green!0} 0.520\,$\pm$\,0.095 & \cellcolor{green!9} 1.325\,$\pm$\,0.111$^{\circ}$ & \cellcolor{green!7} 1.549\,$\pm$\,0.039$^{\ddagger}$ & \cellcolor{green!6} \textbf{0.774}\,$\pm$\,0.509 \\
  \midrule
  \multicolumn{8}{l}{\textbf{BERT}} \\
  \midrule
  \textsc{Retrain} & \cellcolor{green!55} 0.398\,$\pm$\,0.027$^{\ddagger\P\bullet\circ\diamond\triangle}$ & \cellcolor{green!55} 0.317\,$\pm$\,0.016$^{\ddagger\S\P\star\bullet\circ\diamond\triangle}$ & \cellcolor{green!55} 0.132\,$\pm$\,0.007$^{\ddagger\star\bullet\diamond\triangle}$ & \cellcolor{green!55} 0.338\,$\pm$\,0.070$^{\bullet}$ & \cellcolor{green!55} 0.835\,$\pm$\,0.042$^{\ddagger\P\bullet\circ\diamond\triangle}$ & \cellcolor{green!55} 1.117\,$\pm$\,0.019$^{\ddagger\S\P\star\bullet\circ\diamond\triangle}$ & \cellcolor{green!55} \textbf{0.523}\,$\pm$\,0.348$^{\ddagger\S\P\star\bullet\circ\diamond\triangle}$ \\
  \textsc{FineTune} & \cellcolor{green!8} 0.600\,$\pm$\,0.095 & \cellcolor{green!16} 0.398\,$\pm$\,0.014$^{\diamond}$ & \cellcolor{green!12} 0.157\,$\pm$\,0.017 & \cellcolor{green!27} 0.389\,$\pm$\,0.033 & \cellcolor{green!0} 1.434\,$\pm$\,0.074 & \cellcolor{green!8} 1.539\,$\pm$\,0.063$^{\bullet\diamond}$ & \cellcolor{green!1} \textbf{0.753}\,$\pm$\,0.547 \\
  \textsc{NewOnly} & \cellcolor{green!44} 0.445\,$\pm$\,0.037$^{\ddagger\bullet\circ\diamond\triangle}$ & \cellcolor{green!23} 0.383\,$\pm$\,0.033 & \cellcolor{green!48} 0.136\,$\pm$\,0.007$^{\bullet}$ & \cellcolor{green!35} 0.374\,$\pm$\,0.018 & \cellcolor{green!50} 0.888\,$\pm$\,0.056$^{\ddagger\P\bullet\circ\diamond\triangle}$ & \cellcolor{green!29} 1.357\,$\pm$\,0.118$^{\bullet\circ\diamond\triangle}$ & \cellcolor{green!38} \textbf{0.597}\,$\pm$\,0.417$^{\ddagger\P\bullet\circ\diamond\triangle}$ \\
  \textsc{\textbf{HybridAL}} ($\Delta\alpha$) & \cellcolor{green!19} 0.554\,$\pm$\,0.137 & \cellcolor{green!12} 0.405\,$\pm$\,0.026 & \cellcolor{green!24} 0.150\,$\pm$\,0.017 & \cellcolor{green!25} 0.393\,$\pm$\,0.052 & \cellcolor{green!12} 1.300\,$\pm$\,0.110 & \cellcolor{green!17} 1.460\,$\pm$\,0.206 & \cellcolor{green!11} \textbf{0.710}\,$\pm$\,0.509 \\
  \textsc{\textbf{HybridAL}} ($\Delta$Acc) & \cellcolor{green!44} 0.447\,$\pm$\,0.088$^{\ddagger\circ\triangle}$ & \cellcolor{green!0} 0.430\,$\pm$\,0.046 & \cellcolor{green!17} 0.154\,$\pm$\,0.012 & \cellcolor{green!0} 0.438\,$\pm$\,0.080 & \cellcolor{green!55} 0.835\,$\pm$\,0.042$^{\ddagger\P\bullet\circ\diamond\triangle}$ & \cellcolor{green!23} 1.403\,$\pm$\,0.189 & \cellcolor{green!33} \textbf{0.618}\,$\pm$\,0.419$^{\ddagger\P\bullet\circ\diamond\triangle}$ \\
  \textsc{FixedSwitch} (3) & \cellcolor{green!26} 0.525\,$\pm$\,0.037$^{\circ}$ & \cellcolor{green!9} 0.411\,$\pm$\,0.049 & \cellcolor{green!0} 0.164\,$\pm$\,0.015 & \cellcolor{green!12} 0.415\,$\pm$\,0.065 & \cellcolor{green!8} 1.345\,$\pm$\,0.117 & \cellcolor{green!3} 1.586\,$\pm$\,0.060 & \cellcolor{green!4} \textbf{0.741}\,$\pm$\,0.541 \\
  \textsc{FixedSwitch} (5) & \cellcolor{green!0} 0.635\,$\pm$\,0.091 & \cellcolor{green!7} 0.416\,$\pm$\,0.027 & \cellcolor{green!10} 0.158\,$\pm$\,0.014 & \cellcolor{green!28} 0.387\,$\pm$\,0.046 & \cellcolor{green!9} 1.335\,$\pm$\,0.097 & \cellcolor{green!7} 1.550\,$\pm$\,0.046 & \cellcolor{green!3} \textbf{0.747}\,$\pm$\,0.527 \\
  \textsc{FixedSwitch} (7) & \cellcolor{green!25} 0.527\,$\pm$\,0.088$^{\circ}$ & \cellcolor{green!1} 0.427\,$\pm$\,0.028 & \cellcolor{green!24} 0.150\,$\pm$\,0.009 & \cellcolor{green!21} 0.400\,$\pm$\,0.052 & \cellcolor{green!1} 1.426\,$\pm$\,0.153 & \cellcolor{green!0} 1.616\,$\pm$\,0.070 & \cellcolor{green!0} \textbf{0.758}\,$\pm$\,0.568 \\
  \textsc{FixedSwitch} (10) & \cellcolor{green!11} 0.589\,$\pm$\,0.049 & \cellcolor{green!13} 0.403\,$\pm$\,0.020 & \cellcolor{green!7} 0.159\,$\pm$\,0.022 & \cellcolor{green!32} 0.379\,$\pm$\,0.058 & \cellcolor{green!14} 1.278\,$\pm$\,0.076$^{\ddagger}$ & \cellcolor{green!7} 1.553\,$\pm$\,0.077 & \cellcolor{green!7} \textbf{0.727}\,$\pm$\,0.520 \\
  \midrule
  \multicolumn{8}{l}{\textbf{RoBERTa}} \\
  \midrule
  \textsc{Retrain} & \cellcolor{green!52} 0.366\,$\pm$\,0.099$^{\ddagger\bullet\circ}$ & \cellcolor{green!55} 0.285\,$\pm$\,0.015$^{\ddagger\S\P\star\bullet\circ\diamond\triangle}$ & \cellcolor{green!55} 0.127\,$\pm$\,0.016$^{\ddagger\star\circ\triangle}$ & \cellcolor{green!55} 0.290\,$\pm$\,0.043$^{\bullet}$ & \cellcolor{green!55} 0.823\,$\pm$\,0.070$^{\ddagger\bullet\circ\diamond\triangle}$ & \cellcolor{green!55} 1.094\,$\pm$\,0.018$^{\ddagger\P\bullet\circ\diamond\triangle}$ & \cellcolor{green!55} \textbf{0.498}\,$\pm$\,0.352$^{\ddagger\S\P\star\bullet\circ\diamond\triangle}$ \\
  \textsc{FineTune} & \cellcolor{green!11} 0.490\,$\pm$\,0.071 & \cellcolor{green!0} 0.439\,$\pm$\,0.042 & \cellcolor{green!17} 0.163\,$\pm$\,0.022 & \cellcolor{green!24} 0.331\,$\pm$\,0.056 & \cellcolor{green!12} 1.370\,$\pm$\,0.115 & \cellcolor{green!2} 1.610\,$\pm$\,0.107 & \cellcolor{green!6} \textbf{0.734}\,$\pm$\,0.563 \\
  \textsc{NewOnly} & \cellcolor{green!55} 0.358\,$\pm$\,0.036$^{\ddagger\bullet\circ\diamond\triangle}$ & \cellcolor{green!40} 0.327\,$\pm$\,0.009$^{\ddagger\star\bullet\circ\diamond\triangle}$ & \cellcolor{green!52} 0.130\,$\pm$\,0.009$^{\star\bullet\circ\triangle}$ & \cellcolor{green!27} 0.328\,$\pm$\,0.032 & \cellcolor{green!52} 0.867\,$\pm$\,0.088$^{\ddagger\bullet\circ\diamond\triangle}$ & \cellcolor{green!50} 1.145\,$\pm$\,0.066$^{\ddagger\P\bullet\circ\diamond\triangle}$ & \cellcolor{green!49} \textbf{0.526}\,$\pm$\,0.365$^{\ddagger\P\star\bullet\circ\diamond\triangle}$ \\
  \textsc{\textbf{HybridAL}} ($\Delta\alpha$) & \cellcolor{green!47} 0.381\,$\pm$\,0.093$^{\bullet}$ & \cellcolor{green!25} 0.369\,$\pm$\,0.046$^{\ddagger}$ & \cellcolor{green!51} 0.131\,$\pm$\,0.018$^{\star\triangle}$ & \cellcolor{green!38} 0.313\,$\pm$\,0.039$^{\bullet}$ & \cellcolor{green!24} 1.225\,$\pm$\,0.318 & \cellcolor{green!11} 1.523\,$\pm$\,0.078 & \cellcolor{green!22} \textbf{0.657}\,$\pm$\,0.545$^{\ddagger\bullet\circ}$ \\
  \textsc{\textbf{HybridAL}} ($\Delta$Acc) & \cellcolor{green!42} 0.396\,$\pm$\,0.080$^{\circ}$ & \cellcolor{green!10} 0.410\,$\pm$\,0.044 & \cellcolor{green!18} 0.161\,$\pm$\,0.017 & \cellcolor{green!38} 0.312\,$\pm$\,0.043 & \cellcolor{green!35} 1.085\,$\pm$\,0.334$^{\bullet}$ & \cellcolor{green!39} 1.247\,$\pm$\,0.210$^{\ddagger\P\bullet\circ\diamond\triangle}$ & \cellcolor{green!33} \textbf{0.602}\,$\pm$\,0.443$^{\ddagger\bullet\circ\diamond\triangle}$ \\
  \textsc{FixedSwitch} (3) & \cellcolor{green!18} 0.471\,$\pm$\,0.044 & \cellcolor{green!12} 0.404\,$\pm$\,0.041 & \cellcolor{green!0} 0.178\,$\pm$\,0.028 & \cellcolor{green!0} 0.363\,$\pm$\,0.031 & \cellcolor{green!0} 1.528\,$\pm$\,0.202 & \cellcolor{green!0} 1.631\,$\pm$\,0.098 & \cellcolor{green!0} \textbf{0.763}\,$\pm$\,0.602 \\
  \textsc{FixedSwitch} (5) & \cellcolor{green!0} 0.523\,$\pm$\,0.069 & \cellcolor{green!9} 0.413\,$\pm$\,0.027 & \cellcolor{green!22} 0.157\,$\pm$\,0.019 & \cellcolor{green!29} 0.325\,$\pm$\,0.043 & \cellcolor{green!5} 1.469\,$\pm$\,0.069 & \cellcolor{green!7} 1.562\,$\pm$\,0.064 & \cellcolor{green!4} \textbf{0.742}\,$\pm$\,0.570 \\
  \textsc{FixedSwitch} (7) & \cellcolor{green!16} 0.474\,$\pm$\,0.039 & \cellcolor{green!13} 0.402\,$\pm$\,0.019 & \cellcolor{green!13} 0.166\,$\pm$\,0.037 & \cellcolor{green!21} 0.335\,$\pm$\,0.038 & \cellcolor{green!19} 1.287\,$\pm$\,0.130$^{\circ}$ & \cellcolor{green!5} 1.586\,$\pm$\,0.080 & \cellcolor{green!11} \textbf{0.708}\,$\pm$\,0.543$^{\bullet\circ}$ \\
  \textsc{FixedSwitch} (10) & \cellcolor{green!29} 0.437\,$\pm$\,0.056$^{\ddagger\circ}$ & \cellcolor{green!20} 0.382\,$\pm$\,0.039 & \cellcolor{green!11} 0.167\,$\pm$\,0.014 & \cellcolor{green!19} 0.338\,$\pm$\,0.045 & \cellcolor{green!17} 1.305\,$\pm$\,0.099$^{\circ}$ & \cellcolor{green!7} 1.560\,$\pm$\,0.040 & \cellcolor{green!13} \textbf{0.698}\,$\pm$\,0.542$^{\ddagger\bullet\circ}$ \\
  \bottomrule
  \end{tabular}%
  }
  \caption{Full per-(method $\times$ backbone $\times$ dataset) test \textbf{NLL} (lower is better). Each cell reports mean $\pm$ std across 5 seeds. Cell shading is per-column within each backbone (greener $=$ lower $=$ better). The \textbf{Mean} column averages across the 30 (dataset, seed) cells per (method, backbone). Superscripts indicate that the cell is significantly lower than the reference method (paired $t$-test, $p<0.05$, two-sided): $\dagger$ \textsc{Retrain}, $\ddagger$ \textsc{FineTune}, $\S$ \textsc{NewOnly}, $\P$ \textsc{\textbf{HybridAL}} ($\Delta\alpha$), $\star$ \textsc{\textbf{HybridAL}} ($\Delta$Acc), $\bullet$ \textsc{FixedSwitch} (3), $\circ$ \textsc{FixedSwitch} (5), $\diamond$ \textsc{FixedSwitch} (7), $\triangle$ \textsc{FixedSwitch} (10).}
  \label{tab:appendix_nll_full}
\end{table*}

\begin{table*}[t]
  \centering
  \setlength{\tabcolsep}{3pt}
  \resizebox{\textwidth}{!}{%
  \begin{tabular}{l|ccccccc}
  \toprule
  Method & IMDb & AGNews & Jigsaw & SST-2 & TwtEv & Yahoo & \textbf{Mean} \\
  \midrule
  \multicolumn{8}{l}{\textbf{DistilBERT}} \\
  \midrule
  \textsc{Retrain} & \cellcolor{green!0} 526\,$\pm$\,19 & \cellcolor{green!0} 690\,$\pm$\,33 & \cellcolor{green!0} 719\,$\pm$\,41 & \cellcolor{green!0} 834\,$\pm$\,39 & \cellcolor{green!0} 886\,$\pm$\,22 & \cellcolor{green!0} 1375\,$\pm$\,44 & \cellcolor{green!0} \textbf{838}\,$\pm$\,272 \\
  \textsc{FineTune} & \cellcolor{green!23} 367\,$\pm$\,11$^{\dagger\P\star\diamond\triangle}$ & \cellcolor{green!42} 407\,$\pm$\,17$^{\dagger\P\star\circ\diamond\triangle}$ & \cellcolor{green!41} 459\,$\pm$\,12$^{\dagger\star\diamond\triangle}$ & \cellcolor{green!16} 621\,$\pm$\,14$^{\dagger\P\star\circ\diamond\triangle}$ & \cellcolor{green!22} 564\,$\pm$\,13$^{\dagger\star\circ\diamond\triangle}$ & \cellcolor{green!32} 675\,$\pm$\,19$^{\dagger\P\star\circ\diamond\triangle}$ & \cellcolor{green!28} \textbf{515}\,$\pm$\,115$^{\dagger\P\star\circ\diamond\triangle}$ \\
  \textsc{NewOnly} & \cellcolor{green!55} 153\,$\pm$\,6$^{\dagger\ddagger\P\star\bullet\circ\diamond\triangle}$ & \cellcolor{green!55} 324\,$\pm$\,15$^{\dagger\ddagger\P\star\bullet\circ\diamond\triangle}$ & \cellcolor{green!55} 374\,$\pm$\,13$^{\dagger\ddagger\P\star\bullet\circ\diamond\triangle}$ & \cellcolor{green!55} 122\,$\pm$\,3$^{\dagger\ddagger\P\star\bullet\circ\diamond\triangle}$ & \cellcolor{green!55} 94\,$\pm$\,7$^{\dagger\ddagger\P\star\bullet\circ\diamond\triangle}$ & \cellcolor{green!55} 162\,$\pm$\,12$^{\dagger\ddagger\P\star\bullet\circ\diamond\triangle}$ & \cellcolor{green!55} \textbf{205}\,$\pm$\,107$^{\dagger\ddagger\P\star\bullet\circ\diamond\triangle}$ \\
  \textsc{\textbf{HybridAL}} ($\Delta\alpha$) & \cellcolor{green!7} 478\,$\pm$\,84 & \cellcolor{green!19} 564\,$\pm$\,102 & \cellcolor{green!23} 572\,$\pm$\,115$^{\dagger}$ & \cellcolor{green!4} 780\,$\pm$\,97 & \cellcolor{green!14} 687\,$\pm$\,126$^{\dagger}$ & \cellcolor{green!14} 1061\,$\pm$\,258$^{\dagger}$ & \cellcolor{green!13} \textbf{690}\,$\pm$\,234$^{\dagger}$ \\
  \textsc{\textbf{HybridAL}} ($\Delta$Acc) & \cellcolor{green!2} 510\,$\pm$\,70 & \cellcolor{green!27} 512\,$\pm$\,42$^{\dagger}$ & \cellcolor{green!31} 523\,$\pm$\,21$^{\dagger}$ & \cellcolor{green!8} 727\,$\pm$\,84$^{\dagger}$ & \cellcolor{green!8} 765\,$\pm$\,102 & \cellcolor{green!17} 1000\,$\pm$\,189$^{\dagger}$ & \cellcolor{green!14} \textbf{673}\,$\pm$\,204$^{\dagger}$ \\
  \textsc{FixedSwitch} (3) & \cellcolor{green!24} 365\,$\pm$\,11$^{\dagger\P\star\diamond\triangle}$ & \cellcolor{green!41} 420\,$\pm$\,18$^{\dagger\P\star\circ\diamond\triangle}$ & \cellcolor{green!42} 457\,$\pm$\,9$^{\dagger\star\diamond\triangle}$ & \cellcolor{green!17} 617\,$\pm$\,13$^{\dagger\P\star\circ\diamond\triangle}$ & \cellcolor{green!22} 567\,$\pm$\,9$^{\dagger\star\circ\diamond\triangle}$ & \cellcolor{green!32} 674\,$\pm$\,30$^{\dagger\P\star\circ\diamond\triangle}$ & \cellcolor{green!28} \textbf{517}\,$\pm$\,113$^{\dagger\P\star\circ\diamond\triangle}$ \\
  \textsc{FixedSwitch} (5) & \cellcolor{green!23} 370\,$\pm$\,8$^{\dagger\P\star\diamond\triangle}$ & \cellcolor{green!36} 452\,$\pm$\,13$^{\dagger\star\diamond\triangle}$ & \cellcolor{green!39} 476\,$\pm$\,15$^{\dagger\star\diamond\triangle}$ & \cellcolor{green!15} 644\,$\pm$\,8$^{\dagger\P\triangle}$ & \cellcolor{green!20} 601\,$\pm$\,19$^{\dagger\star\diamond\triangle}$ & \cellcolor{green!29} 746\,$\pm$\,30$^{\dagger\P\star\diamond\triangle}$ & \cellcolor{green!25} \textbf{548}\,$\pm$\,130$^{\dagger\P\star\diamond\triangle}$ \\
  \textsc{FixedSwitch} (7) & \cellcolor{green!16} 415\,$\pm$\,10$^{\dagger\star}$ & \cellcolor{green!30} 491\,$\pm$\,21$^{\dagger}$ & \cellcolor{green!33} 512\,$\pm$\,25$^{\dagger\triangle}$ & \cellcolor{green!14} 656\,$\pm$\,16$^{\dagger\P\triangle}$ & \cellcolor{green!19} 607\,$\pm$\,15$^{\dagger\star\triangle}$ & \cellcolor{green!27} 777\,$\pm$\,26$^{\dagger\triangle}$ & \cellcolor{green!23} \textbf{576}\,$\pm$\,122$^{\dagger\P\star\triangle}$ \\
  \textsc{FixedSwitch} (10) & \cellcolor{green!16} 415\,$\pm$\,16$^{\dagger\star}$ & \cellcolor{green!28} 505\,$\pm$\,16$^{\dagger}$ & \cellcolor{green!29} 538\,$\pm$\,9$^{\dagger}$ & \cellcolor{green!12} 680\,$\pm$\,3$^{\dagger}$ & \cellcolor{green!16} 662\,$\pm$\,18$^{\dagger}$ & \cellcolor{green!23} 866\,$\pm$\,13$^{\dagger}$ & \cellcolor{green!20} \textbf{611}\,$\pm$\,149$^{\dagger\P\star}$ \\
  \midrule
  \multicolumn{8}{l}{\textbf{BERT}} \\
  \midrule
  \textsc{Retrain} & \cellcolor{green!0} 960\,$\pm$\,51 & \cellcolor{green!0} 1442\,$\pm$\,155 & \cellcolor{green!0} 1296\,$\pm$\,61 & \cellcolor{green!0} 1560\,$\pm$\,57 & \cellcolor{green!0} 1581\,$\pm$\,62 & \cellcolor{green!0} 2752\,$\pm$\,185 & \cellcolor{green!0} \textbf{1599}\,$\pm$\,574 \\
  \textsc{FineTune} & \cellcolor{green!24} 651\,$\pm$\,22$^{\dagger\star\diamond\triangle}$ & \cellcolor{green!40} 774\,$\pm$\,28$^{\dagger\star\bullet\circ\diamond\triangle}$ & \cellcolor{green!39} 820\,$\pm$\,33$^{\dagger\P\star\bullet\circ\diamond\triangle}$ & \cellcolor{green!18} 1127\,$\pm$\,15$^{\dagger}$ & \cellcolor{green!21} 1042\,$\pm$\,32$^{\dagger\star}$ & \cellcolor{green!34} 1202\,$\pm$\,24$^{\dagger}$ & \cellcolor{green!29} \textbf{936}\,$\pm$\,205$^{\dagger\star}$ \\
  \textsc{NewOnly} & \cellcolor{green!55} 242\,$\pm$\,10$^{\dagger\ddagger\P\star\bullet\circ\diamond\triangle}$ & \cellcolor{green!55} 528\,$\pm$\,28$^{\dagger\ddagger\P\star\bullet\circ\diamond\triangle}$ & \cellcolor{green!55} 621\,$\pm$\,28$^{\dagger\ddagger\P\star\bullet\circ\diamond\triangle}$ & \cellcolor{green!55} 236\,$\pm$\,22$^{\dagger\ddagger\P\star\bullet\circ\diamond\triangle}$ & \cellcolor{green!55} 168\,$\pm$\,25$^{\dagger\ddagger\P\star\bullet\circ\diamond\triangle}$ & \cellcolor{green!55} 269\,$\pm$\,14$^{\dagger\ddagger\star\bullet\circ\diamond\triangle}$ & \cellcolor{green!55} \textbf{344}\,$\pm$\,172$^{\dagger\ddagger\P\star\bullet\circ\diamond\triangle}$ \\
  \textsc{\textbf{HybridAL}} ($\Delta\alpha$) & \cellcolor{green!15} 760\,$\pm$\,98$^{\dagger\star}$ & \cellcolor{green!27} 994\,$\pm$\,200$^{\dagger}$ & \cellcolor{green!30} 926\,$\pm$\,48$^{\dagger}$ & \cellcolor{green!39} 621\,$\pm$\,73$^{\dagger\ddagger}$ & \cellcolor{green!40} 549\,$\pm$\,32$^{\dagger\ddagger\star}$ & \cellcolor{green!37} 1066\,$\pm$\,821$^{\dagger}$ & \cellcolor{green!34} \textbf{819}\,$\pm$\,372$^{\dagger\star}$ \\
  \textsc{\textbf{HybridAL}} ($\Delta$Acc) & \cellcolor{green!1} 947\,$\pm$\,55$^{\dagger}$ & \cellcolor{green!28} 972\,$\pm$\,69$^{\dagger}$ & \cellcolor{green!27} 964\,$\pm$\,60$^{\dagger}$ & \cellcolor{green!24} 994\,$\pm$\,570 & \cellcolor{green!0} 1584\,$\pm$\,60 & \cellcolor{green!37} 1062\,$\pm$\,190$^{\dagger}$ & \cellcolor{green!22} \textbf{1087}\,$\pm$\,323$^{\dagger}$ \\
  \textsc{FixedSwitch} (3) & \cellcolor{green!26} 625\,$\pm$\,21$^{\dagger\P\star\diamond\triangle}$ & \cellcolor{green!38} 813\,$\pm$\,33$^{\dagger\star\diamond\triangle}$ & \cellcolor{green!36} 858\,$\pm$\,28$^{\dagger\star\diamond\triangle}$ & \cellcolor{green!42} 542\,$\pm$\,8$^{\dagger\ddagger\diamond\triangle}$ & \cellcolor{green!42} 505\,$\pm$\,11$^{\dagger\ddagger\P\star\circ\triangle}$ & \cellcolor{green!48} 584\,$\pm$\,34$^{\dagger\ddagger\star\circ\diamond\triangle}$ & \cellcolor{green!41} \textbf{654}\,$\pm$\,138$^{\dagger\ddagger\P\star\circ\diamond\triangle}$ \\
  \textsc{FixedSwitch} (5) & \cellcolor{green!23} 658\,$\pm$\,21$^{\dagger\star\diamond\triangle}$ & \cellcolor{green!36} 837\,$\pm$\,28$^{\dagger\star\diamond\triangle}$ & \cellcolor{green!34} 883\,$\pm$\,23$^{\dagger\triangle}$ & \cellcolor{green!42} 553\,$\pm$\,29$^{\dagger\ddagger\triangle}$ & \cellcolor{green!40} 562\,$\pm$\,6$^{\dagger\ddagger\star}$ & \cellcolor{green!46} 682\,$\pm$\,16$^{\dagger\ddagger\star\triangle}$ & \cellcolor{green!40} \textbf{696}\,$\pm$\,129$^{\dagger\ddagger\star\diamond\triangle}$ \\
  \textsc{FixedSwitch} (7) & \cellcolor{green!20} 697\,$\pm$\,4$^{\dagger\star\triangle}$ & \cellcolor{green!30} 948\,$\pm$\,41$^{\dagger}$ & \cellcolor{green!30} 924\,$\pm$\,36$^{\dagger}$ & \cellcolor{green!41} 562\,$\pm$\,8$^{\dagger\ddagger\triangle}$ & \cellcolor{green!41} 526\,$\pm$\,23$^{\dagger\ddagger\P\star\circ\triangle}$ & \cellcolor{green!45} 711\,$\pm$\,28$^{\dagger\ddagger\star\triangle}$ & \cellcolor{green!38} \textbf{728}\,$\pm$\,166$^{\dagger\ddagger\star\triangle}$ \\
  \textsc{FixedSwitch} (10) & \cellcolor{green!18} 724\,$\pm$\,9$^{\dagger\star}$ & \cellcolor{green!27} 987\,$\pm$\,87$^{\dagger}$ & \cellcolor{green!27} 965\,$\pm$\,13$^{\dagger}$ & \cellcolor{green!39} 618\,$\pm$\,19$^{\dagger\ddagger}$ & \cellcolor{green!39} 574\,$\pm$\,14$^{\dagger\ddagger\star}$ & \cellcolor{green!43} 815\,$\pm$\,35$^{\dagger\ddagger}$ & \cellcolor{green!36} \textbf{781}\,$\pm$\,165$^{\dagger\ddagger\star}$ \\
  \midrule
  \multicolumn{8}{l}{\textbf{RoBERTa}} \\
  \midrule
  \textsc{Retrain} & \cellcolor{green!0} 950\,$\pm$\,59 & \cellcolor{green!0} 1072\,$\pm$\,42 & \cellcolor{green!0} 1176\,$\pm$\,82 & \cellcolor{green!0} 749\,$\pm$\,17 & \cellcolor{green!0} 693\,$\pm$\,28 & \cellcolor{green!0} 873\,$\pm$\,39 & \cellcolor{green!0} \textbf{919}\,$\pm$\,178 \\
  \textsc{FineTune} & \cellcolor{green!25} 620\,$\pm$\,20$^{\dagger\star\circ\diamond\triangle}$ & \cellcolor{green!32} 704\,$\pm$\,20$^{\dagger\P\star\circ\diamond\triangle}$ & \cellcolor{green!29} 838\,$\pm$\,46$^{\dagger\P\triangle}$ & \cellcolor{green!22} 535\,$\pm$\,16$^{\dagger\star\circ\diamond\triangle}$ & \cellcolor{green!20} 491\,$\pm$\,21$^{\dagger\star\circ\diamond\triangle}$ & \cellcolor{green!28} 529\,$\pm$\,14$^{\dagger\P\star\circ\diamond\triangle}$ & \cellcolor{green!26} \textbf{619}\,$\pm$\,124$^{\dagger\P\star\bullet\circ\diamond\triangle}$ \\
  \textsc{NewOnly} & \cellcolor{green!55} 219\,$\pm$\,9$^{\dagger\ddagger\P\star\bullet\circ\diamond\triangle}$ & \cellcolor{green!55} 432\,$\pm$\,9$^{\dagger\ddagger\P\star\bullet\circ\diamond\triangle}$ & \cellcolor{green!55} 538\,$\pm$\,40$^{\dagger\ddagger\P\star\bullet\circ\diamond\triangle}$ & \cellcolor{green!55} 203\,$\pm$\,10$^{\dagger\ddagger\P\star\bullet\circ\diamond\triangle}$ & \cellcolor{green!55} 138\,$\pm$\,4$^{\dagger\ddagger\P\star\bullet\circ\diamond\triangle}$ & \cellcolor{green!55} 204\,$\pm$\,18$^{\dagger\ddagger\P\star\bullet\circ\diamond\triangle}$ & \cellcolor{green!55} \textbf{289}\,$\pm$\,148$^{\dagger\ddagger\P\star\bullet\circ\diamond\triangle}$ \\
  \textsc{\textbf{HybridAL}} ($\Delta\alpha$) & \cellcolor{green!9} 836\,$\pm$\,175 & \cellcolor{green!7} 995\,$\pm$\,111 & \cellcolor{green!2} 1154\,$\pm$\,168 & \cellcolor{green!10} 654\,$\pm$\,104 & \cellcolor{green!11} 583\,$\pm$\,115 & \cellcolor{green!19} 637\,$\pm$\,44$^{\dagger\star}$ & \cellcolor{green!10} \textbf{810}\,$\pm$\,241$^{\dagger}$ \\
  \textsc{\textbf{HybridAL}} ($\Delta$Acc) & \cellcolor{green!11} 803\,$\pm$\,110$^{\dagger}$ & \cellcolor{green!21} 823\,$\pm$\,70$^{\dagger\P}$ & \cellcolor{green!23} 909\,$\pm$\,34$^{\dagger\P\triangle}$ & \cellcolor{green!5} 703\,$\pm$\,87 & \cellcolor{green!5} 645\,$\pm$\,80 & \cellcolor{green!5} 815\,$\pm$\,108 & \cellcolor{green!12} \textbf{783}\,$\pm$\,117$^{\dagger}$ \\
  \textsc{FixedSwitch} (3) & \cellcolor{green!24} 627\,$\pm$\,17$^{\dagger\star\circ\diamond\triangle}$ & \cellcolor{green!29} 736\,$\pm$\,47$^{\dagger\P\triangle}$ & \cellcolor{green!29} 840\,$\pm$\,29$^{\dagger\P\star\diamond\triangle}$ & \cellcolor{green!20} 549\,$\pm$\,23$^{\dagger\star\circ\diamond\triangle}$ & \cellcolor{green!20} 495\,$\pm$\,24$^{\dagger\star\circ\triangle}$ & \cellcolor{green!26} 561\,$\pm$\,33$^{\dagger\P\star\diamond\triangle}$ & \cellcolor{green!25} \textbf{635}\,$\pm$\,124$^{\dagger\P\star\circ\diamond\triangle}$ \\
  \textsc{FixedSwitch} (5) & \cellcolor{green!19} 696\,$\pm$\,25$^{\dagger\diamond}$ & \cellcolor{green!26} 768\,$\pm$\,27$^{\dagger\P\triangle}$ & \cellcolor{green!27} 860\,$\pm$\,39$^{\dagger\P\diamond\triangle}$ & \cellcolor{green!15} 602\,$\pm$\,25$^{\dagger\star}$ & \cellcolor{green!17} 524\,$\pm$\,14$^{\dagger\star}$ & \cellcolor{green!25} 567\,$\pm$\,13$^{\dagger\P\star\diamond\triangle}$ & \cellcolor{green!22} \textbf{669}\,$\pm$\,122$^{\dagger\P\star\diamond\triangle}$ \\
  \textsc{FixedSwitch} (7) & \cellcolor{green!18} 711\,$\pm$\,19$^{\dagger}$ & \cellcolor{green!24} 794\,$\pm$\,46$^{\dagger\P}$ & \cellcolor{green!24} 899\,$\pm$\,34$^{\dagger\P\triangle}$ & \cellcolor{green!16} 592\,$\pm$\,17$^{\dagger}$ & \cellcolor{green!17} 520\,$\pm$\,9$^{\dagger\star\triangle}$ & \cellcolor{green!19} 640\,$\pm$\,20$^{\dagger\star}$ & \cellcolor{green!20} \textbf{693}\,$\pm$\,131$^{\dagger\P\star\triangle}$ \\
  \textsc{FixedSwitch} (10) & \cellcolor{green!19} 699\,$\pm$\,16$^{\dagger}$ & \cellcolor{green!18} 862\,$\pm$\,53$^{\dagger}$ & \cellcolor{green!16} 993\,$\pm$\,35$^{\dagger}$ & \cellcolor{green!15} 596\,$\pm$\,16$^{\dagger\star}$ & \cellcolor{green!14} 552\,$\pm$\,22$^{\dagger\star}$ & \cellcolor{green!17} 668\,$\pm$\,13$^{\dagger\star}$ & \cellcolor{green!17} \textbf{728}\,$\pm$\,158$^{\dagger\P\star}$ \\
  \bottomrule
  \end{tabular}%
  }
  \caption{Full per-(method $\times$ backbone $\times$ dataset) \textbf{training time} in seconds (lower is better). Each cell reports mean $\pm$ std across 5 seeds. Cell shading is per-column within each backbone (greener $=$ lower $=$ better). The \textbf{Mean} column averages across the 30 (dataset, seed) cells per (method, backbone). Superscripts indicate that the cell is significantly lower than the reference method (paired $t$-test, $p<0.05$, two-sided): $\dagger$ \textsc{Retrain}, $\ddagger$ \textsc{FineTune}, $\S$ \textsc{NewOnly}, $\P$ \textsc{\textbf{HybridAL}} ($\Delta\alpha$), $\star$ \textsc{\textbf{HybridAL}} ($\Delta$Acc), $\bullet$ \textsc{FixedSwitch} (3), $\circ$ \textsc{FixedSwitch} (5), $\diamond$ \textsc{FixedSwitch} (7), $\triangle$ \textsc{FixedSwitch} (10).}
  \label{tab:appendix_time_full}
\end{table*}

\subsection{Post-Switch Signal Stability}
\label{app:post_switch}

\begin{figure}[t]
  \centering
  \subcaptionbox{Fast switch.\label{fig:ca_fast}}
    {\includegraphics[width=0.48\linewidth]{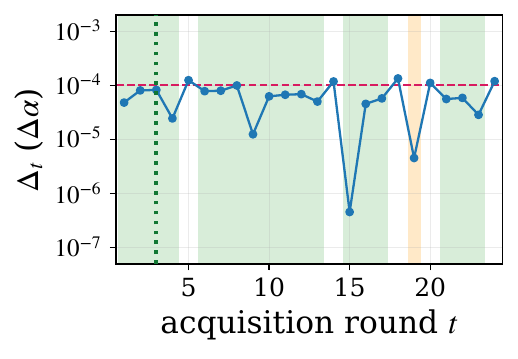}}\hfill
  \subcaptionbox{Slow switch.\label{fig:ca_slow}}
    {\includegraphics[width=0.48\linewidth]{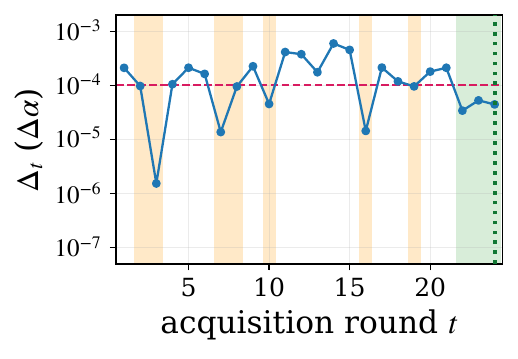}}
  \\[2pt]
  \subcaptionbox{Dips filtered by $k$.\label{fig:ca_dips}}
    {\includegraphics[width=0.48\linewidth]{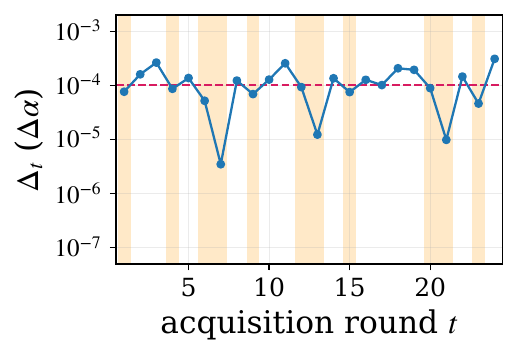}}\hfill
  \subcaptionbox{No switch.\label{fig:ca_none}}
    {\includegraphics[width=0.48\linewidth]{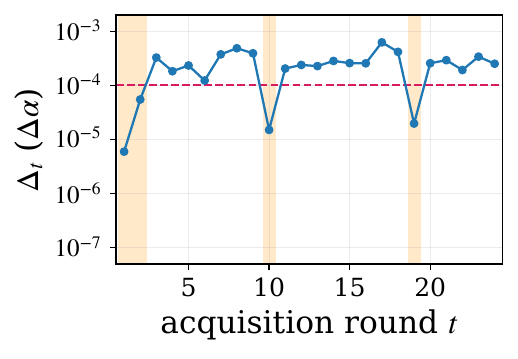}}
  \\[4pt]
  \includegraphics[width=\linewidth]{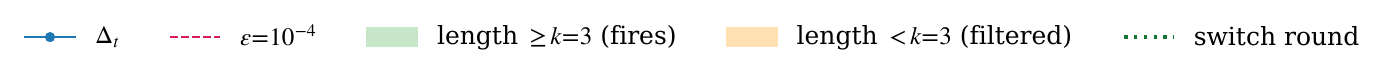}
  \caption{Per-round $\Delta\alpha$ trajectories
  ($\varepsilon{=}10^{-4}$, $k{=}3$) for four representative runs.
  Dashed line marks $\varepsilon$; the marker marks
  $t^{\star}$.}
  \label{fig:control_alpha}
\end{figure}

\begin{figure}[t]
  \centering
  \subcaptionbox{Fast switch.\label{fig:cc_fast}}
    {\includegraphics[width=0.48\linewidth]{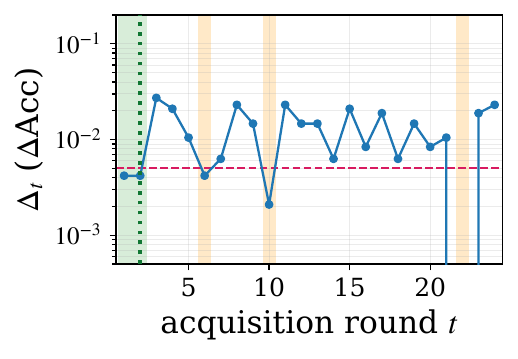}}\hfill
  \subcaptionbox{Slow switch.\label{fig:cc_slow}}
    {\includegraphics[width=0.48\linewidth]{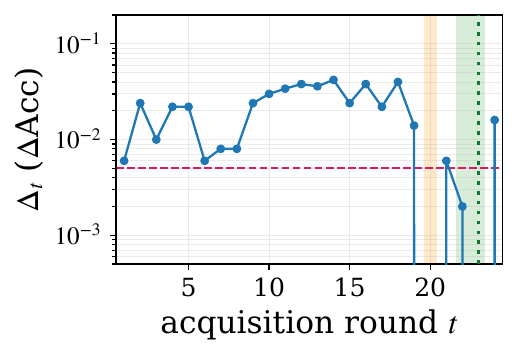}}
  \\[2pt]
  \subcaptionbox{Dips filtered by $k$.\label{fig:cc_dips}}
    {\includegraphics[width=0.48\linewidth]{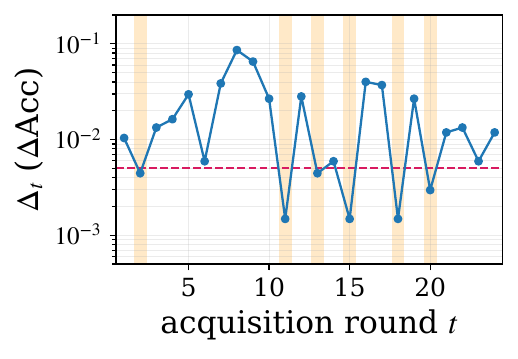}}\hfill
  \subcaptionbox{No switch.\label{fig:cc_none}}
    {\includegraphics[width=0.48\linewidth]{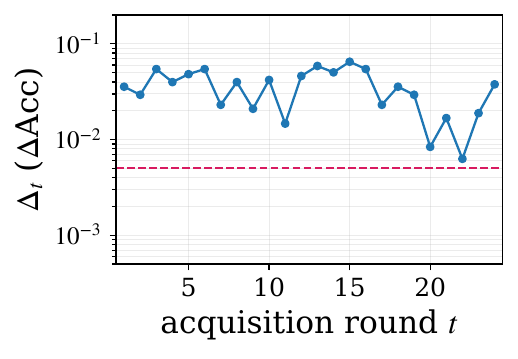}}
  \\[4pt]
  \includegraphics[width=\linewidth]{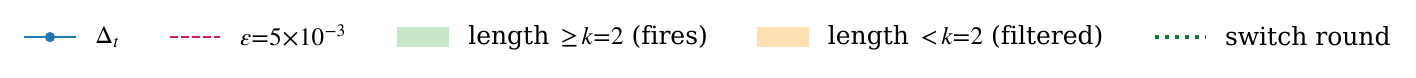}
  \caption{Per-round $\Delta$Acc trajectories
  ($\varepsilon{=}5{\times}10^{-3}$, $k{=}2$) for the same four
  cases as Figure~\ref{fig:control_alpha}.}
  \label{fig:control_acc}
\end{figure}

\begin{figure}[t]
  \centering
  \subcaptionbox{$\Delta\alpha$ ($k{=}3$).\label{fig:sh_alpha}}
    {\includegraphics[width=0.48\linewidth]{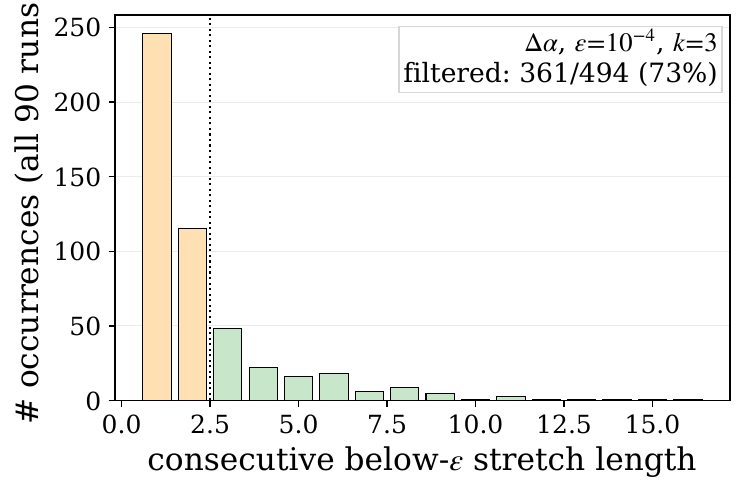}}\hfill
  \subcaptionbox{$\Delta$Acc ($k{=}2$).\label{fig:sh_acc}}
    {\includegraphics[width=0.48\linewidth]{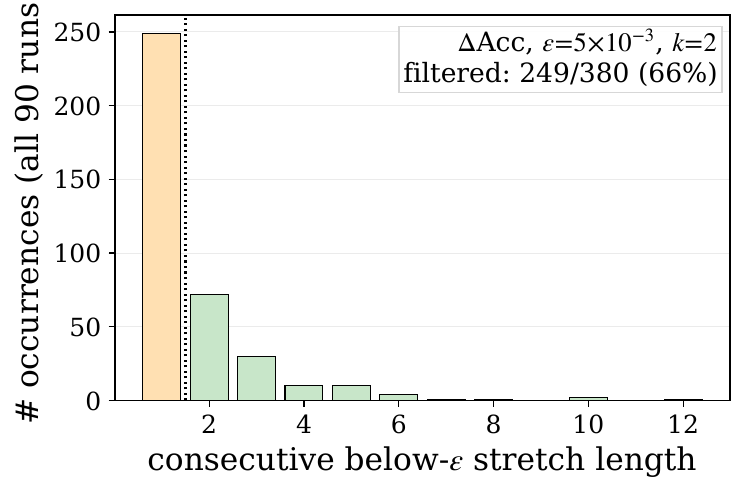}}
  \\[4pt]
  \includegraphics[width=\linewidth]{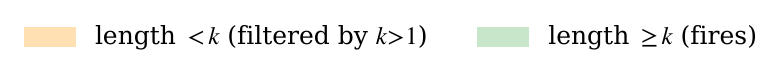}
 
  \caption{Distribution of below-$\varepsilon$ stretch lengths,
  $90$ runs per signal. Stretches shorter than $k$ are filtered
  by the patience parameter and do not trigger a switch.}
  \label{fig:stretch_hist}
\end{figure}

Tables~\ref{tab:post_switch_stability} and
\ref{tab:main_t_star_per_cell} report post-switch dynamics and
per-cell $t^{\star}$. After switching, both signals frequently
re-cross their thresholds ($31\%$ to $46\%$ of post-switch rounds
for $\Delta\alpha$; $36\%$ to $83\%$ for $\Delta$Acc), and
$t^{\star}$ itself spans $4.0$ to $23.0$ across cells. Endpoint
\fone{} is non-inferior (\S\ref{sec:f1_parity}), so the excursions
reflect post-switch optimization dynamics rather than failed
stabilization. A reversible variant would treat them as
instability and oscillate, breaking the monotonic cost
guarantee; hence the irreversibility in
Algorithm~\ref{alg:hybridal}.
Figures~\ref{fig:control_alpha} and~\ref{fig:control_acc} show
representative per-round trajectories with $\varepsilon$ and
$t^{\star}$ marked: a fast switcher, a slow switcher, a run whose
sub-threshold dips are filtered by the patience parameter, and a
run that never fires. Figure~\ref{fig:stretch_hist} aggregates this behaviour across all
$90$ runs per signal:
$73\%$ ($\Delta\alpha$) and $66\%$ ($\Delta$Acc) of
below-$\varepsilon$ stretches are shorter than $k$ and are
therefore filtered rather than triggering a switch.


\begin{table}[H]
  \centering
  \scriptsize
  \setlength{\tabcolsep}{3pt}
  \resizebox{\columnwidth}{!}{%
  \begin{tabular}{l|ccc|ccc}
  \toprule
  & \multicolumn{3}{c|}{$\Delta\alpha$ ($\varepsilon{=}10^{-4}$)}
  & \multicolumn{3}{c}{$\Delta$Acc ($\varepsilon{=}5{\times}10^{-3}$)} \\
  Dataset & $t^{\!\star}$ & max post & \%${>}\varepsilon$
          & $t^{\!\star}$ & max post & \%${>}\varepsilon$ \\
  \midrule
  IMDb      &  8.4 & 4.2e\!-\!4 & 43 & 17.8 & 3.8e\!-\!2 & 83 \\
  AG News   & 10.7 & 4.4e\!-\!4 & 46 &  8.0 & 2.5e\!-\!2 & 54 \\
  Jigsaw    &  6.9 & 2.9e\!-\!4 & 31 &  9.0 & 1.6e\!-\!2 & 36 \\
  SST-2     &  8.4 & 4.6e\!-\!4 & 33 & 10.8 & 3.9e\!-\!2 & 67 \\
  TweetEval &  6.3 & 4.5e\!-\!4 & 41 & 12.5 & 6.7e\!-\!2 & 80 \\
  Yahoo     & 10.7 & 4.6e\!-\!4 & 40 & 15.2 & 5.4e\!-\!2 & 79 \\
  \bottomrule
  \end{tabular}}
  \caption{Post-switch signal dynamics (3 backbones $\times$ 5
  seeds; switched cells only). $t^{\!\star}$: mean switch round
  averaged unweightedly across the 3 backbones; max post:
  largest post-switch signal value pooled across all switched
  cells; \%${>}\varepsilon$: fraction of post-switch rounds
  above threshold.}
  \label{tab:post_switch_stability}
\end{table}

\begin{table}[H]
  \centering
  \scriptsize
  \setlength{\tabcolsep}{2pt}
  \renewcommand{\arraystretch}{1.0}
  \resizebox{\columnwidth}{!}{%
  \begin{tabular}{l|cccc|cccc}
  \toprule
  & \multicolumn{4}{c|}{HybridAL ($\Delta\alpha$)} & \multicolumn{4}{c}{HybridAL ($\Delta$Acc)} \\
  Dataset & DistilBERT & BERT & RoBERTa & \textit{Mean} & DistilBERT & BERT & RoBERTa & \textit{Mean} \\
  \midrule
  IMDb      & 12.3 &  8.5 &  4.5 &  8.4 & 15.5 & 23.0 & 14.8 & 17.8 \\
  AG News   & 11.3 &  8.8 & 12.0 & 10.7 &  8.6 &  8.0 &  7.4 &  8.0 \\
  Jigsaw    & 11.2 &  5.4 &  4.0 &  6.9 &  7.4 &  9.6 & 10.0 &  9.0 \\
  SST-2     &  7.5 &  9.6 &  8.0 &  8.4 &  9.8 &  9.0 & 13.5 & 10.8 \\
  TweetEval &  7.5 &  6.6 &  4.7 &  6.3 & 14.0 & \ding{55} & 11.0 & 12.5 \\
  Yahoo     & 11.8 & 10.8 &  9.4 & 10.7 & 13.4 & 15.8 & 16.5 & 15.2 \\
  \midrule
  \textit{Mean} & 10.3 & 8.3 & 7.1 & 8.6 & 11.4 & 13.1 & 12.2 & 12.2 \\
  \bottomrule
  \end{tabular}}
    \caption{Mean empirical switch round $t^{\star}$ per
  (method, backbone, dataset), averaged across the seeds that
  fired. \ding{55}: no seed switched within $T{=}25$ rounds;
  such cells are excluded from the means.}
  \label{tab:main_t_star_per_cell}
\end{table}

\subsection{Calibration Mechanism: Temperature Analysis}
\label{app:temp_scaling}

To understand the mechanism behind the calibration differences
in \S\ref{sec:time_nll}, we fit a single temperature parameter
$\tau$ per run on the validation
set~\citep{guo2017calibration} and examine the fitted values
(Table~\ref{tab:appendix_ece_temp}).

The fitted temperatures reveal a clear pattern:
\textsc{Retrain} is intrinsically well-calibrated
($\tau{=}0.97$, requiring almost no correction), while
\textsc{FineTune} is severely overconfident ($\tau{=}2.29$).
Both \textsc{\textbf{HybridAL}} variants fall between these
extremes ($\tau{\approx}1.8$--$2.0$), consistent with the
design intent: retraining during the early high-variance
rounds produces less overconfident predictions than switching
to warm-starting immediately. The monotonic ordering
\textsc{Retrain} $<$ \textsc{\textbf{HybridAL}}($\Delta$Acc)
$<$ \textsc{\textbf{HybridAL}}($\Delta\alpha$) $<$
\textsc{FineTune} in both $\tau$ and raw NLL confirms that
later switching preserves more of \textsc{Retrain}'s
calibration, as predicted by the stabilization hypothesis
(\S\ref{subsec:stab}).

After temperature scaling, all pool-trained methods converge
to similar NLL ($0.517$--$0.533$) and ECE
($2.5$--$3.0\%$), confirming that they can reach the same
calibration ceiling given a held-out validation set for
post-hoc fitting. This is expected: all pool-trained methods
reach similar \fone{} (\S\ref{sec:f1_parity}), so their
learned representations carry similar discriminative
information; the differences lie in how well-scaled the
probabilities are \emph{during training}, when
uncertainty-based acquisition functions use them to select
examples. \textsc{NewOnly} is the exception: even after
scaling, it retains the highest NLL ($0.560$) and ECE
($3.3\%$), indicating that discarding historical data harms
representation quality, not just probability scaling. This
provides additional evidence for excluding \textsc{NewOnly}
from the substantive Pareto frontier
(\S\ref{sec:time_nll}). Acquisition in this work uses the
model's native probabilities, so probability quality during
training affects which examples are queried. Applying
per-round temperature scaling before acquisition is possible
in principle but would require refitting $\tau_t$ at every
round; we leave this controlled comparison to future work.
In low-resource AL settings where the validation set itself
is expensive, \textsc{\textbf{HybridAL}}'s intrinsic
calibration ($\tau{\approx}1.8$--$2.0$ vs.\
\textsc{FineTune}'s $2.29$) may be the only available option.

\begin{table}[H]
  \centering
  \scriptsize
  \setlength{\tabcolsep}{3pt}
  \resizebox{\columnwidth}{!}{%
  \begin{tabular}{l|c|cc|cc}
  \toprule
  Method & $\tau$ & NLL & NLL ($\tau$) & ECE & ECE ($\tau$) \\
  \midrule
  \textsc{Retrain}                   & 0.97 & 0.532 & 0.517 & 5.4\%  & 2.6\% \\
  \textsc{FineTune}                  & 2.29 & 0.777 & 0.533 & 13.7\% & 2.5\% \\
  \textsc{NewOnly}                   & 1.20 & 0.596 & 0.560 & 8.0\%  & 3.3\% \\
  \textsc{\textbf{HybridAL}} ($\Delta\alpha$) & 1.97 & 0.722 & 0.532 & 12.1\% & 2.8\% \\
  \textsc{\textbf{HybridAL}} ($\Delta$Acc)    & 1.80 & 0.682 & 0.525 & 11.8\% & 3.0\% \\
  \bottomrule
  \end{tabular}}
  \caption{Temperature scaling on DistilBERT (6 datasets
  $\times$ 5 seeds). $\tau$: fitted temperature; NLL($\tau$), ECE($\tau$):
  post-scaling values.}
  \label{tab:appendix_ece_temp}
\end{table}

\section{Validation-Label Assumptions and Size Sensitivity}
\label{app:val_size}

$\mathcal{V}$ is fixed, held out, and disjoint from
$\mathcal{L}_t$ and $\mathcal{U}_t$; it is shared by \emph{all}
methods for early stopping and per-round evaluation, so it is not
a cost specific to \textsc{\textbf{HybridAL}}. Its size ranges
from $477$ to $1{,}596$ labels
(Table~\ref{tab:datasets-class-distribution}), comparable to or
larger than the $1{,}000$-label acquisition budget. The two
signals differ in how they use it: $\Delta$Acc reads its value
from the early-stopping forward pass and adds no extra
computation, while $\Delta\alpha$ is computed from weight
matrices alone and depends on $\mathcal{V}$ only through early
stopping. Replacing early stopping with a fixed-epoch schedule
would remove that dependence entirely, making $\Delta\alpha$
validation-free, but would forfeit the time savings, which come
from \textsc{FineTune} converging in $3.4$ epochs vs.\
\textsc{Retrain}'s $5.5$ under early stopping
(\S\ref{app:early_stopping}).

Table~\ref{tab:val_size} reports a focused study of how much
validation data $\Delta$Acc actually needs. We subsample
$\mathcal{V}$ to $|\mathcal{V}_{\text{sig}}| \in \{25, 50, 100,
200, \text{full}\}$ labels on DistilBERT with IMDb and AG News,
3 seeds per cell, at the tuned
$(\varepsilon, k){=}(5{\times}10^{-3}, 2)$; subsampling is
class-stratified, performed once at initialisation, and
deterministic per seed. The subsampled set drives both early
stopping and the $\Delta$Acc signal, and endpoint \fone{} is
scored on the untouched test set. Endpoint \fone{} varies by at
most $0.007$ within each dataset across all sizes (spread
$0.0040$ on AG News, $0.0069$ on IMDb), inside the $0.013$
seed-noise floor of \S\ref{app:tost}, so a few dozen validation
labels suffice for the signal.

\begin{table}[H]
  \centering
  \small
  \setlength{\tabcolsep}{4pt}
  \resizebox{\columnwidth}{!}{%
  \begin{tabular}{l|c|ccc}
  \toprule
  Dataset & $|\mathcal{V}_{\text{sig}}|$ & Fire rate & $t^{\star}$ & Test \fone{} \\
  \midrule
  \multirow{5}{*}{AG News}
    & 25            & 100\% & $8.0$\,$\pm$\,$2.6$  & $0.9010$\,$\pm$\,$0.0024$ \\
    & 50            & 100\% & $10.0$\,$\pm$\,$8.9$ & $0.9044$\,$\pm$\,$0.0010$ \\
    & 100           & 67\%  & $10.5$\,$\pm$\,$9.2$ & $0.9022$\,$\pm$\,$0.0001$ \\
    & 200           & 33\%  & $11$                 & $0.9004$\,$\pm$\,$0.0061$ \\
    & full ($1{,}200$) & 100\% & $7.7$\,$\pm$\,$2.5$ & $0.9041$\,$\pm$\,$0.0065$ \\
  \midrule
  \multirow{5}{*}{IMDb}
    & 25            & 33\% & $11$ & $0.8236$\,$\pm$\,$0.0086$ \\
    & 50            & 33\% & $23$ & $0.8305$\,$\pm$\,$0.0037$ \\
    & 100           & 33\% & $17$ & $0.8237$\,$\pm$\,$0.0173$ \\
    & 200           & 0\%  & ---  & $0.8295$\,$\pm$\,$0.0032$ \\
    & full ($500$)  & 33\% & $24$ & $0.8262$\,$\pm$\,$0.0077$ \\
  \bottomrule
  \end{tabular}}
  \caption{Validation-size sensitivity for
  \textsc{\textbf{HybridAL}}($\Delta$Acc) on DistilBERT, 3 seeds
  per cell. $t^{\star}$ is averaged over firing seeds only;
  single values indicate one firing seed.}
  \label{tab:val_size}
\end{table}

Fire rate should not be read as a trend in this table. It is
non-monotonic in $|\mathcal{V}_{\text{sig}}|$, and at
$|\mathcal{V}_{\text{sig}}|{=}25$ validation accuracy is
quantised in steps of $1/25{=}0.04$, so $\Delta$Acc is either
exactly $0$ or at least $0.04$, well above
$\varepsilon{=}5{\times}10^{-3}$: sub-threshold movement is
numerically unobservable, and the signal can fire because every
reading below $0.04$ registers as zero rather than because the
trajectory has stabilised. IMDb's lower fire rate persists at
full $\mathcal{V}$ and matches its largest post-switch
re-crossing rate in
Table~\ref{tab:post_switch_stability}, so it is a property of the
dataset rather than of validation size.

\section{Post-Switch Batch Composition}
\label{app:batch_composition}

This appendix asks what changes about acquisition after the
switch. We log per-round selected indices, acquired-batch
entropies, and full validation logits for \textsc{Retrain},
\textsc{\textbf{HybridAL}}($\Delta\alpha$), and
\textsc{\textbf{HybridAL}}($\Delta$Acc) on DistilBERT with IMDb
and AG News, seeds $42$--$44$ (18 runs). This is a focused study
on two datasets and one backbone, not the full grid.

\subsection{Batch Overlap}
\label{app:jaccard}

Table~\ref{tab:jaccard} reports the Jaccard overlap between each
\textsc{\textbf{HybridAL}} run's acquired batch and that of the
paired \textsc{Retrain} run at the same round, pooled over the six
(dataset, seed) cells. Before the switch the two are identical by
construction, since both are retraining from the same
initialisation on the same pool. After the switch the overlap
collapses to under $0.02$: the model selects almost entirely
different examples. Endpoint \fone{} is nonetheless non-inferior
(\S\ref{app:tost}), so a different batch is not a worse batch at
this budget.

\begin{table}[H]
  \centering
  \small
  \setlength{\tabcolsep}{4pt}
  \resizebox{\columnwidth}{!}{%
  \begin{tabular}{l|cc}
  \toprule
  Comparison vs.\ \textsc{Retrain} & Pre-switch & Post-switch \\
  \midrule
  \textsc{\textbf{HybridAL}} ($\Delta\alpha$) & $1.0000$\,$\pm$\,$0.0000$ & $0.0174$\,$\pm$\,$0.0427$ \\
  \textsc{\textbf{HybridAL}} ($\Delta$Acc)    & $0.9992$\,$\pm$\,$0.0071$ & $0.0197$\,$\pm$\,$0.0515$ \\
  \bottomrule
  \end{tabular}}
  \caption{Jaccard overlap of acquired batches with the paired
  \textsc{Retrain} run, pooled over 6 (dataset, seed) cells.}
  \label{tab:jaccard}
\end{table}

\subsection{Class Balance}
\label{app:class_balance}

Table~\ref{tab:class_balance} gives the class distribution of
acquired batches before and after the switch. No class collapses:
the largest shift is $8.0$ pp on AG News \emph{Sci/Tech} for
$\Delta\alpha$, and the binary IMDb splits stay within $5$ pp of
even. Low overlap therefore does not come from the sampler
concentrating on a single class.

\begin{table}[H]
  \centering
  \small
  \setlength{\tabcolsep}{4pt}
  \resizebox{\columnwidth}{!}{%
  \begin{tabular}{ll|cc}
  \toprule
  Signal & Dataset & Pre-switch & Post-switch \\
  \midrule
  $\Delta\alpha$ & IMDb    & $(47.6, 52.4)$ & $(51.4, 48.6)$ \\
  $\Delta\alpha$ & AG News & $(29.0, 11.9, 33.1, 26.1)$ & $(25.0, 9.8, 31.1, 34.1)$ \\
  $\Delta$Acc    & IMDb    & $(52.5, 47.5)$ & $(51.5, 48.5)$ \\
  $\Delta$Acc    & AG News & $(30.6, 13.5, 29.2, 26.7)$ & $(24.6, 11.0, 32.8, 31.5)$ \\
  \bottomrule
  \end{tabular}}
  \caption{Class composition of acquired batches, in percent.
  IMDb: (negative, positive). AG News: (World, Sports, Business,
  Sci/Tech).}
  \label{tab:class_balance}
\end{table}

\subsection{Acquired-Batch Entropy}
\label{app:entropy}

Table~\ref{tab:batch_entropy} reports the mean predictive entropy
of the acquired batch. Post-switch entropy falls by $0.09$
($\Delta\alpha$) and $0.08$ ($\Delta$Acc) relative to pre-switch.
This is consistent with the sharper post-fine-tuning softmax
documented in \S\ref{app:temp_scaling} ($\tau$ up to $2.29$),
which compresses all entropies toward zero without reordering
them, so top-$n$ selection would be unaffected. We note that this
is an interpretation of the temperature evidence rather than a
direct measurement: we did not compute rank correlation between
pre- and post-switch entropy orderings.

\begin{table}[H]
  \centering
  \small
  \setlength{\tabcolsep}{4pt}
  \resizebox{\columnwidth}{!}{%
  \begin{tabular}{l|ccc}
  \toprule
  Method & Pre-switch & Post-switch & Change \\
  \midrule
  \textsc{\textbf{HybridAL}} ($\Delta\alpha$) & $0.8455$\,$\pm$\,$0.1665$ & $0.7547$\,$\pm$\,$0.0907$ & $-0.0909$ \\
  \textsc{\textbf{HybridAL}} ($\Delta$Acc)    & $0.8392$\,$\pm$\,$0.1835$ & $0.7624$\,$\pm$\,$0.0884$ & $-0.0768$ \\
  \textsc{Retrain}                            & $0.8419$\,$\pm$\,$0.1617$ & ---                        & --- \\
  \bottomrule
  \end{tabular}}
  \caption{Mean predictive entropy of the acquired batch.
  \textsc{Retrain} never switches, so its trajectory is reported
  as a single reference value.}
  \label{tab:batch_entropy}
\end{table}

\subsection{Per-Round Calibration}
\label{app:ece_trajectory}

Table~\ref{tab:ece_trajectory} tracks 15-bin validation ECE
through the AL loop. \textsc{\textbf{HybridAL}} matches
\textsc{Retrain} in the early rounds, where the pool is small and
each batch reshapes the labeled distribution most, and drifts
above it later: over the last five rounds the gap averages
$2.90$ pp for $\Delta\alpha$ and $5.06$ pp for $\Delta$Acc.
Endpoint \fone{} is unchanged over the same window
(\S\ref{app:tost}). Late calibration drift therefore does not
translate into a measurable endpoint cost here, and it is
correctable at inference by temperature scaling
(\S\ref{app:temp_scaling}); but the pattern also means these
analyses cannot support a claim that better calibration yields
better acquisition, which we do not make.

\begin{table}[H]
  \centering
  \small
  \setlength{\tabcolsep}{4pt}
  \resizebox{\columnwidth}{!}{%
  \begin{tabular}{c|ccc}
  \toprule
  Round & \textsc{Retrain} & $\Delta\alpha$ & $\Delta$Acc \\
  \midrule
  5  & $6.09$\,$\pm$\,$2.44$ & $5.90$\,$\pm$\,$2.47$ & $6.58$\,$\pm$\,$2.85$ \\
  10 & $5.28$\,$\pm$\,$2.37$ & $7.27$\,$\pm$\,$3.35$ & $7.60$\,$\pm$\,$4.23$ \\
  15 & $4.28$\,$\pm$\,$1.31$ & $6.43$\,$\pm$\,$2.91$ & $7.15$\,$\pm$\,$4.95$ \\
  20 & $5.43$\,$\pm$\,$4.01$ & $7.36$\,$\pm$\,$3.95$ & $9.87$\,$\pm$\,$4.27$ \\
  24 & $4.70$\,$\pm$\,$1.03$ & $8.47$\,$\pm$\,$4.01$ & $10.06$\,$\pm$\,$4.11$ \\
  \midrule
  Last 5 & $5.11$\,$\pm$\,$3.08$ & $8.01$\,$\pm$\,$3.76$ & $10.17$\,$\pm$\,$3.87$ \\
  \bottomrule
  \end{tabular}}
  \caption{Validation ECE (\%, 15-bin) by round, mean over 6
  (dataset, seed) cells.}
  \label{tab:ece_trajectory}
\end{table}

\section{Signal Ablation}
\label{app:signal_ablation}

\begin{figure}[t]
  \centering
  \includegraphics[width=\columnwidth]{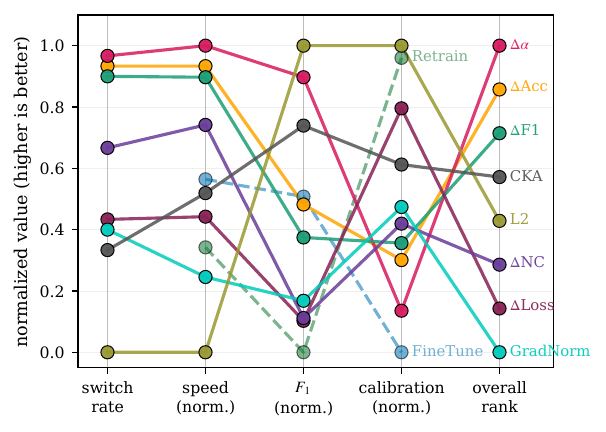}
  \caption{Parallel-coordinates profile of the eight switching
  signals on five normalised axes (higher~=~better); Retrain
  and \textsc{FineTune} (dashed) are references. $\Delta\alpha$ and
  $\Delta$Acc dominate on overall rank, motivating their
  selection as \textsc{\textbf{HybridAL}} variants.}
  \label{fig:signal_ablation_pareto}
\end{figure}

We evaluated eight candidate switching signals
(\S\ref{sec:setup}) on DistilBERT with the entropy sampler,
across all six datasets and $5$ seeds. Because the raw
signals have different units and magnitudes, we normalise
each per-round value by the signal's maximum over rounds
$2$--$5$ of a calibration run, producing unit-free values
that share a common threshold $\varepsilon{=}0.5$ with
patience $k{=}3$. Under this normalisation, a switch fires when the
round-to-round change $\Delta S_t$
(Definition~\ref{def:stabilization}) drops below
$\varepsilon{=}0.5$ for $k$ consecutive rounds. This shared configuration is
intended for ranking signals against each other; the absolute
time and \fone{} numbers are not directly comparable to the
per-signal tuned hyperparameters used in the main results
(Appendix~\ref{app:hyperparam_tuning}).
Table~\ref{tab:appendix_signal_ablation}
reports per-(signal $\times$ dataset) test \fone{}, test NLL,
total training time, and switching rate (fraction of seeds
where the signal triggered a switch within $T{=}25$ rounds);
Figure~\ref{fig:signal_ablation_pareto} summarises the same
data as a parallel-coordinates profile across five normalised
axes (switch rate, speed, \fone{}, calibration, overall rank).

\begin{figure}[t]
  \centering
  \includegraphics[width=\columnwidth]{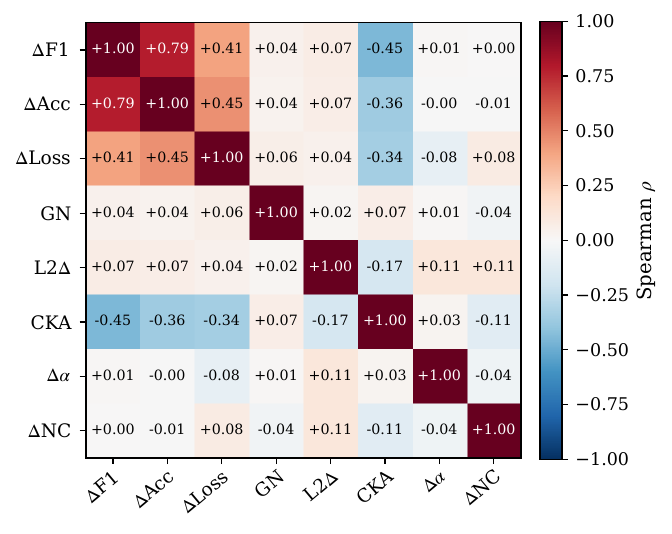}
  \caption{Spearman correlation between the eight candidate
  switching signals, $n{=}720$ (run $\times$ round)
  observations from \textsc{Retrain} DistilBERT runs. Performance
  signals ($\Delta$F1, $\Delta$Acc, $\Delta$Loss) cluster at
  $\rho{\leq}0.79$; $\Delta\alpha$ is orthogonal to every
  other signal, motivating its selection as the second
  \textsc{\textbf{HybridAL}} variant alongside $\Delta$Acc.}
  \label{fig:signal_correlation}
\end{figure}

\begin{figure}[t]
  \centering
  \includegraphics[width=0.9\columnwidth]{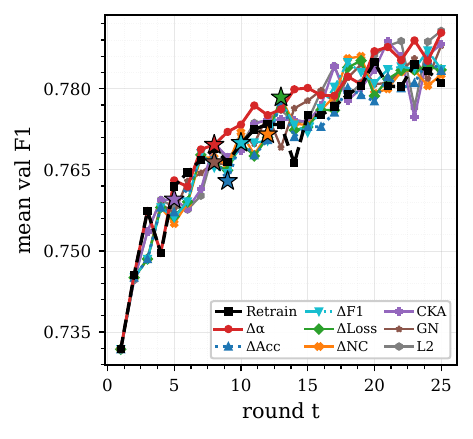}
  \caption{Per-round mean validation \fone{} averaged across
  all six datasets (equal weight per dataset); $\star$ marks
  each signal's mean switch round across the dataset/seed
  firings.}
  \label{fig:signal_ablation_mean_f1}
\end{figure}

Among the firing signals, $\Delta\alpha$ achieves the highest
mean test \fone{} ($0.793$) at the lowest training cost
($424$~s), and $\Delta$Acc is the closest performance-based
runner-up (\fone{}\,$=0.791$, time $438$~s) while requiring
only a standard validation forward pass; the two signals dominate the eight
on overall rank in Figure~\ref{fig:signal_ablation_pareto}.
We select $\Delta$Acc over the nearly redundant $\Delta$F1
(Spearman $\rho{=}0.8$ between per-round signal values,
Figure~\ref{fig:signal_correlation}) for its computational
simplicity: accuracy is a single scalar, whereas macro-F1
requires per-class aggregation. $\Delta\alpha$ is orthogonal
to every other signal ($\rho \approx 0$,
Figure~\ref{fig:signal_correlation}), confirming it captures
complementary information: weight spectral statistics rather
than task-level performance. We therefore adopt both as
independent \textsc{\textbf{HybridAL}} variants in the main results
(\S\ref{sec:results}). Figure~\ref{fig:signal_ablation_mean_f1}
confirms that all firing signals track \textsc{Retrain}'s
per-round validation \fone{} throughout training, with
$\Delta\alpha$ switching earliest.

Test NLL exposes the calibration cost of switching: the
firing performance-based signals ($\Delta$Acc, $\Delta$F1,
$\Delta$Loss, $\Delta$NC) average $0.57$--$0.70$ vs.\
\textsc{Retrain}'s $0.53$, and the model-based $\Delta\alpha$ averages
$0.74$. Even the best signal ($\Delta$Loss at $0.57$) does
not match \textsc{Retrain} on NLL, consistent with the
calibration-time trade-off in \S\ref{sec:time_nll}; we adopt
$\Delta\alpha$ and $\Delta$Acc despite this trade-off because
their (\fone{}, time) positions are Pareto-dominant among the
eight, and calibration is then traded against time via the
choice of signal. Although $\Delta$Loss achieves the best
NLL among firing signals ($0.574$), it fires on only $43\%$
of cells and fails entirely on IMDb, defaulting to
\textsc{Retrain}'s cost on non-firing datasets. Finally, $\ell_2$ weight distance never fires on
any of the six datasets, CKA fires only on AG News and Yahoo,
and gradient norm fires only on IMDb, SST-2, and TweetEval;
the corresponding cells in
Table~\ref{tab:appendix_signal_ablation} sit within one seed
std of \textsc{Retrain} on every dataset, confirming that running
\textsc{\textbf{HybridAL}} with a non-firing signal is statistically
indistinguishable from \textsc{Retrain}.

\begin{table*}[t]
  \centering
  \scriptsize
  \setlength{\tabcolsep}{3pt}
  \begin{tabular}{lcccccc|c}
  \toprule
  \multicolumn{8}{l}{\textbf{Test $F_1$ $\uparrow$}} \\
  Signal & IMDb & AGNews & Jigsaw & SST-2 & TwtEv & Yahoo & \textbf{Mean} \\
  \midrule
  \textsc{Retrain} & \cellcolor{green!30} 0.819\,$\pm$\,0.021 & \cellcolor{green!0} 0.900\,$\pm$\,0.004 & \cellcolor{green!29} 0.881\,$\pm$\,0.003 & \cellcolor{green!0} 0.835\,$\pm$\,0.021 & \cellcolor{green!52} 0.634\,$\pm$\,0.023 & \cellcolor{green!55} 0.667\,$\pm$\,0.004 & \cellcolor{green!0} \textbf{0.789}\,$\pm$\,0.105 \\
  $\Delta$Acc & \cellcolor{green!43} 0.825\,$\pm$\,0.009 & \cellcolor{green!45} 0.904\,$\pm$\,0.003 & \cellcolor{green!31} 0.881\,$\pm$\,0.007 & \cellcolor{green!23} 0.847\,$\pm$\,0.023 & \cellcolor{green!51} 0.633\,$\pm$\,0.009 & \cellcolor{green!17} 0.657\,$\pm$\,0.006 & \cellcolor{green!27} \textbf{0.791}\,$\pm$\,0.109 \\
  $\Delta\alpha$ & \cellcolor{green!44} 0.826\,$\pm$\,0.006 & \cellcolor{green!13} 0.901\,$\pm$\,0.005 & \cellcolor{green!12} 0.880\,$\pm$\,0.007 & \cellcolor{green!50} 0.861\,$\pm$\,0.010 & \cellcolor{green!50} 0.633\,$\pm$\,0.012 & \cellcolor{green!16} 0.657\,$\pm$\,0.004 & \cellcolor{green!49} \textbf{0.793}\,$\pm$\,0.109 \\
  $\Delta$F1 & \cellcolor{green!53} 0.830\,$\pm$\,0.011 & \cellcolor{green!45} 0.904\,$\pm$\,0.003 & \cellcolor{green!20} 0.880\,$\pm$\,0.005 & \cellcolor{green!23} 0.847\,$\pm$\,0.023 & \cellcolor{green!42} 0.628\,$\pm$\,0.011 & \cellcolor{green!14} 0.656\,$\pm$\,0.006 & \cellcolor{green!21} \textbf{0.791}\,$\pm$\,0.111 \\
  $\Delta$Loss & \cellcolor{green!0} 0.804\,$\pm$\,0.022 & \cellcolor{green!12} 0.901\,$\pm$\,0.005 & \cellcolor{green!55} 0.883\,$\pm$\,0.005 & \cellcolor{green!35} 0.853\,$\pm$\,0.020 & \cellcolor{green!55} 0.636\,$\pm$\,0.006 & \cellcolor{green!35} 0.662\,$\pm$\,0.005 & \cellcolor{green!6} \textbf{0.790}\,$\pm$\,0.107 \\
  $\Delta$NC & \cellcolor{green!18} 0.812\,$\pm$\,0.024 & \cellcolor{green!26} 0.903\,$\pm$\,0.004 & \cellcolor{green!8} 0.879\,$\pm$\,0.006 & \cellcolor{green!55} 0.863\,$\pm$\,0.011 & \cellcolor{green!42} 0.628\,$\pm$\,0.013 & \cellcolor{green!0} 0.653\,$\pm$\,0.009 & \cellcolor{green!6} \textbf{0.790}\,$\pm$\,0.112 \\
  GradNorm & \cellcolor{green!55} 0.831\,$\pm$\,0.008 & \cellcolor{green!55} 0.905\,$\pm$\,0.002 & \cellcolor{green!53} 0.883\,$\pm$\,0.008 & \cellcolor{green!48} 0.860\,$\pm$\,0.010 & \cellcolor{green!0} 0.601\,$\pm$\,0.039 & \cellcolor{green!32} 0.661\,$\pm$\,0.007 & \cellcolor{green!9} \textbf{0.790}\,$\pm$\,0.119 \\
  $\ell_2$ & \cellcolor{green!40} 0.823\,$\pm$\,0.010 & \cellcolor{green!21} 0.902\,$\pm$\,0.003 & \cellcolor{green!0} 0.879\,$\pm$\,0.005 & \cellcolor{green!52} 0.862\,$\pm$\,0.008 & \cellcolor{green!45} 0.630\,$\pm$\,0.022 & \cellcolor{green!45} 0.664\,$\pm$\,0.006 & \cellcolor{green!55} \textbf{0.793}\,$\pm$\,0.109 \\
  CKA & \cellcolor{green!40} 0.823\,$\pm$\,0.010 & \cellcolor{green!12} 0.901\,$\pm$\,0.009 & \cellcolor{green!0} 0.879\,$\pm$\,0.005 & \cellcolor{green!52} 0.862\,$\pm$\,0.008 & \cellcolor{green!45} 0.630\,$\pm$\,0.022 & \cellcolor{green!23} 0.659\,$\pm$\,0.004 & \cellcolor{green!41} \textbf{0.792}\,$\pm$\,0.110 \\
  \midrule
  \addlinespace[2pt]
  \multicolumn{8}{l}{\textbf{Test NLL $\downarrow$}} \\
  Signal & IMDb & AGNews & Jigsaw & SST-2 & TwtEv & Yahoo & \textbf{Mean} \\
  \midrule
  \textsc{Retrain} & \cellcolor{green!52} 0.426\,$\pm$\,0.039 & \cellcolor{green!52} 0.320\,$\pm$\,0.008 & \cellcolor{green!48} 0.133\,$\pm$\,0.008 & \cellcolor{green!30} 0.391\,$\pm$\,0.029 & \cellcolor{green!54} 0.814\,$\pm$\,0.066 & \cellcolor{green!55} 1.108\,$\pm$\,0.018 & \cellcolor{green!52} \textbf{0.532}\,$\pm$\,0.336 \\
  $\Delta$Acc & \cellcolor{green!13} 0.599\,$\pm$\,0.137 & \cellcolor{green!7} 0.432\,$\pm$\,0.028 & \cellcolor{green!22} 0.147\,$\pm$\,0.015 & \cellcolor{green!0} 0.451\,$\pm$\,0.076 & \cellcolor{green!28} 1.104\,$\pm$\,0.261 & \cellcolor{green!13} 1.470\,$\pm$\,0.216 & \cellcolor{green!11} \textbf{0.700}\,$\pm$\,0.477 \\
  $\Delta\alpha$ & \cellcolor{green!0} 0.658\,$\pm$\,0.052 & \cellcolor{green!1} 0.446\,$\pm$\,0.018 & \cellcolor{green!22} 0.147\,$\pm$\,0.016 & \cellcolor{green!1} 0.448\,$\pm$\,0.063 & \cellcolor{green!8} 1.321\,$\pm$\,0.202 & \cellcolor{green!17} 1.435\,$\pm$\,0.190 & \cellcolor{green!0} \textbf{0.743}\,$\pm$\,0.494 \\
  $\Delta$F1 & \cellcolor{green!17} 0.583\,$\pm$\,0.119 & \cellcolor{green!7} 0.432\,$\pm$\,0.028 & \cellcolor{green!33} 0.141\,$\pm$\,0.014 & \cellcolor{green!0} 0.451\,$\pm$\,0.076 & \cellcolor{green!35} 1.025\,$\pm$\,0.298 & \cellcolor{green!11} 1.486\,$\pm$\,0.222 & \cellcolor{green!14} \textbf{0.686}\,$\pm$\,0.475 \\
  $\Delta$Loss & \cellcolor{green!50} 0.435\,$\pm$\,0.037 & \cellcolor{green!13} 0.418\,$\pm$\,0.025 & \cellcolor{green!0} 0.158\,$\pm$\,0.020 & \cellcolor{green!37} 0.376\,$\pm$\,0.037 & \cellcolor{green!49} 0.875\,$\pm$\,0.163 & \cellcolor{green!46} 1.184\,$\pm$\,0.176 & \cellcolor{green!42} \textbf{0.574}\,$\pm$\,0.364 \\
  $\Delta$NC & \cellcolor{green!44} 0.463\,$\pm$\,0.081 & \cellcolor{green!5} 0.435\,$\pm$\,0.028 & \cellcolor{green!20} 0.148\,$\pm$\,0.017 & \cellcolor{green!12} 0.427\,$\pm$\,0.087 & \cellcolor{green!30} 1.085\,$\pm$\,0.262 & \cellcolor{green!14} 1.462\,$\pm$\,0.210 & \cellcolor{green!18} \textbf{0.670}\,$\pm$\,0.479 \\
  GradNorm & \cellcolor{green!27} 0.537\,$\pm$\,0.094 & \cellcolor{green!55} 0.313\,$\pm$\,0.013 & \cellcolor{green!55} 0.129\,$\pm$\,0.015 & \cellcolor{green!10} 0.431\,$\pm$\,0.062 & \cellcolor{green!0} 1.414\,$\pm$\,0.162 & \cellcolor{green!54} 1.114\,$\pm$\,0.022 & \cellcolor{green!22} \textbf{0.656}\,$\pm$\,0.469 \\
  $\ell_2$ & \cellcolor{green!55} 0.414\,$\pm$\,0.037 & \cellcolor{green!51} 0.323\,$\pm$\,0.012 & \cellcolor{green!47} 0.134\,$\pm$\,0.011 & \cellcolor{green!55} 0.339\,$\pm$\,0.008 & \cellcolor{green!55} 0.805\,$\pm$\,0.045 & \cellcolor{green!54} 1.117\,$\pm$\,0.019 & \cellcolor{green!55} \textbf{0.522}\,$\pm$\,0.341 \\
  CKA & \cellcolor{green!55} 0.414\,$\pm$\,0.037 & \cellcolor{green!0} 0.449\,$\pm$\,0.027 & \cellcolor{green!47} 0.134\,$\pm$\,0.011 & \cellcolor{green!55} 0.339\,$\pm$\,0.008 & \cellcolor{green!55} 0.805\,$\pm$\,0.045 & \cellcolor{green!0} 1.585\,$\pm$\,0.069 & \cellcolor{green!30} \textbf{0.621}\,$\pm$\,0.484 \\
  \midrule
  \addlinespace[2pt]
  \multicolumn{8}{l}{\textbf{Training time (s) $\downarrow$}} \\
  Signal & IMDb & AGNews & Jigsaw & SST-2 & TwtEv & Yahoo & \textbf{Mean} \\
  \midrule
  \textsc{Retrain} & \cellcolor{green!21} 533\,$\pm$\,23 & \cellcolor{green!17} 690\,$\pm$\,34 & \cellcolor{green!22} 720\,$\pm$\,39 & \cellcolor{green!29} 372\,$\pm$\,18 & \cellcolor{green!16} 402\,$\pm$\,10 & \cellcolor{green!10} 655\,$\pm$\,21 & \cellcolor{green!19} \textbf{562}\,$\pm$\,141 \\
  $\Delta$Acc & \cellcolor{green!30} 497\,$\pm$\,90 & \cellcolor{green!54} 457\,$\pm$\,12 & \cellcolor{green!52} 530\,$\pm$\,10 & \cellcolor{green!47} 329\,$\pm$\,63 & \cellcolor{green!40} 335\,$\pm$\,66 & \cellcolor{green!42} 478\,$\pm$\,130 & \cellcolor{green!51} \textbf{438}\,$\pm$\,104 \\
  $\Delta\alpha$ & \cellcolor{green!55} 410\,$\pm$\,38 & \cellcolor{green!51} 478\,$\pm$\,45 & \cellcolor{green!55} 509\,$\pm$\,46 & \cellcolor{green!55} 312\,$\pm$\,18 & \cellcolor{green!48} 314\,$\pm$\,51 & \cellcolor{green!34} 520\,$\pm$\,134 & \cellcolor{green!55} \textbf{424}\,$\pm$\,106 \\
  $\Delta$F1 & \cellcolor{green!30} 499\,$\pm$\,85 & \cellcolor{green!54} 457\,$\pm$\,12 & \cellcolor{green!50} 545\,$\pm$\,39 & \cellcolor{green!47} 330\,$\pm$\,63 & \cellcolor{green!24} 378\,$\pm$\,52 & \cellcolor{green!44} 465\,$\pm$\,132 & \cellcolor{green!49} \textbf{445}\,$\pm$\,100 \\
  $\Delta$Loss & \cellcolor{green!3} 595\,$\pm$\,27 & \cellcolor{green!30} 607\,$\pm$\,88 & \cellcolor{green!36} 632\,$\pm$\,34 & \cellcolor{green!16} 401\,$\pm$\,51 & \cellcolor{green!17} 397\,$\pm$\,32 & \cellcolor{green!18} 613\,$\pm$\,141 & \cellcolor{green!24} \textbf{541}\,$\pm$\,123 \\
  $\Delta$NC & \cellcolor{green!22} 528\,$\pm$\,89 & \cellcolor{green!50} 482\,$\pm$\,48 & \cellcolor{green!33} 651\,$\pm$\,88 & \cellcolor{green!28} 375\,$\pm$\,29 & \cellcolor{green!23} 382\,$\pm$\,34 & \cellcolor{green!47} 450\,$\pm$\,149 & \cellcolor{green!41} \textbf{478}\,$\pm$\,123 \\
  GradNorm & \cellcolor{green!44} 448\,$\pm$\,116 & \cellcolor{green!0} 797\,$\pm$\,31 & \cellcolor{green!0} 865\,$\pm$\,54 & \cellcolor{green!18} 398\,$\pm$\,72 & \cellcolor{green!55} 294\,$\pm$\,27 & \cellcolor{green!4} 692\,$\pm$\,16 & \cellcolor{green!14} \textbf{582}\,$\pm$\,224 \\
  $\ell_2$ & \cellcolor{green!0} 606\,$\pm$\,28 & \cellcolor{green!3} 776\,$\pm$\,17 & \cellcolor{green!6} 826\,$\pm$\,44 & \cellcolor{green!0} 438\,$\pm$\,12 & \cellcolor{green!0} 445\,$\pm$\,16 & \cellcolor{green!0} 713\,$\pm$\,23 & \cellcolor{green!0} \textbf{634}\,$\pm$\,156 \\
  CKA & \cellcolor{green!0} 604\,$\pm$\,28 & \cellcolor{green!55} 451\,$\pm$\,18 & \cellcolor{green!7} 817\,$\pm$\,43 & \cellcolor{green!3} 431\,$\pm$\,11 & \cellcolor{green!1} 441\,$\pm$\,16 & \cellcolor{green!55} 404\,$\pm$\,65 & \cellcolor{green!29} \textbf{525}\,$\pm$\,152 \\
  \midrule
  \addlinespace[2pt]
  \multicolumn{8}{l}{\textbf{Switching rate (\%)}} \\
  Signal & IMDb & AGNews & Jigsaw & SST-2 & TwtEv & Yahoo & \textbf{Mean} \\
  \midrule
  $\Delta$Acc & \cellcolor{green!55} 100 & \cellcolor{green!55} 100 & \cellcolor{green!55} 100 & \cellcolor{green!55} 100 & \cellcolor{green!33} 60 & \cellcolor{green!55} 100 & \cellcolor{green!53} \textbf{93} \\
  $\Delta\alpha$ & \cellcolor{green!55} 100 & \cellcolor{green!55} 100 & \cellcolor{green!55} 100 & \cellcolor{green!55} 100 & \cellcolor{green!55} 100 & \cellcolor{green!44} 80 & \cellcolor{green!55} \textbf{97} \\
  $\Delta$F1 & \cellcolor{green!55} 100 & \cellcolor{green!55} 100 & \cellcolor{green!55} 100 & \cellcolor{green!55} 100 & \cellcolor{green!22} 40 & \cellcolor{green!55} 100 & \cellcolor{green!51} \textbf{90} \\
  $\Delta$Loss & \cellcolor{green!0} 0 & \cellcolor{green!55} 100 & \cellcolor{green!55} 100 & \cellcolor{green!11} 20 & \cellcolor{green!11} 20 & \cellcolor{green!11} 20 & \cellcolor{green!25} \textbf{43} \\
  $\Delta$NC & \cellcolor{green!11} 20 & \cellcolor{green!55} 100 & \cellcolor{green!44} 80 & \cellcolor{green!33} 60 & \cellcolor{green!33} 60 & \cellcolor{green!44} 80 & \cellcolor{green!38} \textbf{67} \\
  GradNorm & \cellcolor{green!44} 80 & \cellcolor{green!0} 0 & \cellcolor{green!0} 0 & \cellcolor{green!33} 60 & \cellcolor{green!55} 100 & \cellcolor{green!0} 0 & \cellcolor{green!23} \textbf{40} \\
  $\ell_2$ & \cellcolor{green!0} 0 & \cellcolor{green!0} 0 & \cellcolor{green!0} 0 & \cellcolor{green!0} 0 & \cellcolor{green!0} 0 & \cellcolor{green!0} 0 & \cellcolor{green!0} \textbf{0} \\
  CKA & \cellcolor{green!0} 0 & \cellcolor{green!55} 100 & \cellcolor{green!0} 0 & \cellcolor{green!0} 0 & \cellcolor{green!0} 0 & \cellcolor{green!55} 100 & \cellcolor{green!19} \textbf{33} \\
  \bottomrule
  \end{tabular}
  \caption{Per-dataset signal-ablation on DistilBERT (5 seeds per cell; all \textsc{\textbf{HybridAL}} signals at $\varepsilon{=}0.5$, $k{=}3$). Four sub-tables stack the per-(signal $\times$ dataset) means of test \fone{}, test NLL, total training time, and switching rate (fraction of runs in which the signal triggered a switch within 25 rounds). The final \textbf{Mean} column averages across the six datasets. Cell shading is per column; darker green = better.}
  \label{tab:appendix_signal_ablation}
\end{table*}

\begin{figure*}[h]
  \centering
  \includegraphics[width=\textwidth]{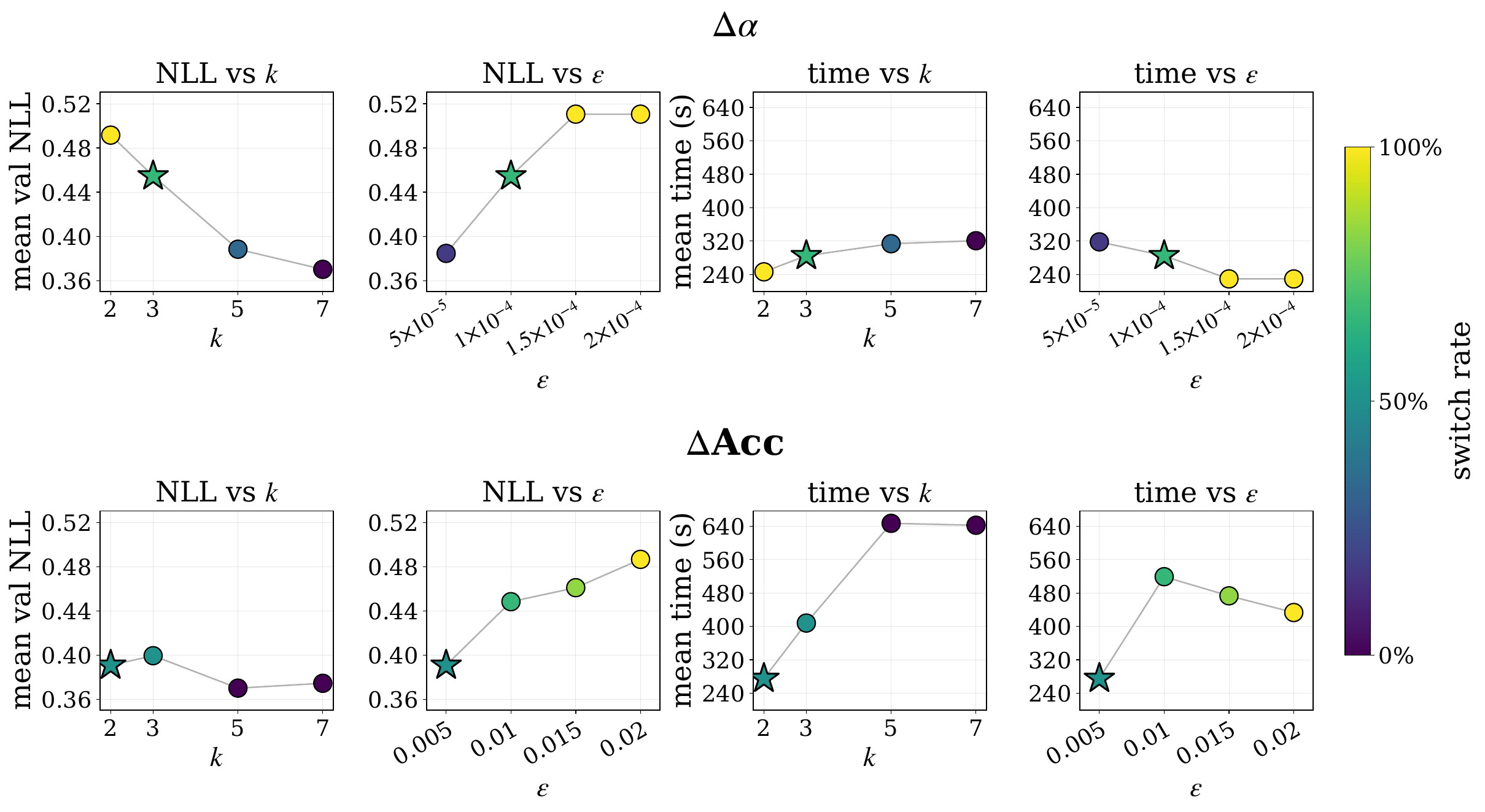}
  \caption{Hyperparameter cross-sections through
    $(\varepsilon^{\!*}, k^{\!*})$ for $\Delta\alpha$ (top row,
    $\varepsilon^{\!*}{=}10^{-4}$, $k^{\!*}{=}3$) and $\Delta$Acc
    (bottom row, $\varepsilon^{\!*}{=}5{\times}10^{-3}$,
    $k^{\!*}{=}2$). In each row, the leftmost two panels sweep
    $k$ (with $\varepsilon{=}\varepsilon^{\!*}$) and $\varepsilon$
    (with $k{=}k^{\!*}$) for mean val NLL; the rightmost two
    panels do the same for mean training time. Marker color
    encodes switch rate; the chosen cell is marked with $\star$.}
  \label{fig:hyperparameter_sweep}
\end{figure*}

\section{Hyperparameter Tuning}
\label{app:hyperparam_tuning}

Table~\ref{tab:hyperparameter_grid} and
Figure~\ref{fig:hyperparameter_sweep} give the $(\varepsilon, k)$
sensitivity for $\Delta\alpha$ and $\Delta$Acc, whose
informative $\varepsilon$ regimes differ by two orders of
magnitude. We sweep $\{5{\times}10^{-5},\, 10^{-4},\,
1.5{\times}10^{-4},\, 2{\times}10^{-4}\}$ for $\Delta\alpha$ and
$\{5{\times}10^{-3},\, 10^{-2},\, 1.5{\times}10^{-2},\,
2{\times}10^{-2}\}$ for $\Delta$Acc, each anchored so the
smallest $\varepsilon$ fires within the first five rounds and
the largest within the budget; $k \in \{2, 3, 5, 7\}$ is
shared.

We pick $(\varepsilon^{\!*}, k^{\!*})$ in two stages. First,
restrict to cells whose mean validation \fone{} is within
$0.5\%$ of the grid top:
\begin{equation}
  \mathcal{C}
  = \{(\varepsilon, k):\ \fone{}(\varepsilon, k)
       \geq 0.995\,\fone{}_{\max}\}.
  \label{eq:hp_C}
\end{equation}
This tolerance sits well below the per-cell seed standard
deviation ($\sim 5$ pp), so $\mathcal{C}$ contains every cell
statistically tied with the \fone{} best.

Second, among cells in $\mathcal{C}$ we need to trade off
time against calibration, since the fastest cell is not
always the best-calibrated. We minimise a combined score of
wall-clock training time and validation NLL, each normalised
by the in-set minimum:
\begin{equation}
  \mathrm{score}(\varepsilon, k)
  = \frac{\mathrm{Time}(\varepsilon, k)}
         {\mathrm{Time}^{\mathcal{C}}_{\min}}
  + \lambda \cdot
    \frac{\mathrm{NLL}(\varepsilon, k)}
         {\mathrm{NLL}^{\mathcal{C}}_{\min}},
  \label{eq:hp_score}
\end{equation}
\begin{equation}
  (\varepsilon^{\!*}, k^{\!*})
  = \arg\min_{(\varepsilon, k) \in \mathcal{C}}
    \mathrm{score}(\varepsilon, k),
  \label{eq:hp_select}
\end{equation}
with $\lambda = 0.5$. Both normalised terms lie in
$[1, \infty)$ and per-axis normalisation removes their unit
gap, so $\lambda$ controls the calibration-vs.-time weight on
$\mathcal{C}$. We pick $\lambda<1$ to favour time savings,
since \fone{} within $\mathcal{C}$ is already statistically
equivalent to the grid top; smaller $\lambda$ would over-weight
time, while larger $\lambda$ would push toward \textsc{Retrain}
on both axes. The selected cells are stable across
$\lambda \in [0.3, 0.7]$; we report $\lambda{=}0.5$ as the
midpoint.

This selects $(10^{-4}, 3)$ for $\Delta\alpha$ (2 candidates in
$\mathcal{C}$; score $1.546$ vs.\ $1.561$) and
$(5{\times}10^{-3}, 2)$ for $\Delta$Acc (7 candidates; the
chosen cell jointly attains $\mathrm{Time}^{\mathcal{C}}_{\min}$
and $\mathrm{NLL}^{\mathcal{C}}_{\min}$, score $1.500$); both
are applied to all main-results experiments without retuning.

Across the grid, mean validation \fone{} varies by only $1.1$
pp while mean training time varies by $1.5\times$ to
$2.5\times$, so the choice is insensitive to \fone{} and driven
by the (time, NLL) trade-off captured by
Eq.~\ref{eq:hp_score}.

\begin{table*}[t]
  \centering
  \footnotesize
  \setlength{\tabcolsep}{4pt}
  \begin{tabular}{lcccccccccccc}
  \toprule
   & \multicolumn{4}{c}{mean val \fone{}} & \multicolumn{4}{c}{mean val NLL} & \multicolumn{4}{c}{mean time (s)} \\
  \cmidrule(lr){2-5}\cmidrule(lr){6-9}\cmidrule(lr){10-13}
  $\varepsilon$ & $k{=}2$ & $k{=}3$ & $k{=}5$ & $k{=}7$ & $k{=}2$ & $k{=}3$ & $k{=}5$ & $k{=}7$ & $k{=}2$ & $k{=}3$ & $k{=}5$ & $k{=}7$ \\
  \midrule
  \multicolumn{13}{l}{$\Delta\alpha$} \\
  5e-5 & 0.858 & 0.857 & 0.855 & 0.855 & 0.447 & 0.385 & 0.370 & 0.370 & 290.8 & 318.0 & 331.9 & 322.5 \\
  \cellcolor{gray!25} 1e-4 & 0.854 & \cellcolor{gray!25} 0.865 & 0.856 & 0.855 & 0.492 & \cellcolor{gray!25} 0.455 & 0.388 & 0.370 & 246.3 & \cellcolor{gray!25} 284.8 & 313.7 & 320.8 \\
  1.5e-4 & 0.855 & 0.855 & 0.858 & 0.855 & 0.498 & 0.511 & 0.435 & 0.370 & 214.3 & 229.4 & 289.3 & 320.9 \\
  2e-4 & 0.855 & 0.855 & 0.859 & 0.861 & 0.498 & 0.511 & 0.436 & 0.416 & 214.3 & 229.3 & 278.9 & 302.3 \\
  \midrule
  \multicolumn{13}{l}{$\Delta$Acc} \\
  \cellcolor{gray!25} 0.005 & \cellcolor{gray!25} 0.857 & 0.858 & 0.855 & 0.850 & \cellcolor{gray!25} 0.391 & 0.400 & 0.370 & 0.375 & \cellcolor{gray!25} 275.3 & 408.1 & 646.9 & 642.3 \\
  0.01 & 0.856 & 0.861 & 0.856 & 0.856 & 0.448 & 0.436 & 0.403 & 0.388 & 519.1 & 528.0 & 569.5 & 592.6 \\
  0.015 & 0.855 & 0.857 & 0.855 & 0.857 & 0.461 & 0.483 & 0.426 & 0.393 & 473.3 & 481.4 & 558.0 & 578.1 \\
  0.02 & 0.859 & 0.855 & 0.855 & 0.857 & 0.487 & 0.482 & 0.431 & 0.395 & 433.3 & 476.8 & 541.2 & 556.1 \\
  \bottomrule
  \end{tabular}
  \caption{$(\varepsilon, k)$ tuning grid for both signals. Three metrics shown side-by-side: mean val \fone{}, mean val NLL, and mean training time (s); for NLL and time, lower is better. \textcolor{gray}{Gray} cells mark the chosen $(\varepsilon^{\!\star}, k^{\!\star})$ applied to the main results.}
  \label{tab:hyperparameter_grid}
\end{table*}

\section{Robustness Ablations}
\label{app:robustness}

We test \textsc{\textbf{HybridAL}}'s sensitivity to five implementation choices
on DistilBERT (all ablations use the $\Delta\alpha$ variant
unless noted otherwise): the per-round training schedule
(\S\ref{app:early_stopping}), the entropy pre-filter
subset size $N$ (\S\ref{app:subset_ablation}), the initial pool size
(\S\ref{app:pool_size}), the acquisition batch size
(\S\ref{app:batch_size}), and the acquisition sampler
(\S\ref{app:sampler_ablation}).

\subsection{Early Stopping}
\label{app:early_stopping}
Our main experiments use max $10$ epochs per round with early
stopping (patience $2$ on val loss); the alternative is a
fixed budget of $5$ epochs. Both schedules deliver equivalent
\fone{} on AG News and IMDb (max gap $0.014$,
Table~\ref{tab:appendix_early_stopping}); early stopping is the
default because it lets \textsc{FineTune} converge in $3.4$ epochs vs.\
\textsc{Retrain}'s $5.5$ (\S\ref{sec:setup}), the source of \textsc{\textbf{HybridAL}}'s
time savings.

\begin{table}[H]
  \centering
  \small
  \setlength{\tabcolsep}{4pt}
  \resizebox{\columnwidth}{!}{%
  \begin{tabular}{ll|cc}
  \toprule
  Configuration & Method & AG News & IMDb \\
  \midrule
  fixed $5$ ep & \textsc{FineTune} & $0.892$\,$\pm$\,$0.005$ & $0.808$\,$\pm$\,$0.009$ \\
  fixed $5$ ep & \textsc{Retrain} & $0.897$\,$\pm$\,$0.004$ & $0.822$\,$\pm$\,$0.010$ \\
  max $10$ + ES & \textsc{FineTune} & $0.895$\,$\pm$\,$0.004$ & $0.818$\,$\pm$\,$0.007$ \\
  max $10$ + ES & \textsc{Retrain} & $0.892$\,$\pm$\,$0.007$ & $0.821$\,$\pm$\,$0.012$ \\
  \bottomrule
  \end{tabular}
  }
  \caption{Early-stopping spot check, 3 seeds per cell.}
  \label{tab:appendix_early_stopping}
\end{table}

\subsection{Subset Size Sensitivity}\label{app:subset_ablation}

Entropy-based acquisition ranks unlabelled candidates by predictive
entropy and selects the top $n$ for labelling each round. Scoring the
entire unlabelled pool every round is expensive: on IMDb's
${\sim}40\text{k}$ pool the entropy pass dominates the per-round
wall-clock cost for DistilBERT. We instead score a uniformly random subsample of $N$ candidates each round. To select $N$, we run both training strategies (\textsc{Retrain} and \textsc{FineTune}) for
$N \in \{100, 500, 1000, 2000, 5000, 10\text{k}, 20\text{k},
|\text{pool}|\}$ on IMDb and SST-2 with three seeds each, measuring
final test F1 (macro). Since $N$ is fixed once before all subsequent 
experiments and is not tuned per dataset, reporting test F1 here does 
not introduce data leakage.
\begin{figure}[H]
  \centering
  \includegraphics[width=\columnwidth]{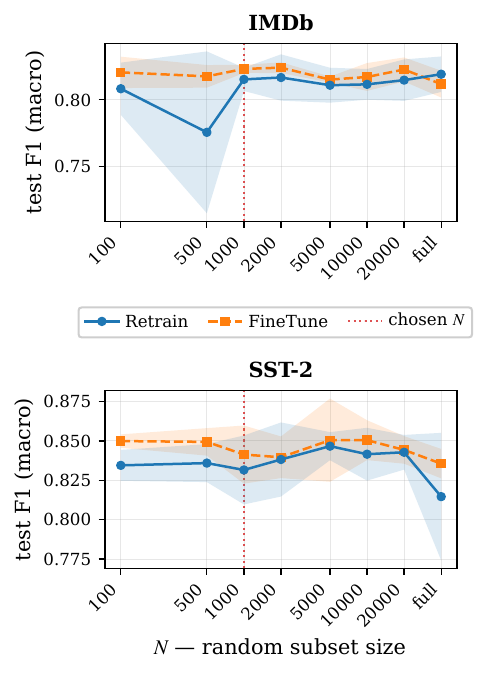}
   \caption{Final test \fone{} vs.\ entropy acquisition subsample
  size $N$ on IMDb (top) and SST-2 (bottom), mean and $\pm 1$ std
  across three seeds. Red dotted line: the chosen $N{=}1000$.}
  \label{fig:n_sensitivity}
\end{figure}

Figure~\ref{fig:n_sensitivity} shows the results. Test F1 is
essentially flat for $N \ge 1000$ on both datasets and both
strategies: the F1 band over $N \in [1000, |\text{pool}|]$ stays
within seed noise. At $N{=}500$ the IMDb \textsc{Retrain} mean drops
noticeably, suggesting that this subsample is too small to reliably 
surface the highest entropy candidates. Increasing $N$ to 
$|\text{pool}|$ (the full pool) provides no benefit and on SST-2 
even lowers F1 slightly, likely because the top entropy region is 
dominated by hard, noisy examples that the stochastic sampling at 
smaller $N$ smooths out. We therefore fix $N{=}1000$ for all 
subsequent experiments: the smallest value at which final F1 is 
statistically indistinguishable from larger $N$ on both datasets, 
while keeping the per round acquisition cost negligible.

\subsection{Initial Pool Size}
\label{app:pool_size}

We vary $|\mathcal{L}_0| \in \{50, 100, 200, 500\}$ for \textsc{Retrain}
and \textsc{\textbf{HybridAL}} (Table~\ref{tab:appendix_pool_size}).
\textsc{\textbf{HybridAL}} is within seed noise of \textsc{Retrain}
in 11/12 cells; the empirical switch round shifts inversely
with pool size (mean $7.8$--$12.0$), confirming that the stabilization
signal adapts to data availability.

\begin{table}[H]
  \centering
  \small
  \setlength{\tabcolsep}{4pt}
  \resizebox{\columnwidth}{!}{%
  \begin{tabular}{l|c|ccc}
  \toprule
  Method & $|\mathcal{L}_0|$ & AG News & IMDb & Yahoo \\
  \midrule
  \textsc{Retrain} & 50 & $0.899$\,$\pm$\,$0.003$ & $0.820$\,$\pm$\,$0.007$ & $0.658$\,$\pm$\,$0.004$ \\
  \textsc{\textbf{HybridAL}} & 50 & $0.902$\,$\pm$\,$0.004$ & $0.828$\,$\pm$\,$0.008$ & $0.654$\,$\pm$\,$0.008$ \\
  \textsc{Retrain} & 100 & $0.898$\,$\pm$\,$0.004$ & $0.821$\,$\pm$\,$0.015$ & $0.658$\,$\pm$\,$0.009$ \\
  \textsc{\textbf{HybridAL}} & 100 & $0.898$\,$\pm$\,$0.004$ & $0.826$\,$\pm$\,$0.014$ & $0.643$\,$\pm$\,$0.010$ \\
  \textsc{Retrain} & 200 & $0.900$\,$\pm$\,$0.004$ & $0.819$\,$\pm$\,$0.021$ & $0.667$\,$\pm$\,$0.004$ \\
  \textsc{\textbf{HybridAL}} & 200 & $0.902$\,$\pm$\,$0.007$ & $0.814$\,$\pm$\,$0.014$ & $0.656$\,$\pm$\,$0.004$$^{*}$ \\
  \textsc{Retrain} & 500 & $0.902$\,$\pm$\,$0.006$ & $0.836$\,$\pm$\,$0.004$ & $0.679$\,$\pm$\,$0.004$ \\
  \textsc{\textbf{HybridAL}} & 500 & $0.903$\,$\pm$\,$0.003$ & $0.829$\,$\pm$\,$0.011$ & $0.667$\,$\pm$\,$0.014$ \\
  \bottomrule
  \end{tabular}
  }
  \caption{Pool size ablation, 5 seeds per cell. $^{*}$
  $p<0.05$ vs.\ \textsc{Retrain} (paired $t$-test).}
  \label{tab:appendix_pool_size}
\end{table}

\subsection{Acquisition Batch Size}
\label{app:batch_size}
We vary $n \in \{16, 32, 64, 128\}$ on the same three datasets
(Table~\ref{tab:appendix_batch_size}). \textsc{\textbf{HybridAL}} is within seed noise of \textsc{Retrain}
in 11/12 cells; both methods scale similarly with $n$ ($+0.02$ to $+0.05$ \fone{} from
$n{=}16$ to $n{=}128$), so the parity claim is not specific to
$n{=}32$.

\begin{table}[H]
  \centering
  \small
  \setlength{\tabcolsep}{4pt}
  \resizebox{\columnwidth}{!}{%
  \begin{tabular}{l|c|ccc}
  \toprule
  Method & $n$ & AG News & IMDb & Yahoo \\
  \midrule
  \textsc{Retrain} & 16 & $0.896$\,$\pm$\,$0.005$ & $0.809$\,$\pm$\,$0.009$ & $0.646$\,$\pm$\,$0.013$ \\
  \textsc{\textbf{HybridAL}} & 16 & $0.895$\,$\pm$\,$0.006$ & $0.816$\,$\pm$\,$0.011$ & $0.651$\,$\pm$\,$0.008$ \\
  \textsc{Retrain} & 32 & $0.900$\,$\pm$\,$0.004$ & $0.819$\,$\pm$\,$0.021$ & $0.667$\,$\pm$\,$0.004$ \\
  \textsc{\textbf{HybridAL}} & 32 & $0.902$\,$\pm$\,$0.007$ & $0.814$\,$\pm$\,$0.014$ & $0.656$\,$\pm$\,$0.004$$^{*}$ \\
  \textsc{Retrain} & 64 & $0.912$\,$\pm$\,$0.003$ & $0.836$\,$\pm$\,$0.008$ & $0.679$\,$\pm$\,$0.003$ \\
  \textsc{\textbf{HybridAL}} & 64 & $0.911$\,$\pm$\,$0.003$ & $0.834$\,$\pm$\,$0.005$ & $0.668$\,$\pm$\,$0.009$ \\
  \textsc{Retrain} & 128 & $0.915$\,$\pm$\,$0.006$ & $0.846$\,$\pm$\,$0.010$ & $0.691$\,$\pm$\,$0.007$ \\
  \textsc{\textbf{HybridAL}} & 128 & $0.916$\,$\pm$\,$0.005$ & $0.852$\,$\pm$\,$0.004$ & $0.687$\,$\pm$\,$0.010$ \\
  \bottomrule
  \end{tabular}
  }
  \caption{Batch size ablation, 5 seeds per cell. $^{*}$
  $p<0.05$ vs.\ \textsc{Retrain} (paired $t$-test).}
  \label{tab:appendix_batch_size}
\end{table}

\subsection{Acquisition Sampler}
\label{app:sampler_ablation}
We compare \textsc{\textbf{HybridAL}} and \textsc{Retrain} across \textsc{Entropy} (the
default), \textsc{Random}, and
\textsc{BADGE}~\citep{ash2019deep}
(Table~\ref{tab:appendix_sampler}). \textsc{\textbf{HybridAL}} is within seed noise of \textsc{Retrain}
in 7/9 cells; both exceptions are on Yahoo Answers, where \textsc{\textbf{HybridAL}} is weakest in the main
results. Parity holds across uncertainty-based,
diversity-based, and random acquisition.

\begin{table}[H]
  \centering
  \small
  \setlength{\tabcolsep}{4pt}
  \resizebox{\columnwidth}{!}{%
  \begin{tabular}{l|l|ccc}
  \toprule
  Sampler & Method & AG News & IMDb & Yahoo \\
  \midrule
  \textsc{Entropy} & \textsc{Retrain} & $0.900$\,$\pm$\,$0.004$ & $0.819$\,$\pm$\,$0.021$ & $0.667$\,$\pm$\,$0.004$ \\
  \textsc{Entropy} & \textsc{\textbf{HybridAL}} & $0.902$\,$\pm$\,$0.007$ & $0.814$\,$\pm$\,$0.014$ & $0.656$\,$\pm$\,$0.004$$^{*}$ \\
  \textsc{Random} & \textsc{Retrain} & $0.887$\,$\pm$\,$0.006$ & $0.825$\,$\pm$\,$0.007$ & $0.662$\,$\pm$\,$0.009$ \\
  \textsc{Random} & \textsc{\textbf{HybridAL}} & $0.886$\,$\pm$\,$0.007$ & $0.820$\,$\pm$\,$0.006$ & $0.651$\,$\pm$\,$0.014$ \\
  \textsc{BADGE} & \textsc{Retrain} & $0.897$\,$\pm$\,$0.004$ & $0.828$\,$\pm$\,$0.007$ & $0.668$\,$\pm$\,$0.004$ \\
  \textsc{BADGE} & \textsc{\textbf{HybridAL}} & $0.898$\,$\pm$\,$0.004$ & $0.833$\,$\pm$\,$0.003$ & $0.651$\,$\pm$\,$0.007$$^{*}$ \\
  \bottomrule
  \end{tabular}
  }
  \caption{Sampler ablation, 5 seeds per cell. $^{*}$ $p<0.05$
  vs.\ \textsc{Retrain} (paired $t$-test).}
  \label{tab:appendix_sampler}
\end{table}

\end{document}